\documentclass[journal]{IEEEtran}
\usepackage{amsmath,amscd}
\usepackage{graphicx}
\usepackage{algorithm}
\usepackage{algorithmic}
\usepackage{amssymb,array}
\usepackage{multirow}
\usepackage{multicol}
\usepackage{amsthm}

\usepackage{enumitem}
\usepackage{hyperref}
\usepackage{bm}
\usepackage{xcolor}
\usepackage[font=small,labelsep=period]{caption}
\usepackage{subcaption}
\usepackage{url}

\DeclareMathOperator*{\minimize}{minimize\,}
\DeclareMathOperator*{\st}{subject\ to\,}

\def\x{\bm{x}}
\def\l{\bm{l}}
\def\a{\bm{a}}
\def\s{\bm{s}}
\def\y{\bm{y}}
\def\n{\bm{n}}

\def\A{\bm{A}}

\def\v{\bm{v}}

\def\u{\bm{u}}
\def\v{\bm{v}}
\def\W{\bm{W}}

\def\N{\bm{N}}
\def\I{\bm{I}}
\def\V{\bm{V}}

\def\U{\bm{U}}

\def\Y{\bm{Y}}

\def\L{\bm{L}}
\def\X{\bm{X}}
\def\D{\bm{D}}

\def\S{\bm{S}}
\newcommand{\E}{\mathrm{E}}

\newcommand{\BE}{\begin{equation}}
\newcommand{\EE}{\end{equation}}
\newcommand{\BS}{\begin{subequations}}
\newcommand{\ES}{\end{subequations}}

\makeatletter

\allowdisplaybreaks

\makeatother

\begin{document}

\title{Denoising-based Turbo Message Passing for Compressed Video Background Subtraction}
\author{Zhipeng~Xue,~Xiaojun~Yuan,~\IEEEmembership{Senior~Member,~IEEE}, and Yang~Yang,~\IEEEmembership{Fellow,~IEEE}}

% \thanks{Z. Xue is with the School of Information Science and Technology, ShanghaiTech University, Shanghai 201210, China, also with the Shanghai Institute of Microsystem and Information Technology, Chinese Academy of Sciences, Shanghai, China, and also with University of Chinese Academy of Sciences, Beijing, China  (e-mail: xuezhp@shanghaitech.edu.cn). X. Yuan is with the Center for Intelligent Networking and Communications (CINC), University of Electronic Science and Technology of China, Chengdu 611731, China (e-mail: xjyuan@uestc.edu.cn). Y. Yang is with the School of Information Science and Technology, ShanghaiTech University, Shanghai 201210, China and also with the Shanghai Institute of Fog Computing Technology (SHIFT), Shanghai, China (e-mail: yangyang@shanghaitech.edu.cn). This work was supported in part by the National Key R$\&$D Program of China under Grant 2018YFB1801105.}

\maketitle

\begin{abstract}
In this paper, we consider the compressed video background subtraction problem that separates the background and foreground of a video from its compressed measurements. The background of a video usually lies in a low dimensional space and the foreground is usually sparse. More importantly, each video frame is a natural image that has textural patterns. By exploiting these properties, we develop a message passing algorithm termed offline denoising-based turbo message passing (DTMP). We show that these structural properties can be efficiently handled by the existing denoising techniques under the turbo message passing framework. We further extend the DTMP algorithm to the online scenario where the video data is collected in an online manner. The extension is based on the similarity/continuity between adjacent video frames. We adopt the optical flow method to refine the estimation of the foreground. We also adopt the sliding window based background estimation to reduce complexity. By exploiting the Gaussianity of messages, we develop the state evolution to characterize the per-iteration performance of offline and online DTMP. Comparing to the existing algorithms, DTMP can work at much lower compression rates, and can subtract the background successfully with a lower mean squared error and better visual quality for both offline and online compressed video background subtraction.
\end{abstract}


\IEEEpeerreviewmaketitle

\section{Introduction}
\label{sec:introduction}

Video background subtraction (VBS) is important for many applications including video surveillance, object detection, visual hull computation, hunman-machine interaction, background substitution, gesture recognition, etc. \cite{shah2014video,bouwmans2019background}. Many approaches have been proposed for VBS. Traditional simple approaches in \cite{lee2002back,mcfarlane1995segmentation,zheng2006extracting} use average, median or histogram analysis over time to represent the video background. Statistical approaches use single Gaussian \cite{wren1997pfinder}, Mixture of Gaussian \cite{stauffer1999adaptive}, Kernel density methods \cite{elgammal2000non}, and principal component analysis \cite{oliver2000bayesian, bouwmans2014robust} to represent background. In \cite{sigari2008fuzzy,el2008type,zhang2006fusing,el2008fuzzy} fuzzy background modeling is introduced that models the background using fuzzy running average \cite{sigari2008fuzzy}, type-2 fuzzy mixture of Gaussian \cite{el2008type}, Sugeno integral \cite{zhang2006fusing} or Choquet integral \cite{el2008fuzzy}. Recently, deep learning based approaches have been extensive studied \cite{bouwmans2019deep}. These learning-based approaches achieves good performance when dataset is large. Among the existing approaches robust principal component analysis (RPCA) \cite{candes2011robust,xu2010robust,ding2011bayesian,bouwmans2014robust} provides a robust model for separating video background and foreground. A video modeled as a data matrix can be decomposed as an addition of two component matrices, namely the background and the foreground. The background is modeled by a low-rank subspace since the background is usually static or changes slowly over time, while the moving foreground is modeled as a sparse matrix. Besides, the enormous success of information retrieval methods such as compressive sensing \cite{donoho2006compressed} and low-rank matrix completion \cite{candes2009exact} shows that structured data can be recovered from highly incomplete measurements. This inspires us to study the compressed VBS problem, i.e., to separate the background and the foreground of a video based on its compressed measurements.

The compressed VBS problem can be formulated as a compressed RPCA problem. As a generalization of the RPCA problem, compressed RPCA \cite{ganesh2012principal} aims to recover a low-rank matrix and a sparse matrix from the compressed measurements of their sum. Several algorithms for compressed RPCA have been proposed recently \cite{ganesh2012principal,waters2011sparcs,aravkin2014variational,xue2018turbo}. In \cite{ganesh2012principal}, a variant of the principal component pursuit (PCP) method is proposed for noiseless compressed RPCA. In \cite{waters2011sparcs}, a greedy algorithm is proposed to iteratively estimate the low-rank component and the sparse component. In \cite{aravkin2014variational}, a stable PCP method is proposed to handle the measurement noise. In \cite{xue2018turbo} a message passing based algorithm is proposed to achieve low-complexity recovery with fewer measurements.

For real applications, video data are mostly delivered in the form of video streams. This leads to a more challenging problem of online compressed VBS, i.e., to separate the background and the foreground of a video whenever a compressed data frame arrives. In \cite{van2018compressive}, the authors proposed an online algorithm by taking the similarities of the foreground frames into consideration. In \cite{prativadibhayankaram2017compressive}, the authors proposed an online algorithm that leverages the information from the previously separated foreground frames by using optical flow.

In the above mentioned approaches, the compressed VBS problem is formulated as an optimization problem to induce the low-rankness of the background by using the nuclear norm and the sparsity of the foreground by using the $l_1$-norm. However, besides low-rankness and sparsity, video frames possess far richer local and global structural properties (such as textures, edges, etc.) since each frame is a natural image. These image structures, if not appropriately exploited, may cause significant performance loss in solving the compressed VBS problem.

In this paper, we incorporate the image structural information into the formulation of the offline and online compressed VBS problems. We first design a message passing based algorithm termed offline denoising based turbo message passing (DTMP) for the offline compressed VBS problem. More specifically, we establish a probability model for the offline compressed VBS problem by considering the image structure information in addition to the low-rankness of the background and the sparsity of the foreground. Based on that, we develop a factor graph representation of the offline compressed VBS problem, and design the message passing algorithm by basically following the sum-product rule. To reduce computational complexity, appropriate approximations are introduced based on the turbo message passing principle \cite{xue2017denoising}. We further extend the DTMP algorithm to the online scenario. We use the sliding-window based background estimation to reduce the computation complexity, and exploit the continuity of the foreground frames by using the optical flow method. We characterize the performance of both the offline and online DTMP algorithms using two scalar functions termed state evolution. Simulation on real video datasets demonstrates that the offline and online DTMP achieve significant performance gains compared to their counterpart algorithms in terms of both visual quality and normalized mean-square error.

In our prior work \cite{xue2018turbo}, we proposed to solve the compressed RPCA problem by using the turbo message passing (TMP) framework, where low-rank and sparsity denoisers are adopted for the estimation of the low-rank background matrix and the sparse foreground matrix. The TMP algorithm can be applied to the offline compressed VBS problem. Compared with \cite{xue2018turbo}, the work in this paper has the following novelties and contributions.
\begin{itemize}
	\item We introduce appropriate approximations to the joint probability density function by taking into account the sparsity of the foreground, the low-rankness of the background, and more importantly the image structural information, based on which a factor graph of the offline compressed VBS problem is constructed.
	\item Based on the factor graph, we develop the offline DTMP algorithm by using the sum-product rule and Gaussian message approximations. Compared to the TMP algorithm in \cite{xue2018turbo}, the main difference of the DTMP algorithm is the inclusion of an image denoiser to handle the image structural information. To ensure good performance, we carefully design the input of the image denoiser (by dropping a non-Gaussian input component), based on which the extrinsic denoising principle developed in \cite{xue2017denoising} can be applied.
	\item We further consider the online scenario and develop the online DTMP algorithm. In specific, we employ a sliding-window based background estimation approach to reduce the computation complexity, and exploit the continuity of the foreground frames by using the optical flow method. 
	\item We characterize the behavior of both the offline and online DTMP algorithms by using the state evolution. We show that the performances of the DTMP algorithms can be accurately predicted by the state evolution.
\end{itemize}

% Among the existing approaches robust principal component analysis (RPCA) \cite{candes2011robust,xu2010robust,ding2011bayesian,bouwmans2014robust} provides a robust model for separating video background and foreground. 
% A video modeled as a data matrix can be decomposed as an addition of two component matrices, namely the background and the foreground. The background is modeled by a low-rank subspace since the background is usually static or changes slowly over time, while the moving foreground is modeled as a sparse matrix. Besides, the enormous success of information retrieval methods such as compressive sensing \cite{donoho2006compressed} and low-rank matrix completion \cite{candes2009exact} shows that structured data can be recovered from highly incomplete measurements.

The remainder of this paper is organized as follows. In Section \ref{sec_related}, we introduce the related works of this paper briefly. In Section \ref{sec_offline_dtmp}, we present the offline VBS problem, establish the offline DTMP algorithm based on message passing, and analyze the complexity of the offline DTMP algorithm. In Section \ref{sec_online_dtmp}, we present the online VBS problem, establish the online DTMP algorithm, and analyze its complexity. In Section \ref{sec_state_evo}, we develop the state evolution analysis for both the offline and online DTMP algorithms. In Section \ref{section_results}, we present the numerical results of the state evolution of the offline and online DTMP algorithms and compare the performance of the algorithms with their counterparts in different video datasets. In Section \ref{sec_conc}, we conclude the paper.

In this paper, we use bold capital letters to denote matrices and use bold lowercase letters to denote vectors. Denote by $\X^T$, $\text{rank}(\X)$, and $\text{Tr}(\X)$ the transpose, the rank, and the trace of matrix $\X$, respectively. Denote by $X_{i,j}$ the $(i,j)$-th entry of matrix $\X$, and by $\text{vec}(\X)$ the vector obtained by sequentially stacking the columns of $\X$. Denote by $\mathcal{A}$ a linear operator, and by $\mathcal{A}^T$ its adjoint linear operator. The inner product of two matrices is defined by $\left<\X,\Y\right>=\text{Tr}(\X\Y^T)$. $\I$ denotes the identity matrix with an appropriate size. $\|\X\|_F$ denotes the Frobenius norm of matrix $\X$ and $\|\x\|_2$ denotes the $l_2$ norm of vector $\x$.

\section{Related Work}\label{sec_related}
A large amount of work in the field of video background subtraction has been published in the literature. In this section, we give a brief review of the developments in this field from three aspects.

\subsection{Video Background Subtraction Methods}
Background subtraction methods can be mainly classified into the following categories:
\begin{itemize}
	\item \textbf{Conventional background subtraction methods} use the average \cite{lee2002back}, or the median \cite{mcfarlane1995segmentation}, or the histogram over time \cite{zheng2006extracting} to represent the video background. These methods can be easily implemented. However, modeling background with a single image requires a fixed background without noise and artifacts, which makes the performance of these methods not robust to real applications.
	\item \textbf{Statistical background subtraction methods} basicaly model the background pixels by using a probability density function (PDF) and learn the PDF from the video frames. The single-Gaussian method \cite{wren1997pfinder} assumes that the intensity values of a pixel over time can be modeled by a single Gaussian distribution. To handle dynamic backgrounds, a mixture-of-Gaussian method \cite{stauffer1999adaptive} is proposed. However, a fast varying background cannot be modeled accurately by a few Gaussians. To solve this problem, the kernel density estimation (KDE) method \cite{elgammal2000non} was proposed. The KDE method is time-consuming which limits its application. Recently, principal component analysis based subspace learning \cite{oliver2000bayesian, bouwmans2014robust} are widely used to construct a background model and RPCA based methods \cite{candes2011robust,xu2010robust,ding2011bayesian,bouwmans2014robust} provide a robust model for video background and foreground seperation. Other methods such as support vector based methods, and subspace learning methods also fall into this category.
	\item \textbf{Fuzzy background subtraction methods} use fuzzy running average \cite{sigari2008fuzzy} or type-2 fuzzy mixture of Gaussian \cite{el2008type} to model the video background. The forground is detcted using the Sugeno integral \cite{zhang2006fusing} or Choquet integral \cite{el2008fuzzy}.
	\item \textbf{Neural network background subtraction methods} model the background as the weights of a neural network which can be trained by using training video frames. The network is trained to classify each pixel of the input frame as background or foreground \cite{bouwmans2019deep}. In \cite{culibrk2006neural}, a neural network which forms an unsupervised Bayesian classifier for background modeling and foreground detection is proposed. In \cite{luque2008video}, a multivalued discrete neural network is used to detect and correct the deficiencies and errors of the Mixture of Gaussian algorithm. In \cite{luque2008neural}, an unsupervised competitive neural network to represent the background is constructed based on adaptive neighborhoods. 
	% Recently, many works been proposed \cite{luque2008dipolar}.
	\item \textbf{Clustering based background subtraction methods} suppose that pixels in the input frame can be represented by clusters. The K-means based method \cite{butler2005real}, the codebook based method \cite{kim2004background} and the sequential clustering method \cite{xiao2006background} belongs to this category.

\end{itemize}
Interested readers may refer to \cite{bouwmans2011recent,bouwmans2014traditional} for a more detailed review of the video background subtraction methods.

\subsection{Compressed Video Background Subtraction Methods}
In this paper, we consider the compressed video background subtraction problem. Several compressive RPCA methods have been proposed in \cite{ganesh2012principal,waters2011sparcs,zonoobi2013low,li2014recursive,aravkin2014variational,pan2017online,van2018online,van2018compressive,prativadibhayankaram2018compressive,xue2018turbo}. In \cite{ganesh2012principal}, a variant of the PCP method is proposed for noiseless compressed RPCA. In \cite{waters2011sparcs}, a greedy algorithm is proposed to iteratively estimate the low-rank component and the sparse component. In \cite{zonoobi2013low}, an improvement of SpaRCS is introduced to incorporate the extracted prior knowledge of the sparse component (foreground) and applied the method to surveillance video reconstruction. In \cite{li2014recursive}, an efficient algorithm is proposed to solve the compressed video background subtraction problem. The algorithm first solves the single-frame compressed sensing problem and then initialize the low-rank background based on a few recovered frames. After that, the background and the foreground are seperated in a ``frame-by-frame" fashion. In \cite{aravkin2014variational}, a stable PCP method is proposed to handle the measurement noise. In \cite{pan2017online}, an online compressed RPCA method is proposed to reduce the memory cost. In \cite{van2018online}, a decomposition method is proposed to solve an n-$l_1$ cluster-weighted minimization problem to decompose a sequence of frames into sparse and low-rank matrices. In \cite{van2018compressive}, the structure of the foreground is considered to further reduce the number of measurements. In \cite{prativadibhayankaram2018compressive}, based on the method in \cite{van2018online,van2018compressive}, the optial flow is further introduced to estimate motions between foreground frames. In \cite{xue2018turbo} a message passing based algorithm is proposed to achieve low-complexity recovery with fewer measurements.

\subsection{Message Passing Based Video Background Subtraction Methods}
Message passing based algorithms have been widely used in many areas including compressed sensing \cite{rangan2011generalized}, and low-rank matrix recovery \cite{parker2014bilinear}. Recently, the message passing approach has been extended for solving the compressed RPCA problem. In \cite{parker2014bilinear}, the authors extended the generalized approximate message passing (GAMP) \cite{rangan2011generalized} algorithm for solving the RPCA problem via matrix factorization, with the resulting algorithm termed bilinear GAMP (BiG-AMP). In \cite{parker2016parametric}, as an extension of BiG-AMP, the authors developed an algorithm termed parametric BiG-AMP (P-BiG-AMP) to solve problems including compressed RPCA. In our previous work \cite{xue2018turbo}, we proposed a turbo-type message passing (TMP) framework for the compressed RPCA problem. Low-rank and sparsity denoisers are adopted for the estimation of the low-rank background matrix and the sparse foreground matrix. The TMP algorithm can be applied to the offline compressed VBS problem considered in this paper, and will be used as a baseline for comparison.

% /////////////////////////////////////////
\section{Offline Denoising-Based Turbo Message Passing}\label{sec_offline_dtmp}

\subsection{Problem Formulation}\label{off_problem_form}
Consider a video with a sequence of frames. Let $w$ and $h$ be the width and the height of a video frame, respectively. Each frame of a video can be represented by a vector of length $n_1=hw$. Denote by $\x_i \in \mathbb{R}^{n_1\times 1}$ the $i$-th frame of the video, by $\s_i$ the foreground of the $i$-th frame, and by $\l_i$ the background of the $i$-th frame. Denote by $n_2$ the total number of frames. Let $\X=[\x_1, \x_2,\cdots,\x_{n_2}]=\L+\S\in \mathbb{R}^{n_1\times n_2}$ be the video matrix, where $\L=[\l_1, \l_2,\cdots,\l_{n_2}]$ is the background matrix, and $\S=[\s_1, \s_2,\cdots,\s_{n_2}]$ is the foreground matrix.

A linear measurement vector of the video is given by
\begin{align}
	\y = \mathcal{A}(\X)+\n=\mathcal{A}(\L+\S)+\n\label{off_com_vbs}
\end{align}
where $\n\in \mathbb{R}^{m\times 1}$ is an independent Gaussian measurement noise with zero mean and covariance matrix $\sigma^2 \I$, and $\mathcal{A}: \mathbb{R}^{n_1\times n_2}\rightarrow \mathbb{R}^{m\times 1}$ is a linear measurement operator. Then, the offline compressed VBS problem is defined as to recover the foreground component $\S$ and the background component $\L$ from the noisy observation $\y$. Here, ``offline" means that the foreground and background separation is carried out after collecting all the $n_2$ frames of the video.

Existing RPCA based approaches to solving the offline compressed VBS problem mostly exploit the following two properties of the video background and foreground: The background vectors $\{\l_i\}_{i=1}^{n_2}$ reside in a low dimensional subspace of $\mathbb{R}^{n_1}$, i.e., $\L$ is a low-rank matrix; and the foreground vectors $\{\s_i\}_{i=1}^{n_2}$ are sparse, i.e. $\S$ is a sparse matrix. However, these approaches do not take into account the fact that each frame $\x_i$ is a natural image. The textural features of $\x_i$ as an image can be exploited to enhance compressed VBS problem. In what follows, we aim to develop a novel message-passing algorithm for solving the compressed VBS problem by incorporating the textural features of each $\x_i$ as an image.

To start with, suppose that the prior distribution of the low-rank component $\L$ is $p(\L)$ and that the prior distribution of the sparse component $\S$ is $p(\S)$. Then, the joint probability density of $\y$, $\X$, $\L$, and $\S$ is given by
\begin{align}
	p(\y,\X,\L,\S) = p(\y|\X)p(\X|\L,\S)p(\L)p(\S),\label{off_promodel}
\end{align}
where $\L$ and $\S$ are assumed to be independent of each other. Then the offline compressed VBS problem can be expressed as a statistical inference problem:
\begin{align}
	\hat{\L} = \E[\L|\y] \text{\ and\ } \hat{\S}=\E[\S|\y].
	 % (\hat{\L},\hat{\S}) = \argmax_{\L,\S} p(\L,\S|\y).
	 \label{map_prob}
\end{align}
However, the conditional expectations in (\ref{map_prob}) are difficult to calculate since the prior distributions $p(\L)$ and $p(\S)$ are usually unavailable for real-world images and videos. Even if $p(\L)$ and $p(\S)$ are available, the evaluation of these conditional means is still computationally infeasible since high-dimensional integrals are involved. Message passing is a powerful low-complexity tool to provide near-optimal statistical inference. As inspired by the recent successful development of message passing algorithms for structured signal reconstruction problems \cite{donoho2010message,ma2015turbo}, we propose a message-passing based solution in the following subsections.

\subsection{Factor Graph Representation}
We first describe the factor graph representation of $p(\y,\X,\L,\S)$, based on which the message passing algorithm is developed. The factorization of $p(\y,\X,\L,\S)$ in (\ref{off_promodel}) cannot be used directly since the prior distributions $p(\L)$ and $p(\S)$ are difficult to acquire for real-world images and videos. To avoid this difficulty, we replace $p(\L)$ and $p(\S)$ by three separable constraints, namely, $f_L(\L)$ for $\L$ being low-rank, $f_S(\S)$ for $\S$ being sparse, and $f_X(\X)$ for $\X$ being a collection of natural images. Then, by noting $p(\y|\X)=\mathcal{N}(\y;\mathcal{A}(\X),\sigma^2\I)$ and $p(\X|\L,\S)=\delta(\X-\L-\S)$, we approximate the joint distribution of $\y$, $\X$, $\L$, and $\S$ in (\ref{off_promodel}) as
\begin{align}
	p(\y,\X,\L,\S) \approx &\ \mathcal{N}(\y;\mathcal{A}(\X),\sigma^2\I)\delta(\X-\L-\S)\notag\\
	&\times f_{X}(\X)f_{L}(\L)f_{S}(\S)
	\label{joint_prob}
\end{align}
where $\delta(\cdot)$ is the Dirac delta function. Note that the explicit expressions of $f_L(\L)$, $f_S(\S)$, and $f_X(\X)$ are not available in practice. We will discuss how to approximately evaluate the messages involving $f_L(\L)$, $f_S(\S)$, and $f_X(\X)$ in the next subsection.

\begin{figure}
	\centering
	\includegraphics[width=0.85\linewidth]{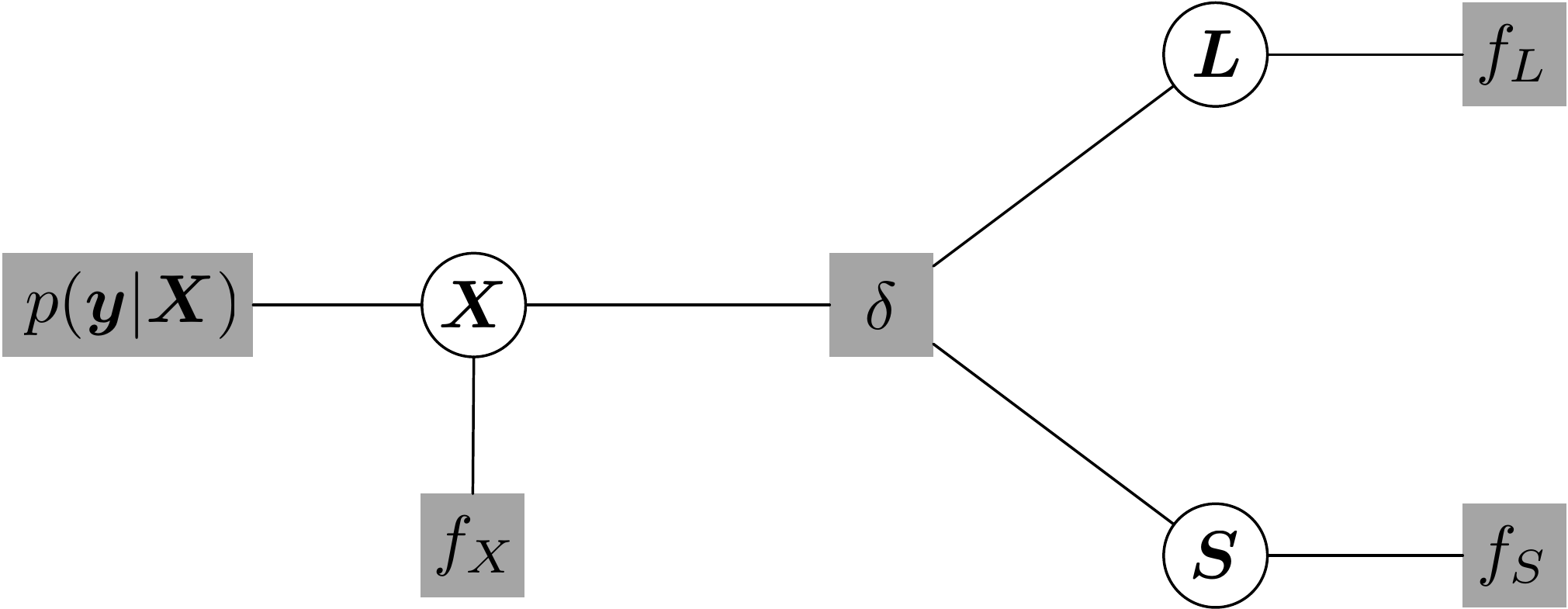}
	\caption{A graph representation of the factorization in (\ref{joint_prob}).}
	\label{factorgraph}
\end{figure}

Based on the factorization (\ref{joint_prob}), we construct the factor graph in Fig. \ref{factorgraph}, where each blank circle represents a variable node, each grey rectangle represents a factor node, and a factor node is connected to a variable node if the variable appears in the factor function. Note that $\delta$ is a shorthand of $\delta(\X-\L-\S)$ for notational brevity, and that the variable node $\y$ is omitted here since $\y$ is observed.

In the next subsection, we will derive the offline DTMP algorihtm based on the factor graph in Fig. \ref{factorgraph}. In our prior work \cite{xue2018turbo}, a turbo-type messgae passing algorithm is proposed for offline compressed RPCA problem that can be applied to offline compressed VBS problem.

\subsection{Derivation of Offline DTMP}\label{derivation_dtmp_batch}
%--------------y to X---------
% We are now ready to present the offline DTMP algorithm based on the factor graph in Fig. \ref{factorgraph}. 

The derivation of the algorithm mainly follows the sum-product rule \cite{kschischang2001factor} and the turbo message passing principle \cite{ma2015turbo}. We emphasize that, due to the unavailability of $f_{L}(\cdot)$, $f_{S}(\cdot)$, and $f_{X}(\cdot)$, exact evaluations of the messages based on the sum product rule are usually impossible. Following the message passing principle in \cite{xue2017denoising}, we will pass means and variances, rather than complete message functions, in the algorithm. Denote by $m_{A\rightarrow \X}(\X)$ the message from factor node $A$ to variable node $\X$. The corresponding mean and variance are respectively denoted by $\X_{A\rightarrow \X}$ and $v_{A\rightarrow \X}$. Similarly, denote by $m_{A\leftarrow \X}(\X)$ the message from variable node $\X$ to factor node $A$, with the corresponding mean and variance expressed as $\X_{A \leftarrow \X}$ and $v_{A\leftarrow \X}$. The messages in the factor graph are described below.

\subsubsection{Message from factor node \texorpdfstring{$p(\y|\X)$}{Lg} to variable node \texorpdfstring{$\X$}{Lg}} \label{derivation_dtmp_batch_1}
% \subsection{The classes \texorpdfstring{$\mathcal{L}(\gamma)$}{Lg}}
Define the message from factor node $p(\y|\X)$ to variable node $\X$ and the message from variable node $\X$ to factor node $p(\y|\X)$ respectively as
\BS
\begin{align}
	m_{\y\rightarrow \X}(\X) &= \prod_{i,j} m_{\y\rightarrow X_{i,j}}(X_{i,j})\\
	m_{\y\leftarrow \X}(\X) &= \prod_{i,j} m_{\y\leftarrow X_{i,j}}(X_{i,j}),
\end{align}\label{m_yx_xy}\ES
where from the sum-product rule, the message from node $p(\y|\X)$ to each variable $X_{i,j}$ is given by
\BS
\begin{align}
	m_{\y \rightarrow X_{i,j}}(X_{i,j}) &= \frac{\int_{\X_{/(i,j)}} p(\y|\X) m_{\y\leftarrow\X}(\X)}{m_{\y\leftarrow X_{i,j}}(X_{i,j})}\\
	&= \frac{m_{\y}(X_{i,j})}{m_{\y\leftarrow X_{i,j}}(X_{i,j})}, \forall i,j,
\end{align}\label{m_p_x}\ES
with 
\begin{align}
	m_{\y}(X_{i,j})=\int_{\X_{/(i,j)}} \mathcal{N}(\y;\mathcal{A}(\X),\sigma^2\I)m_{\y\leftarrow\X}(\X).\label{y_post}
\end{align}
In the above, $\y$ in the subscript is the shorthand for node $p(\y|\X)$, $\X_{/(i,j)}$ denotes the set includes all entries in $\X$ except the $(i,j)$-th entry, and $m_{\y}(X_{i,j})$ is the belief of $X_{i,j}$ at node $p(\y|\X)$. Define
\begin{align}
	m_{\y}(\X)=\prod_{i,j} m_{\y}(X_{i,j}).\label{m_y}
\end{align}
Then, by combining (\ref{m_yx_xy}), (\ref{m_p_x}), and (\ref{m_y}), we obtain an alternative expression of $m_{\y\rightarrow\X}$ given by
\begin{align}
	m_{\y\rightarrow\X}(\X) = \frac{m_{\y}(\X)}{m_{\y\leftarrow\X}(\X)}.\label{xext}
\end{align}

Assume that message $m_{\y\leftarrow \X}(\X)$ is Gaussian with mean $\X_{\y\leftarrow\X}$ and variance $v_{\y\leftarrow \X}$, i.e.
\begin{align}
	m_{\y\leftarrow \X}(\X)=\prod_{i,j} \mathcal{N}(X_{i,j};(\X_{\y\leftarrow\X})_{i,j},v_{\y\leftarrow \X}).\label{x_y_gaus}
\end{align}
Then, from (\ref{y_post}), (\ref{m_y}), and (\ref{x_y_gaus}), message $m_{\y}(\X)$ is also Gaussian with the mean and variance respectively given by $\X_y\in \mathbb{R}^{n_1\times n_2}$ and $v_y$. From \cite[Eq. 8]{ma2015turbo},
\BS
\begin{align}
	\text{vec}(\X_{\y})\! =\!&\, \text{vec}(\X_{\y\leftarrow \X})+ v_{\y\leftarrow\X}\A^T(v_{\y\leftarrow\X}\A\A^T\!+\!\sigma^2\I)^{-1}\notag\\
	&\times(\y\!-\!\A \text{vec}(\X_{\y\leftarrow \X})),\\
	v_{\y}\! =\!\frac{1}{n_1 n_2}&\text{Tr}(v_{\y\leftarrow\X} \I - v_{\y\leftarrow\X}^2\A^T(v_{\y\leftarrow\X}\A\A^T+\sigma^2\I)^{-1}\A),
\end{align}\label{linear_mean_var}\ES
where $\text{vec}(\X)=[\x_{1}^T,\x_{2}^T,\cdots,\x_{n_2}^T]^T$ with $\x_{i}$ being the $i$-th column vector of $\X$, and $\A$ is the matrix form of linear operator $\mathcal{A}$. Note that, when message $m_{\y\leftarrow \X}(\X)$ is not Gaussian, the mean and variance of $m_{\y}(\X)$ can still be well approximated by (\ref{linear_mean_var}) since (\ref{linear_mean_var}) corresponds to the linear minimum-mean square error (LMMSE) estimation of $\X$ given $\y$ with prior mean $\text{vec}(\X_{\y\leftarrow\X})$ and covariance $v_{\y \leftarrow\X}\I$.

From (\ref{xext}), (\ref{x_y_gaus}), and (\ref{linear_mean_var}), the mean and variance of message $m_{\y\rightarrow\X}(\X)$ are respectively given by
\BS 
\begin{align}
	\X_{\y\rightarrow\X}&=v_{\y\rightarrow\X} \left(\frac{\X_{\y}}{v_{\y}}-\frac{\X_{\y\leftarrow\X}}{v_{\y\leftarrow\X}}\right)\\
	v_{\y\rightarrow\X} &= \left(\frac{1}{v_{\y}}-\frac{1}{v_{\y\leftarrow\X}}\right)^{-1}.
\end{align}\label{ext_mean_var}\ES

For the sake of low computational complexity, we usually choose $\mathcal{A}$ as a partial orthogonal linear operator, i.e., $\mathcal{A}(\mathcal{A}^T(\a))=\a$ for an arbitrary vector $\a$. Equivalently, the matrix form of $\mathcal{A}$ satisfies the following property:
\begin{align}
	\A\A^T=\I.\label{part_orth_linop}
\end{align}
Then
\BS
\begin{align}
	 v_{\y} &=\frac{\text{Tr}\left(v_{\y\leftarrow\X} \I \!-\! v_{\y\leftarrow\X}^2\A^T(v_{\y\leftarrow\X}\A\A^T\!+\!\sigma^2\I)^{-1}\A\right)}{n_1 n_2}\\
	 % &=\frac{1}{n_1 n_2}\text{Tr}\left(v_{\y\leftarrow\X} \I - \frac{v_{\y\leftarrow\X}^2}{v_{\y\leftarrow\X}+\sigma^2}\A^T\A\right)\\
	 &=v_{\y\leftarrow\X}  -\frac{m}{n_1 n_2}\frac{v_{\y\leftarrow\X}^2}{v_{\y\leftarrow\X}+\sigma^2}.
\end{align}\label{avg_vx}\ES
Combining (\ref{linear_mean_var})-(\ref{avg_vx}), we obtain a simplified form of $\X_{\y\rightarrow\X}$ and $v_{\y\rightarrow\X}$ given by
\BS
\begin{align}
	\X_{\y\rightarrow\X} &= \X_{\y\leftarrow\X}+\frac{n_1 n_2}{m}\mathcal{A}^T(\y-\mathcal{A}(\X_{\y\leftarrow\X}))\\
	v_{\y\rightarrow\X}&=\frac{n_1 n_2}{m}(v_{\y\leftarrow\X}+\sigma^2)-v_{\y\leftarrow\X}.
\end{align}\label{m_v_m_y_X}\ES

%--------------
\subsubsection{Message from variable node \texorpdfstring{$\X$}{Lg} to factor node \texorpdfstring{$f_{X}$}{Lg}}\label{derivation_dtmp_batch_3_2}
Similarly to (\ref{m_yx_xy}), define the message from variable node $\X$ to factor node $f_{X}$ by
\begin{align}\label{X_fX}
	m_{f_{X} \leftarrow \X}(\X)&=\prod_{i,j}m_{f_{X} \leftarrow X_{i,j}}(X_{i,j})
\end{align}
where the message $m_{f_X\leftarrow X_{i,j}}(X_{i,j})$ is given by
\begin{align}
	m_{f_X\leftarrow X_{i,j}}(X_{i,j}) = m_{\delta\rightarrow X_{i,j}}(X_{i,j}) m_{\y\rightarrow X_{i,j}}(X_{i,j}).\label{m_fX_X}
\end{align}
By combining (\ref{X_fX}) and (\ref{m_fX_X}), we have
\BS
\begin{align}
	m_{f_X\leftarrow \X}(\X) &= \prod_{i,j} m_{\delta\rightarrow X_{i,j}}(X_{i,j}) \prod_{i,j}m_{\y\rightarrow X_{i,j}}(X_{i,j})\\
	&=m_{\delta\rightarrow\X}(\X)m_{\y\rightarrow\X}(\X).
\end{align}\label{m_fX_X_mtx}\ES
Recall from (\ref{ext_mean_var}) that the mean of $m_{\y\rightarrow\X}$ is the LMMSE estimator by treating $m_{\y\leftarrow\X}(\X)$ as the prior. It is known that, based on the central limit theorem, the estimation error of the LMMSE estimator is approximately Gaussian, and hence $m_{\y\rightarrow\X}(\X)$ can be approximated well as a Gaussian message.% \footnote{This Gaussian assumption is used widely in message passing based algorithms \cite{donoho2010message} and is proved later in \cite{bayati2011dynamics}. Later, we will provide the numerical simulations to validate the Gaussian property as shown in Fig. \ref{QQplot1}.}. 
However, due to the lack of explicit expressions of $f_{L}(\cdot)$ and $f_{S}(\cdot)$, it is generally difficult to determine the expression of $m_{\delta\rightarrow\X}(\X)$. From numerical simulations, we observe that $m_{\delta\rightarrow\X}(\X)$ is empirically quite far from a Gaussian distribution. As such, it is difficult to evaluate $m_{f_X\leftarrow\X}(\X)$ by combining $m_{\delta\rightarrow\X}(\X)$ and $m_{\y\rightarrow\X}(\X)$ as in (\ref{m_fX_X_mtx}). Instead, in the algorithm, we ignore $m_{\delta\rightarrow\X}(\X)$ by letting
\begin{align}
	m_{f_X\leftarrow \X}(\X) = m_{\y\rightarrow\X}(\X).\label{msg_fX_X_algo}
\end{align}
Indeed, this treatment may lose some information in message update since the message from node $\delta$ is ignored. However, this treatment facilitates the message update involved in factor node $f_X$. Particularly, when we ignore message $m_{\delta\rightarrow\X}(\X)$, the message passing from $f_X$ to $\X$ can be calculated using the turbo principle, as elaborated in what follows.

%--------------X and fx---------
\subsubsection{Message from factor node \texorpdfstring{$f_{X}$}{Lg} to variable node \texorpdfstring{$\X$}{Lg}}\label{derivation_dtmp_batch_2}

Similarly to (\ref{m_yx_xy}), define the message from factor node $f_{X}$ to variable node $\X$ by
\begin{align}\label{fX_X}
	m_{f_{X} \rightarrow \X}(\X)&=\prod_{i,j}m_{f_{X} \rightarrow X_{i,j}}(X_{i,j})
\end{align}
where the sum-product rule gives
\begin{align}
	m_{f_{X} \rightarrow X_{i,j}}(X_{i,j}) &=\frac{\int_{\X_{/(i,j)}} f_{X}(\X) m_{f_{X} \leftarrow \X}(\X) }{m_{f_{X} \leftarrow X_{i,j}}(X_{i,j})}\notag\\
	&= \frac{m_{f_{X}}(X_{i,j})}{m_{f_{X} \leftarrow X_{i,j}}(X_{i,j})} \label{msg_fxxp_xp}
\end{align}
with 
\begin{align}
	m_{f_{X}}(X_{i,j}) = \int_{\X_{/(i,j)}} f_{X}(\X) m_{f_{X} \leftarrow \X}(\X).\label{xij_post}
\end{align}
Further define 
\begin{align}
	m_{f_{X}}(\X) = \prod_{i,j}m_{f_{X}}(X_{i,j}).\label{msg_fX}
\end{align}
Then, from (\ref{fX_X})-(\ref{msg_fX}), we have
\begin{align}
	m_{f_{X}\rightarrow \X}(\X) = \frac{m_{f_{X}}(\X)}{m_{f_{X} \leftarrow \X}(\X)}.\label{x_ext}
\end{align}
Recall that an explicit expression of $f_X(\X)$ that encodes the image structural information of frames of video matrix $\X$ is unavailable. We next describe how to approximately evaluate the message in (\ref{x_ext}). Instead of calculating the integral in (\ref{xij_post}), we obtain the mean of message $m_{f_{X}}(\X)$ by using an image denoiser that handles the image structural information, such as the wavelet denoiser \cite{chang2000adaptive}, the sliding-window transform denoiser \cite{yaroslavsky2001transform}, or the block matching and 3D filtering (BM3D) denoiser \cite{dabov2006image}.
% \footnote{For example, the denoising process of BM3D denoiser can be described into three steps. Firstly, an image is divided into small blocks and similar blocks are stacked together to form a 3D array. The 3D array exhibits high correlation due to the similarity. Secondly, these 3D array are transformed using a unitary transform, noise are attenuated by shrinkage on the transformed domain. Lastly, inverse 3D transform yields estimates of all matched block.}
Denote by $\mathcal{D}_{x}(\x,v)$ an image denoiser that takes vector $\x$ as a noisy input with noise power $v$. Let $\mathcal{D}_{X}(\X,v)=[\mathcal{D}_{x}(\x_{1},v),\cdots,\mathcal{D}_{x}(\x_{n_2},v)]$. Then, the estimated mean of $m_{f_{X}}(\X)$ is given by
\begin{align}
	\X_{f_{X}} = D_{X}(\X_{f_X\leftarrow\X},v_{f_X\leftarrow\X}),\label{x_denoiser}
\end{align}
where $\X_{f_X\leftarrow\X}$ and $v_{f_X\leftarrow\X}$ are respectively the mean and variance of message $m_{f_X\leftarrow\X}(\X)$. From (\ref{ext_mean_var}a), the mean of message $m_{\y\rightarrow\X}(\X)$ is a linear combination of the mean matrices of messages $m_{\y}(\X)$ and $m_{\y\leftarrow\X}(\X)$. Similarly, we construct $\X_{f_{X}\rightarrow\X}$ as a linear combination of the mean matrices of messages $m_{f_{X}}$ and $m_{\X\rightarrow f_{X}}$:
\begin{align}
	\X_{f_{X} \rightarrow \X}=c_{X}(\mathcal{D}_{X}(\X_{f_X \leftarrow \X},v_{f_{X} \leftarrow \X})\!-\!\alpha_{X}\X_{f_{X} \leftarrow \X})\label{linearcb}
\end{align}
where $\alpha_{X}$ and $c_{X}$ are combination coefficients to be determined. The variance of message $m_{f_{X}\rightarrow\X}$ can be estimated by following \cite[Eq. 20]{xue2018turbo} as
\begin{align}
	v_{f_X\rightarrow\X} = & \frac{1}{m}\left\|\y-\mathcal{A}\left(\X_{f_{X}\rightarrow \X}\right)\right\|_F^2-\sigma^2.\label{linearcb_var}
\end{align}

We next determine the coefficients in (\ref{linearcb}). Note that $m_{f_X}(X_{i,j})$ and $m_{f_X\leftarrow X_{i,j}}(X_{i,j})$ can be respectively regarded as the posterior and prior messages of node $X_{i,j}$ in the processing of the constraint $f_X$. Then, (\ref{msg_fxxp_xp}) can be interpreted as the calculation of extrinsic message $m_{f_X\rightarrow X_{i,j}}(X_{i,j})$ by excluding the prior message from the posterior message. In other words, the extrinsic message $m_{f_X\rightarrow X_{i,j}}(X_{i,j})$ is independent of the prior message $m_{f_X \leftarrow X_{i,j}}(X_{i,j})$ (since the latter does not appear in the calculation of the former according to the sum-product rule). Independence implies uncorrelatedness, yielding
\begin{align}
	\left<\X_{f_{X} \leftarrow \X}-\X,\X_{f_{X}\rightarrow \X}-\X\right>=0,\label{cond1}
\end{align}
where $\left<\X_1,\X_2\right>=\text{trace}(\X_1^T\X_2)$. Further, we require that the output MSE is minimized. Then, $\alpha_x$ and $c_X$ can be determined by solving
\BS
\begin{align}
	\minimize_{\alpha_X, c_X}&\|\X_{f_{X}\rightarrow \X}-\X\|_F^2\\
	\st &\left<\X_{f_{X} \leftarrow \X}-\X,\X_{f_{X}\rightarrow \X}-\X\right>=0
\end{align}\label{cond2}\ES
where $\X_{f_X\rightarrow \X}$ is given by (\ref{linearcb}). By using Stein's unbiased risk estimate and Stein's lemma, it can be shown by following the approach of \cite{xue2017denoising} that the two coefficients in (\ref{linearcb}) are given by
\BS
\begin{align}
	\alpha_{X} &= \frac{\text{div}(\mathcal{D}_{X}(\X_{f_X \leftarrow \X},v_{f_X \leftarrow \X}))}{n}\\
	c_{X} &= \frac{\left<\mathcal{D}_X(\X_{f_X \leftarrow \X},v_{f_X \leftarrow \X})-\alpha_{X} \X_{f_X \leftarrow \X},\X_{f_X \leftarrow \X}\right>}{\|\mathcal{D}_{X}(\X_{f_X \leftarrow \X},v_{f_X \leftarrow \X})-\alpha_{X} \X_{f_X \leftarrow \X}\|_2^2}
\end{align}\label{xparams_c_a}\ES
where $\text{div}(\cdot)$ denotes the divergence. We note that a prerequisite to invoke Stein's lemma is that the input of the denoiser $m_{f_X\leftarrow\X}(\X)$ is a Gaussian message, or more specifically, the input mean $\X_{f_X\leftarrow\X}$ can be modelled as $\X_{f_X\leftarrow\X}=\X+\W$, where $\W$ is an additive Gaussian noise matrix with the elements drawn from the Gaussian distribution $\mathcal{N}(0,v_{f_X\leftarrow \X})$. From (\ref{msg_fX_X_algo}) and the discussions therein, the Gaussianity of message $m_{f_X\leftarrow \X}(\X)$ in (\ref{msg_fX_X_algo}) is ensured by the LMMSE estimation involved in the calculation of message $m_{\y\leftarrow\X}(\X)$.

%--------------delta and X---------
\subsubsection{Messages between variable node \texorpdfstring{$\X$}{Lg} and factor node \texorpdfstring{$\delta$}{Lg}} The message from $\X$ to $\delta$ is given by
\begin{align}
	m_{\delta \leftarrow  \X}(\X) =m_{f_{X} \rightarrow \X}(\X)m_{\y\rightarrow \X}(\X).
	\label{msg_x_delta}
\end{align}
Following \cite[Eq. 11]{ma2015turbo}, the mean and variance of message $m_{\delta\rightarrow X}(\X)$ are respectively given by
\BS
\begin{align}
\X_{\delta\leftarrow \X} &= v_{\delta\leftarrow \X}\left(\frac{\X_{f_{X} \rightarrow \X}}{v_{f_{X} \rightarrow \X}}+\frac{\X_{\y\rightarrow \X}}{v_{\y\rightarrow \X}}\right)\\
v_{\delta\leftarrow \X} &= \left( \frac{1}{v_{f_{X}\rightarrow \X}}+\frac{1}{v_{\y \rightarrow \X}}\right)^{-1}.
\end{align}\label{mv_m_delta_X}\ES

The message in the opposite direction is given by
\begin{align}
	m_{\delta\rightarrow\X}(\X)=\prod_{i,j} m_{\delta\rightarrow X_{i,j}}(X_{i,j})
\end{align}
where
\begin{align}\label{msg_f2_xp}
	m_{\delta \!\rightarrow\! X_{i,j}}(\!X_{i,j}\!)\!=\!\!\!\int_{L_{i,j},S_{i,j}}\!\!\!\!\!\!\!\!\!\!\!\!\!\!\!\! \delta(X_{i,j}\!-\!L_{i,j}\!-\!S_{i,j}) m_{\delta\!\leftarrow \!L_{i,j}}\!(L_{i,j})m_{\delta\! \leftarrow \! S_{i,j}}\!(S_{i,j}).
\end{align}
Denote by $\L_{\delta\leftarrow\L}$ and $\S_{\delta\leftarrow \S}$ the means of messages $m_{\delta\leftarrow\L}(\L)$ and $m_{\delta\leftarrow\S}(\S)$ respectively, and by $v_{\delta\leftarrow\L}$ and $v_{\delta\leftarrow\S}$ the variances of messages $m_{\L \rightarrow \delta}(\L)$ and $m_{\S \rightarrow \delta}(\S)$ respectively. Then, the mean and variance of message $m_{\delta\rightarrow\X}(\X)$ are respectively given by
\BS 
\begin{align}
	\X_{\delta\rightarrow\X} &= \L_{\delta\leftarrow\L}+\S_{\delta\leftarrow\S}\\
	v_{\delta\rightarrow\X} &= v_{\delta\leftarrow\L}+v_{\delta\leftarrow\S}.
\end{align}\label{msg_delta_X_m_v}\ES

%--------------delta to L---------
\subsubsection{Messages between factor node \texorpdfstring{$\delta$}{Lg} and variable node \texorpdfstring{$\L$}{Lg}} From the sum-product rule, we have
\BS\label{msg_delta_L}
\begin{align}
	m_{\delta\leftarrow\L}(\L) &= m_{f_{L} \rightarrow \L}(\L)\\
	m_{\delta\rightarrow\L}(\L)&=\prod_{i,j}m_{\delta \rightarrow L_{i,j}}(L_{i,j})
\end{align}\ES
where
\begin{align}\label{msg_delta_l}
	m_{\delta \!\rightarrow\! L_{i,j}}(L_{i,j}\!)\!=\!\!\!\int_{X_{i,j},S_{i,j}}\!\!\!\!\!\!\!\!\!\!\!\!\!\!\!\!\!\! \delta(X_{i,j}\!\!-\!\!L_{i,j}\!\!-\!\!S_{i,j}) m_{\delta\!\leftarrow\! X_{i,j}}(X_{i,j}) m_{\delta\!\leftarrow\! S_{i,j}}\!(\!S_{i,j}\!).
\end{align}
From (\ref{msg_delta_l}), the mean and variance of $m_{\delta \rightarrow \L}(\L)$ are given by 
\BS
\begin{align}
	\L_{\delta \rightarrow \L} &= \X_{\delta\leftarrow\X}-\S_{\delta\leftarrow\S}\\
	v_{\delta \rightarrow \L} &= v_{\delta\leftarrow\X}+v_{\delta\leftarrow\S}.
\end{align}\label{msg_m_v_delta_L}\ES

%--------------delta to S---------
\subsubsection{Messages between factor node \texorpdfstring{$\delta$}{Lg} and variable node \texorpdfstring{$\S$}{Lg}} Similarly to (\ref{msg_delta_L}), we have
\BS
\begin{align}
	m_{ \delta\leftarrow\S}(\S) &= m_{f_{S} \rightarrow \S}(\S)\\
	m_{\delta\rightarrow\S}(\S)&=\prod_{i,j}m_{\delta \rightarrow S_{i,j}}(S_{i,j})
\end{align}\label{ms_delta_S}\ES
where
\begin{align}\label{msg_delta_S}
	m_{\delta\! \rightarrow \!S_{i,j}}(S_{i,j}\!)\!=\!\!\int_{X_{i,j},L_{i,j}}\!\!\!\!\!\!\!\!\!\!\!\!\!\!\!\!\!\! \delta(X_{i,j}\!\!-\!\!L_{i,j}\!\!-\!\!S_{i,j}) m_{\delta\!\leftarrow\! X_{i,j}}(\!X_{i,j}\!)m_{\delta\!\leftarrow\! L_{i,j}}(\!L_{i,j}\!).
\end{align}
From (\ref{msg_delta_S}), the mean and variance of $m_{\delta \rightarrow \S}(\S)$ are respectively given by 
\BS
\begin{align}
	\S_{\delta \rightarrow \S} &= \X_{\delta\leftarrow\X}-\L_{\delta\leftarrow\L}\\
	v_{\delta \rightarrow \S} &= v_{\delta\leftarrow\X}+v_{\delta\leftarrow\L}.
\end{align}\label{m_v_delta_S}\ES

%-------------fL and L---------
\subsubsection{Messages between variable node \texorpdfstring{$\L$}{Lg} and factor node \texorpdfstring{$f_{L}$}{Lg}} From the sum-product rule, the message from $\L$ to $f_{L}(\L)$ is given by
\begin{align}\label{msg_L_fL}
	m_{f_{L}\leftarrow\L}(\L) = m_{\delta \rightarrow \L}(\L),
\end{align}
and the message from $f_{L}(\L)$ to $\L$ is given by
\begin{align}
	m_{f_L\rightarrow\L}(\L)=\prod_{i,j}m_{f_{L} \rightarrow L_{i,j}}(L_{i,j})
\end{align}
where
\BS
\begin{align}
	m_{f_{L} \rightarrow L_{i,j}}(L_{i,j})&= \frac{\int_{\L_{/(i,j)}} m_{f_{L}\leftarrow \L}(\L) f_{L}(\L)}{m_{f_{L}\leftarrow L_{i,j}}(L_{i,j})}\\
	&=\frac{m_{f_{L}}(L_{i,j})}{m_{f_{L}\leftarrow L_{i,j}}(L_{i,j})}.
\end{align}\label{m_fL_L_ij}\ES
Note that $m_{f_L\rightarrow L_{i,j}}(L_{i,j})$ in (\ref{m_fL_L_ij}) is difficult to determine since $f_{L}(\L)$ is not available. We next calculate the mean and variance of $m_{f_L\rightarrow L_{i,j}}(L_{i,j})$ by mimicking the approach to the approximation of $m_{f_X\rightarrow\X}(\X)$.

Denote by $\mathcal{D}_{L}(\cdot,\cdot)$ the denoiser for low-rank matrix estimation. When the rank of $\L$ is not available, $\mathcal{D}_{L}(\cdot,\cdot)$ can be chosen as the singular value soft thresholding (SVST) denosier or the singular value hard thresholding (SVHT) denoiser \cite{cai2010singular}; when the rank of $\L$ is known, $\mathcal{D}_{L}(\cdot,\cdot)$ can be chosen as the best-rank-$r$ denoiser \cite{eckart1936approximation}. The estimated mean of message $m_{f_{L} \rightarrow \L}(\L)$ is given by
\begin{align}\label{m_fL_L}
	\L_{f_{L} \rightarrow \L}&=c_{L}(\mathcal{D}_{L}(\L_{f_{L} \leftarrow \L},v_{f_{L} \leftarrow \L})-\alpha_{L}\L_{f_{L} \leftarrow \L})
\end{align}
where coefficients $c_{L}$ and $\alpha_{L}$ are given by
\BS
\begin{align}
	\alpha_{L} &= \frac{\text{div}(\mathcal{D}_{L}(\L_{f_{L} \leftarrow \L},v_{f_{L} \leftarrow \L}))}{n},\\
	c_{L} &= \frac{\left<\mathcal{D}_{L}(\L_{f_{L} \leftarrow \L},v_{f_{L} \leftarrow \L})-\alpha_{L} \L_{f_{L} \leftarrow \L},\L_{f_{L} \leftarrow \L}\right>}{\|\mathcal{D}_{L}(\L_{f_{L} \leftarrow \L},v_{f_{L} \leftarrow \L})-\alpha_{L} \L_{f_{L} \leftarrow \L}\|_2^2}.
\end{align}\label{lparams_c_a}\ES

When $\D_{L}(\cdot,\cdot)$ is chosen as the best-rank-$r$ denoiser, the variance of message $m_{f_{L} \rightarrow \L}(\L)$ can be estimated by \cite[Eq. 54(a)]{xue2019tarm}
\begin{align}\label{v_fL_L}
	v_{f_{L} \rightarrow \L} = v_{f_{L} \leftarrow \L}\left(\left(1-\frac{r}{n_1}\left(1+\frac{k}{n_1}\right)\right)\frac{1}{(1-\alpha_L)^2}-1\right).
\end{align}
Alternatively, the variance can be estimated by \cite[Eq. 54(c)]{xue2019tarm}
\begin{align}
	v_{f_{L} \!\rightarrow\! \L} =& -\frac{\left<\mathcal{D}_{L}(\L_{f_{L} \leftarrow \L},v_{f_{L} \leftarrow \L})-\alpha_L \L_{f_{L} \leftarrow \L},\L_{f_{L} \leftarrow \L}\right>^2}{n\|\mathcal{D}_{L}(\L_{f_{L} \leftarrow \L},v_{f_{L} \leftarrow \L})-\alpha_L \L_{f_{L} \leftarrow \L}\|_F^2}\notag\\
	&+\frac{\|\L_{f_{L} \leftarrow \L}\|_F^2}{n}-v_{f_{L} \leftarrow \L}.
\end{align}

% ======S=====f_S===========
\subsubsection{Messages between variable node \texorpdfstring{$\S$}{Lg} and factor node \texorpdfstring{$f_{S}$}{Lg}} The message from $\S$ to $f_{S}$ is given by
\BS
\begin{align}
	m_{f_S \leftarrow \S}(\S) &= m_{\delta \rightarrow \S}(\S)\\
	m_{f_S\rightarrow\S}(\S)&=\prod_{i,j}m_{f_{S} \rightarrow \S}(S_{i,j})
\end{align}\label{msg_S_fS}\ES
where
\BS
\begin{align}
	m_{f_{S} \rightarrow S_{i,j}}(S_{i,j})&= \frac{\int_{\S_{/i,j}} m_{\S\rightarrow f_{S}}(\S) f_{S}(\S) }{m_{f_{S}\leftarrow S_{i,j}}(S_{i,j})}\\
	&= \frac{m_{f_{S}}(S_{i,j})}{m_{f_{S}\leftarrow S_{i,j}}(S_{i,j})}.
\end{align}\label{m_fS_S_ij}\ES
Note that $m_{f_S\rightarrow S_{i,j}}(S_{i,j})$ in (\ref{m_fS_S_ij}) is difficult to determine since $f_{S}(\S)$ is not available. Following the treatment for the low-rank denoiser described above, we calculate the mean and variance of $m_{f_S\rightarrow S_{i,j}}(S_{i,j})$ as follows.

Denote by $\mathcal{D}_{S}(\cdot,\cdot)$ the denoiser for sparse matrix estimation. Denoisers for general sparse data include soft-thresholding \cite{donoho1995noising} and Stein's unbaised risk linear expansion of thresholds (SURE-LET) estimators \cite{blu2007sure}. The estimated mean of message $m_{f_{S} \rightarrow \S}(\S)$ is given by
\begin{align}\label{m_fS_S}
	\S_{f_{S} \rightarrow \S}&=c_{S}(\mathcal{D}_{S}(\S_{f_{S} \leftarrow \S},v_{f_{S} \leftarrow \S})-\alpha_{S}\S_{f_{S} \leftarrow \S})
\end{align}
where $c_{S}$ and $\alpha_{S}$ are linear combination coefficients given by
\BS
\begin{align}
	\alpha_{S} &= \frac{\text{div}(\mathcal{D}_{S}(\S_{f_{S} \leftarrow \S},v_{f_{S} \leftarrow \S}))}{n},\\
	c_{S} &= \frac{\left<\mathcal{D}_{S}(\S_{f_{S} \leftarrow \S},v_{f_{S} \leftarrow \S})-\alpha_{S} \S_{f_{S} \leftarrow \S},\S_{f_{S} \leftarrow \S}\right>}{\|\mathcal{D}_{S}(\S_{f_{S} \leftarrow \S},v_{f_{S} \leftarrow \S})-\alpha_{S} \S_{f_{S} \leftarrow \S}\|_2^2}.
\end{align}\label{sparams_c_a}\ES

For any sparsity denoiser $\mathcal{D}_{S}(\cdot,\cdot)$, the variance of message $m_{f_{S} \rightarrow \S}(\S)$ can be estimated by \cite{xue2019tarm}
\begin{align}\label{v_fS_S}
	v_{f_{S} \rightarrow \S}= & \frac{1}{m}\|\y\!-\!\mathcal{A}(\S_{f_{S} \rightarrow \S}\!+\!\L_{\delta\leftarrow \L}))\|_2^2\!-\!v_{\delta\leftarrow \L}\!-\!\sigma^2.
\end{align}

%------------ X to y 
\subsubsection{Message from variable node \texorpdfstring{$\X$}{Lg} to factor node \texorpdfstring{$\y$}{Lg}}From the sum-product rule, we have
\begin{align}
	m_{\y \leftarrow \X}(\X) =m_{f_{X} \rightarrow \X}(\X)m_{\delta\rightarrow \X}(\X).
	\label{msg_x_y}
\end{align}
Similarly to (\ref{mv_m_delta_X}), the mean and variance of message $m_{\y\leftarrow \X}(\X)$ are respectively given by
\BS
\begin{align}
\X_{\y \leftarrow \X} &= v_{\y \leftarrow \X}\left(\frac{\X_{f_{X} \rightarrow \X}}{v_{f_{X} \rightarrow \X}}+\frac{\X_{\delta\rightarrow \X}}{v_{\delta\rightarrow \X}}\right)\\
v_{\y \leftarrow \X} &= \left(\frac{1}{v_{f_{X} \rightarrow \X}}+\frac{1}{v_{\delta\rightarrow \X}}\right)^{-1}
\end{align}\label{mv_y_x}\ES
It is interesting to note that here we take a different approach from the threatment of $m_{f_X\leftarrow \X}(\X)$ in (\ref{m_fX_X_mtx}). In specific, the approximation of $m_{f_X\leftarrow \X}(\X)$ in (\ref{msg_fX_X_algo}) ignores the non-Gaussian message component, so as to ensure that the approximated $m_{f_X\leftarrow \X}(\X)$ is close to Gaussian and hence the Stein's lemma becomes applicable in determining the coefficients $\alpha_X$ and $c_X$ in (\ref{xparams_c_a}). However, $m_{\y\leftarrow\X}(\X)$ here is not necessarily to be Gaussian. From (\ref{linear_mean_var}) and the discussion therein, the LMMSE estimation involved in the factor node $p(\y|\X)$ only requires the mean and variance of the input message $m_{\y\leftarrow \X}(\X)$. Therefore, both message components are kept in the message update (\ref{msg_x_y}).

\subsection{Overall Algorithm} 
Based on the discussions in the preceding subsections, we summarize the offline denoising-based turbo message passing (DTMP) algorithm in Algorithm \ref{batch_dtmp}. Note that a partial orthogonal linear operator $\mathcal{A}$ is employed in Algorithm \ref{batch_dtmp}. Specifically, Lines 2-3 of Algorithm \ref{batch_dtmp} correspond to the mean and variance of message $m_{\y\rightarrow\X}(\X)$ in (\ref{m_v_m_y_X}) together with the equality in (\ref{msg_fX_X_algo}). Lines 4-5 of Algorithm \ref{batch_dtmp} correspond to the mean and variance of message $m_{f_X\rightarrow\X}(\X)$ in (\ref{linearcb}) and (\ref{linearcb_var}). Lines 6-7 follow from (\ref{mv_m_delta_X}). Lines 8-9 correspond to the mean and variance of messages $m_{\delta\rightarrow\L}(\L)$ in (\ref{msg_m_v_delta_L}) and the equality in (\ref{msg_L_fL}). Lines 10-11 correspond to the mean and variance of message $m_{f_L \rightarrow \L}(\L)$ in (\ref{m_fL_L}) and (\ref{v_fL_L}) and the equality in (\ref{msg_delta_L}a). Lines 12-13 correspond to the mean and variance of messages $m_{\delta\rightarrow\S}(\S)$ in (\ref{m_v_delta_S}) and the equality in (\ref{msg_S_fS}). Lines 14-15 corresponds to the mean and variance of message $m_{f_S\rightarrow\S}(\S)$ in (\ref{m_fS_S}) and (\ref{v_fS_S}) and the equality in (\ref{ms_delta_S}a). Lines 16-17 correspond to the mean and variance of message $m_{\delta\rightarrow\X}(\X)$ in (\ref{msg_delta_X_m_v}). Lines 18-19 correspond to the mean and variance of message $m_{\y\leftarrow\X}(\X)$ in (\ref{mv_y_x}).

% Compared to the TMP algorithm \cite{xue2018turbo}, extra processing steps that exploit the image structural information are introduced in the offline DTMP algorithm (Lines 4-7 and Lines 16-19 of Algorithm \ref{batch_dtmp}) to improve the recover quality of video frames. An image denoiser is employed in Line 4 of Algorithm \ref{batch_dtmp}. To ensure good performance of the offline DTMP algorithm, we carefully design the input of the image denoiser (by dropping a non-Gaussian input component), based on which the extrinsic message passing principle can be applied (parameters $\alpha_X$ and $c_X$ in Line 4).

In offline DTMP, we need to initialize $\X_{\y\leftarrow\X}$, $v_{\y\leftarrow\X}$, $\S_{\delta\leftarrow\S}$ and $v_{\delta\leftarrow\S}$ at the beginning of the algorithm. Ideally, we shall choose the mean and variance of $\X$ as the initial $\X_{\y\leftarrow\X}$ and $v_{\y\leftarrow\X}$, and the mean and variance of $\S$ as the initial $\S_{\delta\leftarrow\S}$ and $v_{\delta\leftarrow\S}$. However, these statistics are usually unavailable in practice. From simulations, we observe that the offline DTMP algorithm is not very sensitive to the initialization of these parameters. Thus, we initialize these parameters by simply setting
\BS
\begin{align}
	\X_{\y\leftarrow\X} &= \S_{\delta\leftarrow\S}=\bm{0},\\
	v_{\y\leftarrow\X} &= v_{\delta\leftarrow\S} = \frac{\|\y\|_2^2}{m}.
\end{align}\label{off_init_x_v}\ES
In addition, the noise power $\sigma^2$ is assumed to be known in prior. In practice, $\sigma^2$ can be estimated by following the standard expectation maximization method.

We now briefly discuss the computational complexity of offline DTMP. The complexity of offline DTMP is dominated by the operations in Lines 2, 4, 10 and 14. Specifically, Line 2 is for the LMMSE estimation of $\X$. The computation complexity of the matrix multiplication operations in Line 2 is $\mathcal{O}(m n_1 n_2)$ flops for a general linear operator $\mathcal{A}$. If we choose a partial orthogonal linear operator such as the partial Discrete Cosine Transform (DCT) operator constructed by randomly selected rows from the DCT matrix of size $n\times n$, the complexity reduces to $\mathcal{O}(n_1n_2 \log (n_1 n_2))$ flops by using the fast cosine transform. The complexity of Line 4 is dominated by the operations of denoiser $\mathcal{D}_{X}$. For frequently used image denoisers, the complexity is typically linear to the image size, i.e. $\mathcal{O}(n_1 n_2)$ flops. The complexity of Line 10 is dominated by the operations of $\mathcal{D}_{L}(\L_{f_L\leftarrow\L},v_{f_L\leftarrow\L})$ which involves the computation of the truncated SVD. The complexity of the truncated SVD is $\mathcal{O}(r n_1 n_2)$ flops. The complexity of Line 14 is dominated by the calculation of $\mathcal{D}_{S}(\S_{f_S\leftarrow\S},v_{f_S\leftarrow\S})$. For the SURE-LET denoiser with the kernel chosen as \cite[Eqs. 28-30]{xue2017denoising}, the complexity is $\mathcal{O}(n_1 n_2)$ flops. Therefore the overall complexity of the offline DTMP algorithm per iteration is given by $\mathcal{O}(r n_1 n_2)+\mathcal{O}(n_1n_2 \log (n_1 n_2))$ flops.

\begin{algorithm}
\caption{Offline DTMP}\label{batch_dtmp}
\begin{algorithmic}[1]
\REQUIRE $\mathcal{A}, \y, \X_{\y\leftarrow\X}, v_{\y\leftarrow\X}, \S_{\delta\leftarrow\S}, v_{\delta\leftarrow\S}, \sigma^2$
\\
\WHILE{the stopping criterion is not met}

\STATE $\X_{f_X\leftarrow\X} =\X_{\y\rightarrow\X} = \X_{\y\leftarrow\X} + \frac{n_1 n_2}{m} \mathcal{A}^T(\y-\mathcal{A}(\X_{\y\leftarrow\X}))$
\STATE $v_{f_X\leftarrow\X} = v_{\y\rightarrow\X} = \frac{n_1 n_2}{m}(v_{\y\leftarrow\X}+\sigma^2)-v_{\y\leftarrow\X}$

\STATE	$\X_{f_{X} \rightarrow \X}=c_{X}(\mathcal{D}_{X}(\X_{f_X \leftarrow \X},v_{f_{X} \leftarrow \X})-\alpha_{X}\X_{f_{X} \leftarrow \X})$ with $\alpha_X$ and $c_{X}$ in (\ref{xparams_c_a})
\STATE $v_{f_X\rightarrow\X} = \frac{1}{m}\left\|\y-\mathcal{A}\left(\X_{f_{X}\rightarrow \X}\right)\right\|_F^2-\sigma^2$

\STATE $v_{\delta\leftarrow\X} = \left(\frac{1}{v_{f_X\rightarrow\X}}+\frac{1}{v_{\y\rightarrow\X}}\right)^{-1}$
\STATE $\X_{\delta\leftarrow\X} = v_{\delta\leftarrow\X}\left(\frac{\X_{f_{X} \rightarrow \X}}{v_{f_{X} \rightarrow \X}}+\frac{\X_{\y\rightarrow\X}}{v_{\y\rightarrow\X}}\right)$

\STATE $\L_{f_{L}\leftarrow\L}=\L_{\delta \rightarrow \L} = \X_{\delta\leftarrow\X}-\S_{\delta\leftarrow\S}$
\STATE $v_{f_{L}\leftarrow\L}=v_{\delta \rightarrow \L} = v_{\delta\leftarrow\X}+v_{\delta\leftarrow\S}$

\STATE	$\L_{\delta\leftarrow\L}=\L_{f_{L} \rightarrow \L}=c_{L}(\mathcal{D}_{L}(\L_{f_{L} \leftarrow \L},v_{f_{L} \leftarrow \L})-\alpha_{L}\L_{f_{L} \leftarrow \L})$ with $\alpha_L$ and $c_{L}$ in (\ref{lparams_c_a})
\STATE $v_{\delta\leftarrow\L}= v_{f_{L} \leftarrow \L}\left(\left(1-\frac{r}{n_1}\left(1+\frac{k}{n_1}\right)\right)\frac{1}{(1-\alpha_L)^2}-1\right)=v_{f_{L} \rightarrow \L} $
\STATE $\S_{f_S \leftarrow \S}=\S_{\delta \rightarrow \S} = \X_{\delta\leftarrow\X}-\L_{\delta\leftarrow\L}$
\STATE	$v_{f_S \leftarrow \S}=v_{\delta \rightarrow \S} = v_{\delta\leftarrow\X}+v_{\delta\leftarrow\L}$
\STATE	$\S_{\delta\leftarrow\S}=\S_{f_{S} \rightarrow \S}=c_{S}(\mathcal{D}_{S}(\S_{f_{S} \leftarrow \S},v_{f_{S} \leftarrow \S})-\alpha_{S}\S_{f_{S} \leftarrow \S})$ with $\alpha_S$ and $c_{S}$ in (\ref{sparams_c_a})
\STATE $v_{\delta\leftarrow\S}=v_{f_{S} \rightarrow \S}= \frac{1}{m}\|\y-\mathcal{A}(\S_{f_{S} \rightarrow \S}+\L_{\delta\leftarrow \L}))\|_2^2-v_{\delta\leftarrow \L}-\sigma^2$

\STATE $\X_{\delta\rightarrow\X} = \L_{\delta\leftarrow\L}+\S_{\delta\leftarrow\S}$
\STATE $v_{\delta\rightarrow\X} = v_{\delta\leftarrow\L}+v_{\delta\leftarrow\S}$
\STATE $v_{\y\leftarrow\X} = \left(\frac{1}{v_{\delta\rightarrow\X}}+\frac{1}{v_{f_X\rightarrow\X}}\right)^{-1}$
\STATE $\X_{\y\leftarrow\X} = v_{\y\leftarrow\X}\left(\frac{\X_{f_{X} \rightarrow \X}}{v_{f_{X} \rightarrow \X}}+\frac{\X_{\delta\rightarrow\X}}{v_{\delta\rightarrow\X}}\right)$
\ENDWHILE
\ENSURE  $\hat{\X}=\hat{\L}+\hat{\S}, \hat{\L}=\mathcal{D}_{L}(\L_{f_{L} \leftarrow \L},v_{f_{L} \leftarrow \L}),\hat{\S}=\mathcal{D}_{S}(\S_{f_{S} \leftarrow \S},v_{f_{S} \leftarrow \S})$
\end{algorithmic}
\end{algorithm}

% /////////////////////////////////////////
\section{Online Denoising-Based Turbo Message Passing}\label{sec_online_dtmp}

\subsection{Problem Formulation}
As discussed in the preceding section, the complexity of the offline DTMP algorithm is linear in the total number of video frames. This complexity is not affordable for online processing of video data, where real-time foreground background separation is required. Specifically, the online version of the compressed VBS problem is defined as to recover the background component $\l_t $ and foreground component $\s_t$ of the current frame $\x_t$ given the current compressed measurement $\y_t$ and the separation results from the previous frames: $\S^{pre}_{t}=[\hat{\s}_{t-k_{S}}, \cdots, \hat{\s}_{t-1}]$ and $\L^{pre}_{t}=[\hat{\l}_{t-k_{L}}, \cdots, \hat{\l}_{t-1}]$, where $k_{S}$ and $k_{L}$ are respectively the window sizes of the recovered foregrounds and backgrounds, and $\hat{\s}_{i}$ and $\hat{\l}_{i}$ are respectively the recovered foreground and background of the $i$-th frame. In addition, each $\y_i$ is modelled by
\begin{align}
	\y_i = \mathcal{A}_{i}(\x_i)+\n_i=\mathcal{A}_{i}(\l_i+\s_i)+\n_i, i=1,\cdots,t \label{on_com_vbs}
\end{align}
where $\n_i\in \mathbb{R}^{m\times 1}$ is the Gaussian measurement noise with mean zero and covariance matrix $\sigma^2 \I$, and $\mathcal{A}_{i}: \mathbb{R}^{hw\times 1}\rightarrow \mathbb{R}^{m\times 1}$ is a linear measurement operator applied to the $i$-th frame.

We aim to extend the offline DTMP algorithm to the online scenario. There are three issues to be addressed in the extension: 
\begin{enumerate}
	\item First, in the online scenario, the linear measurement is conducted in a frame-by-frame manner rather than after collecting all the frames. That is, the linear estimation constraints in (\ref{off_com_vbs}) need to be decoupled as in (\ref{on_com_vbs}).
	\item Second, in the offline scenario, the background frames together are treated as a low-rank matrix, and the singular value decomposition (SVD) is applied in the estimation of the low-rank matrix. However, in the online case, the video frames arrive in a one-by-one manner. As such, it will be very time-consuming if the SVD operation is applied to the whole set of the collected frames whenever a frame arrives.
	\item Third, in offline DTMP, each foreground frame is estimated separately, and the continuity of the foregrounds in adjacent frames is not considered in the algorithm design. Such continuity, if appropriately exploited, can improve the performance of VBS.
\end{enumerate}
To address the above issues, we start with the probability model of the online problem in what follows.

\begin{figure}[!ht]
	\centering
	\includegraphics[width=\linewidth]{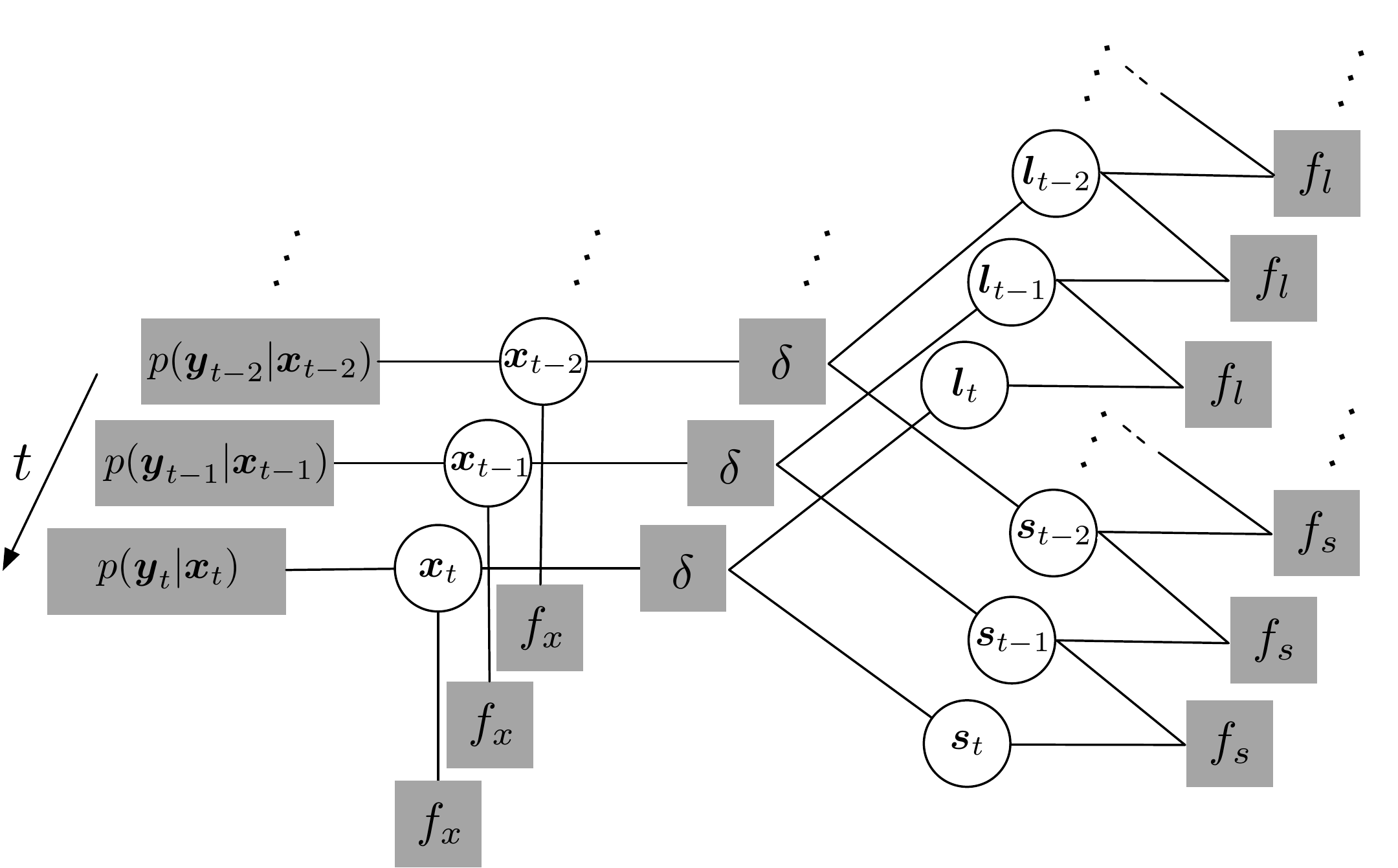}
	\caption{The factor graph of the joint posterior distribution of $\x_t$, $\l_t$ and $\s_t$, where $k_s=k_l=2$.}
	\label{onlinefg}
\end{figure}

\subsection{Factor Graph Representation}
For the online VBS problem, the joint probability density of $\y_{t}$, $\x_t$, $\l_t$, and $\s_t$ conditioned on $\L^{pre}_{t}$ and $\S^{pre}_{t}$ is given by
\begin{align}
	p(\y_t,\x_t,\l_t,\s_t|\L^{pre}_{t},\S^{pre}_{t}) = &\,\, p(\y|\x_t)p(\x_t|\l_t,\s_t) \notag\\
	&\times p(\l_t|\L^{pre}_{t})p(\s_t|\S^{pre}_{t})\label{on_promodel}
\end{align}
where $p(\l_t|\L^{pre}_{t})$ and $p(\s_t|\S^{pre}_{t})$ are respectively the probability densities of the background and the foreground of the $t$-th frame conditioned on the previously recovered results, and 
\BS
\begin{align}
	p(\y_t|\x_t) &= \mathcal{N}(\y_t;\mathcal{A}_{i}(\x_t),\sigma^2\I)\\
	p(\x_t|\s_t,\l_t) &= \delta(\x_t -\s_t-\l_t).
\end{align}\ES
The conditional probability densities $p(\l_t|\L^{pre}_{t})$ and $p(\s_t|\S^{pre}_{t})$ are difficult to acquire in practice. Similarly to the approach in (\ref{joint_prob}), we avoid this difficulty by approximating the joint probability (\ref{on_promodel}) as 
\begin{align}
	\ p(\y_t,\x_t,\l_t,\s_t|\L^{pre}_{t},\S^{pre}_{t})\approx &\,\mathcal{N}(\y_t;\mathcal{A}_{i}(\x_t),\sigma^2\I)\delta(\x_t\!-\!\l_t\!-\!\s_t) \notag\\  &\times\!f_{x}(\x_t)f_{l}(\l_t;\L^{pre}_{t})f_{s}(\s_t;\S^{pre}_{t}),
	\label{appro_joint_online_prob}
\end{align}
where $f_{x}(\x_t)$ represents the constraint of $\x_t$ as a natural image, $f_{l}(\l_t;\L_t^{pre})$ represents the correlation between $\l_t$ and the previous backgrounds $\L_t^{pre}$, and $f_s(\s_t;\S_t^{pre})$ represents the correlation between $\s_t$ and the previous foregrounds $\S_t^{pre}$.

The factor graph of the joint probability distribution in (\ref{appro_joint_online_prob}) is given in Fig. \ref{online_dtmp}. The online DTMP algorithm is established based on the factor graph in Fig. \ref{online_dtmp}, as detailed in the next subsection. We should note that in our prior work \cite{xue2018turbo}, the proposed TMP algorithm can only be used in offline cases. Thus, the extension of offline DTMP to address the online problem is our contribution comparing with the work in \cite{xue2018turbo}.

\subsection{Online DTMP Algorithm}
The online DTMP algorithm is presented in Algorithm  \ref{online_dtmp}. The derivation of the online DTMP algorithm is mostly similar to that of the offline DTMP algorithm. Specifically, Lines 2 and 3 correspond to the means and variances of message $m_{\y_t\rightarrow\x_t}(\x_t)$ and $m_{f_x\leftarrow\x_t}(\x_t)$. In fact, these two lines are simply the vector version of Lines 2 and 3 of Algorithm 1 by noting that (\ref{on_com_vbs}) is a special case of (\ref{off_com_vbs}) by letting $\mathcal{A}(\X)=[\mathcal{A}_{1}(\x_1)^T,\cdots,\mathcal{A}_{n_2}(\x_{n_2})^T]^T$. Lines 4 and 5 correspond to the mean and variance of message $m_{f_x\rightarrow\x_t}(\x_t)$ where $\mathcal{D}_{x}(\x,v)$ is an image denoiser with noisy input $\x$ and noise power $v$. These two lines are similar to Lines 4 and 5 of Algorithm \ref{batch_dtmp} by noting $\mathcal{D}_{X}(\X,v)=[\mathcal{D}_{x}(\x_{1},v),\cdots,\mathcal{D}_{x}(\x_{n_2},v)]$. Lines 6 and 7 correspond to the mean and variance of message $m_{\delta\leftarrow \x_t}(\x_t)$ that can obtained by following (\ref{mv_m_delta_X}). Lines 8 and 9 correspond to the means and variances of messages $m_{\delta\rightarrow\l_t}(\l_t)$ and $m_{f_l\leftarrow\l_t}(\l_t)$ derived by following (\ref{msg_m_v_delta_L}). Lines 10 and 11 correspond to the means and variances of messages $m_{f_l \rightarrow \l_t}(\l_t)$ and $m_{\delta\leftarrow\l_t}(\l_t)$, where $\mathcal{D}_{l}(\l_{f_{l} \leftarrow \l_t};\L_t^{pre},r)\in \mathbb{R}^{hw\times 1}$ is a low-rank denoiser, and coefficients $\alpha_l$ and $c_l$ can be calculated similarly to (\ref{lparams_c_a}) as
	\BS
	\begin{align}
		\alpha_{l} &= \frac{\text{div}(\mathcal{D}_{l}(\l_{f_{l} \leftarrow \l_t},\L_t^{pre},r))}{n},\\
		c_{l} &= \frac{\left<\mathcal{D}_{l}(\l_{f_{l} \leftarrow \l_t},\L_t^{pre},r)-\alpha_{l} \l_{f_{l} \leftarrow \l_t},\l_{f_{l} \leftarrow \l_t}\right>}{\|\mathcal{D}_{l}(\l_{f_{l} \leftarrow \l_t},\L_t^{pre},r)-\alpha_{l} \l_{f_{l} \leftarrow \l_t}\|_2^2}.
	\end{align}\label{on_params_l}\ES
	Lines 12 and 13 correspond to the means and variances of messages $m_{\delta\rightarrow\s_t}(\s_t)$ and $m_{f_s\leftarrow\s_t}(\s_t)$ that can be obtained by following (\ref{m_v_delta_S}). Lines 14 and 15 correspond to the means and variances of messages $m_{f_s\rightarrow\s_t}(\s_t)$ and $m_{\delta\leftarrow\s_t}(\s_t)$ where $\mathcal{D}_{s}(\s_{f_s\leftarrow \s_t},v_{f_s\leftarrow \s_t};\S_t^{pre})$ is a foreground estimator by taking noisy foreground $\s_{f_s\leftarrow \s_t}$, noise power $v_{f_s\leftarrow \s_t}$, and previously recovered foregrounds $\S_t^{pre}$ as input. Similarly to (\ref{on_params_l}), the coefficients $\alpha_s$ and $c_s$ are calculated by
	\BS
	\begin{align}
		\alpha_{s} &= \frac{\text{div}(\mathcal{D}_{s}(\s_{f_{s} \leftarrow \s},v_{f_{s} \leftarrow \s};\S_t^{pre}))}{n},\\
		c_{s} &= \frac{\left<\mathcal{D}_{s}(\s_{f_{s} \leftarrow \s},v_{f_{s} \leftarrow \s};\S_t^{pre})-\alpha_{s} \s_{f_{s} \leftarrow \s},\s_{f_{s} \leftarrow \s}\right>}{\|\mathcal{D}_{s}(\s_{f_{s} \leftarrow \s},v_{f_{s} \leftarrow \s};\S_t^{pre})-\alpha_{s} \s_{f_{s} \leftarrow \s}\|_2^2}.
	\end{align}\label{on_params_s}\ES
	Lines 16 and 17 correspond to the mean and variance of message $m_{\delta\rightarrow\x_t}(\x_t)$ obtained by following (\ref{msg_delta_X_m_v}). Finally, Lines 18 and 19 correspond to the means and variances of messages $m_{\y_t\leftarrow\x_t}(\x_t)$ follows from similar derivations in (\ref{mv_y_x}) and discussions therein. In Algorithm \ref{online_dtmp}, we need to initialize $\x_{\y_t\leftarrow\x_t}$, $v_{\y_t\leftarrow\x_t}$, $\s_{\delta\leftarrow\s_t}$, and $v_{\delta\leftarrow\s_t}$. Similarly to the initialization of offline DTMP in (\ref{off_init_x_v}), we set
\BS
\begin{align}
	\x_{\y_t\leftarrow\x_t} &= \s_{\delta\leftarrow\s_t}=\bm{0},\\
	v_{\y_t\leftarrow\x_t} &= v_{\delta\leftarrow\s_t} = \frac{\|\y_t\|_2^2}{m}.
\end{align}\ES

Compared with the offline version, the main difference of the above online DTMP resides in the realizations of the denoisers $\mathcal{D}_{l}$ and $\mathcal{D}_{s}$. We first describe the realization of the low-rank denoiser $\mathcal{D}_{l}$. Note that $\mathcal{D}_{l}$ takes $\{\l_{f_{l} \leftarrow \l_t},\L_{t}^{pre},r\}$ as input, and outputs a refined background estimate of the current frame. Let $\sigma_{t,i}$ be the $i$-th largest sigular value of $[\L_t^{pre};\l_{f_l\leftarrow \l_t}]$, and $\u_{t,i}$, $\v_{t,i}$ be the corresponding left and right singular vectors. Further denote $\U_t=[\u_{t,1},\cdots,\u_{t,r}]$, $\boldsymbol{\Sigma}_t = \text{diag}(\sigma_{t,1},\cdots,\sigma_{t,r})$, and $\V_t=[\v_{t,1},\cdots,\v_{t,r}]$. Then for the best rank-$r$ denoising, $\hat{\l}_t = \mathcal{D}_{l}(\l_{f_l \leftarrow \l_t};\L_t^{pre},r)$ is given by the last column of $\U_t\boldsymbol{\Sigma}_t\V_t^T$.

In addition, for the $(t+1)$-th frame, the windowed background matrix $\L_{t+1}^{pre}$ is constructed by deleting the first column of $\L_t^{pre}$, and then appending $\hat{\l}_t$ as the last column. 
% \footnote{The complexity of the truncated SVD is high when the frame size is large. We can calculate the SVD of the $[\L_t^{pre};\l_{f_l\leftarrow \l_t}]$ using the incremental SVD \cite{bunch1978updating}. However, when we update $\L_{t+1}^{pre}$, we should delete the first column to maintain the window size unchanged, the calculation of SVD of $\L_{t+1}^{pre}$ require more computational cost than calculate the truncated SVD of $[\L_t^{pre};\l_{f_l\leftarrow \l_t}]$ directly.}.

We now describe the realization of $\mathcal{D}_{s}(\s_{f_{s} \leftarrow \s},v_{f_{s} \leftarrow \s};\S_t^{pre})$ that exploits both the sparsity and the continuity of the foregrounds in the estimation of the current foreground frame. Specifically, 
\begin{align}
\mathcal{D}_{s}(\s_{f_s\!\leftarrow \!\s_t},v_{f_s\!\leftarrow\! \s_t};\S_t^{pre})\!=\!\mathcal{D}_{s,2}(\mathcal{D}_{s,1}(\s_{f_s\!\leftarrow\! \s_t},v_{f_s\!\leftarrow\! \s_t});\S_t^{pre})
\end{align}
where $\mathcal{D}_{s,1}(\hat{\s},\hat{v})$ is a SURE-LET denoiser \cite{blu2007sure} that estimates foreground frame based on inputs and the sparsity of $\s_t$, and $\mathcal{D}_{s,2}(\hat{\s};\S_t^{pre})$ is an optical-flow estimator that enforces the continuity of the foreground frames by using the large displacement optical flow (LDOF) method \cite{brox2011large}. The LDOF estimator $\mathcal{D}_{s,2}(\hat{\s};\S_t^{pre})$ first estimates the horizontal and vertical motion vectors and then outputs an estimate of the current foreground vector by following the motion compensation method in Algorithm 1 of \cite{prativadibhayankaram2017compressive}.

\begin{algorithm}
\caption{Online DTMP}\label{online_dtmp}
\begin{algorithmic}[1]
\REQUIRE $\mathcal{A}_t, \y_t, \x_{\y_t\leftarrow\x_t}, v_{\y_t\leftarrow\x_t}, \s_{\delta\leftarrow\s_t}, v_{\delta\leftarrow\s_t}, \sigma^2, \S_{t}^{pre}, \L_{t}^{pre}.$
\\
\WHILE{the stopping criterion is not met}

\STATE $\x_{f_x\leftarrow\x_t} =\x_{\y_t\rightarrow\x_t} = \x_{\y_t\leftarrow\x_t} + \frac{n}{m} \mathcal{A}^T(\y_t-\mathcal{A}_t(\x_{\y_t\leftarrow\x_t}))$
\STATE $v_{f_x\leftarrow\x_t} = v_{\y_t\rightarrow\x_t} = \frac{n}{m}(v_{\y_t\leftarrow\x_t}+\sigma^2)-v_{\y_t\leftarrow\x_t}$

\STATE	$\x_{f_{x} \rightarrow \x_t}=c_{x}(\mathcal{D}_{x}(\x_{f_x \leftarrow \x_t},v_{f_{x} \leftarrow \x_t})-\alpha_{x}\x_{f_{x} \leftarrow \x_t})$ with $\alpha_x$ and $c_{x}$ in (\ref{xparams_c_a})
\STATE $v_{f_x\rightarrow\x_t} = \frac{1}{m}\left\|\y_t-\mathcal{A}\left(\x_{f_{x}\rightarrow \x_t}\right)\right\|_F^2-\sigma^2$

\STATE $v_{\delta\leftarrow\x_t} = \left(\frac{1}{v_{f_x\rightarrow\x_t}}+\frac{1}{v_{\y_t\rightarrow\x_t}}\right)^{-1}$
\STATE $\x_{\delta\leftarrow\x_t} = v_{\delta\leftarrow\x_t}\left(\frac{\x_{f_{x} \rightarrow \x_t}}{v_{f_{x} \rightarrow \x_t}}+\frac{\x_{\y_t\rightarrow\x_t}}{v_{\y_t\rightarrow\x_t}}\right)$

\STATE $\l_{f_{l}\leftarrow\l_t}=\l_{\delta \rightarrow \l_t} = \x_{\delta\leftarrow\x_t}-\s_{\delta\leftarrow\s_t}$
\STATE $v_{f_{l}\leftarrow\l_t}=v_{\delta \rightarrow \l_t} = v_{\delta\leftarrow\x_t}+v_{\delta\leftarrow\s_t}$

\STATE	$\l_{\delta\leftarrow\l_t}=\l_{f_{l} \rightarrow \l_t}=c_{l}(\mathcal{D}_{l}(\l_{f_{l} \leftarrow \l_t},v_{f_{l} \leftarrow \l_t})-\alpha_{l}\l_{f_{l} \leftarrow \l_t})$ with $\alpha_l$ and $c_{l}$ in (\ref{on_params_l})
\STATE $v_{\delta\leftarrow\l_t} = v_{f_{l} \leftarrow \l}\left(\left(1-\frac{r}{hw}\left(1+\frac{k}{hw}\right)\right)\frac{1}{(1-\alpha_l)^2}-1\right)=v_{f_{l} \rightarrow \l_t}$
\STATE $\s_{f_s \leftarrow \s_t}=\s_{\delta \rightarrow \s_t} = \x_{\delta\leftarrow\x_t}-\l_{\delta\leftarrow\l_t}$
\STATE	$v_{f_s \leftarrow \s_t}=v_{\delta \rightarrow \s_t} = v_{\delta\leftarrow\x_t}+v_{\delta\leftarrow\l_t}$
\STATE	$\s_{\delta\leftarrow\s_t}=\s_{f_{s} \rightarrow \s_t}=c_{s}(\mathcal{D}_{s}(\s_{f_{s} \leftarrow \s_t},v_{f_{s} \leftarrow \s_t})-\alpha_{s}\s_{f_{s} \leftarrow \s_t})$ with $\alpha_s$ and $c_{s}$ in (\ref{on_params_s})
\STATE $v_{\delta\leftarrow\s_t}=v_{f_{s} \rightarrow \s_t}= \frac{1}{m}\|\y_t-\mathcal{A}(\s_{f_{s} \rightarrow \s_t}+\l_{\delta\leftarrow \l_t}))\|_2^2-v_{\delta\leftarrow \l_t}-\sigma^2$

\STATE $\x_{\delta\rightarrow\x_t} = \l_{\delta\leftarrow\l_t}+\s_{\delta\leftarrow\s_t}$
\STATE $v_{\delta\rightarrow\x_t} = v_{\delta\leftarrow\l_t}+v_{\delta\leftarrow\s_t}$
\STATE $v_{\y_t\leftarrow\x_t} = \left(\frac{1}{v_{\delta\rightarrow\x_t}}+\frac{1}{v_{f_x\rightarrow\x_t}}\right)^{-1}$
\STATE $\x_{\y_t\leftarrow\x_t} = v_{\y_t\leftarrow\x_t}\left(\frac{\x_{f_{x} \rightarrow \x_t}}{v_{f_{x} \rightarrow \x_t}}+\frac{\x_{\delta\rightarrow\x_t}}{v_{\delta\rightarrow\x_t}}\right)$
\ENDWHILE
\STATE Update foreground matrix $\S_{t+1}^{pre}$ and background matrix $\L_{t+1}^{pre}$
\ENSURE  $\hat{\x}_t=\hat{\l}_t+\hat{\s}_t, \hat{\l}_t=\mathcal{D}_{l}(\l_{f_{l} \leftarrow \l_t},v_{f_{l} \leftarrow \l_t}),\hat{\s}_t=\mathcal{D}_{s}(\s_{f_{s} \leftarrow \s_t},v_{f_{s} \leftarrow \s_t})$, $\S_{t+1}^{pre}$, $\L_{t+1}^{pre}$.
\end{algorithmic}
\end{algorithm}

% Compared to the TMP algorithm \cite{xue2018turbo}, extra steps that exploit the image structural information is included in the online DTMP algorithm (Lines 4-7 and Lines 16-19 of Algorithm \ref{online_dtmp}.). Sliding window based low-rank denoiser (Line 10 of Algorithm \ref{online_dtmp}) is employed to refine the background estimation. Foreground denoiser that combines sparse denoiser such as SURE-LET and optical-flow estimator is employed to refine the foreground estimation (Line 14 of Algorithm \ref{online_dtmp}).

\subsection{Complexity Analysis}\label{online_complexity}
The computational complexity of the online DTMP algorithm are dominated by the operations in Lines 2, 4, 10 and 14. The computational complexity of Line 2 is $\mathcal{O}(m n_1)$ for a general linear opearator $\mathcal{A}$. For a partial orthogonal DCT operator, the complexity can be reduced to $\mathcal{O}(n_1 \log (n_1))$. The complexity of Line 4 concentrates on the processing of image denoiser $\mathcal{D}_{x}$ which is usually linear to the size of input image. The complexity of the operation in Line 10 mainly concentrates on the step of calculation of $\mathcal{D}_{l}(\l_{f_l\leftarrow\l},v_{f_l\leftarrow\l})$ which involves the computation of the truncated SVD of $[\L_t^{pre};\hat{\l}_t]$ with complexity $\mathcal{O}(r n_1 k_{L})$ (or using incincremental SVD which has $\mathcal{O}(r n_1 k_{L})$ complexity). The complexity of the operation in Line 14 mainly concentrates on the calculation of $\mathcal{D}_{s}(\s_{f_s\leftarrow\s},v_{f_s\leftarrow\s})$. When the SURE-LET denoiser with the kernel given by \cite[Eqs. 28-30]{xue2017denoising} is chosen, the complexity is $\mathcal{O}(n_1)$. Thus, the complexity of the online DTMP algorithm is $\mathcal{O}(r n_1 k_{L})+\mathcal{O}(n_1 \log (n_1))$ flops when a partial orthogonal DCT operator is adopted.

% /////////////////////////////////////////
\section{State Evolution}\label{sec_state_evo}
In this section, we present the state evolution analysis to characterize the behavior of the offline and online DTMP algorithms. We focus on the state evolution of the offline DTMP algorithm and will briefly discuss the extension to the online version at the end of this section.

\subsection{State Evolution of Offline DTMP}
\begin{figure}
	\centering
	\includegraphics[width=0.8\linewidth]{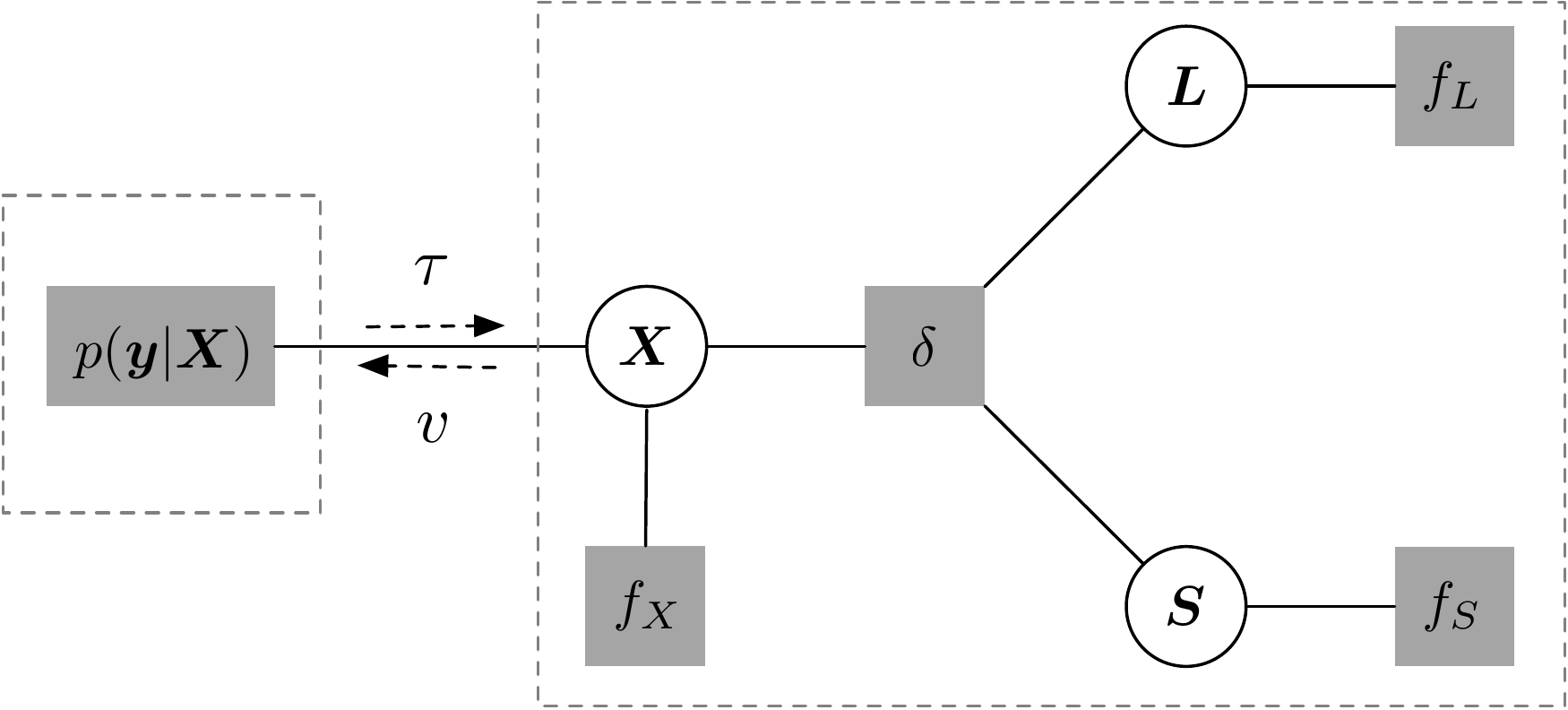}
	\caption{The state variables of the offline DTMP algorithm.}
	\label{states}
\end{figure}

In the state evolution analysis, we characterize the behavior of the offline DTMP algorithm by tracking two statistics: the mean squared error (MSE) of $\X_{\y\rightarrow \X}$ denoted by $\tau$ and the MSE of $\X_{\y\leftarrow \X}$ denoted by $v$. We add subscript $t$ to represent the state variables at iteration $t$. An illustration of the two state variables is plotted in Fig. \ref{states}. In Fig. \ref{states}, the factor graph of offline DTMP is separated into two parts. The left part of the factor graph corresponds to Lines 2-3 of Algorithm \ref{batch_dtmp} that outputs a linear estimation of $\X$ based on the observation $\y$. The right part corresponds to Lines 4-19 of Algorithm \ref{batch_dtmp} that outputs a nonlinear estimation of $\X$ based on the three constraints $f_X$, $f_L$, and $f_S$. Then, the performance of the offline DTMP algorithm can be characterized by
\BS
\begin{align}
\tau_t &= f(v_t)\\
v_{t+1} &= g(\tau_t)
\end{align}\label{se_offline_gf}\ES
where $f$ and $g$ are the the MSE transfer functions of the left part and the right part of Fig. \ref{states} respectively.

\begin{figure}
	\centering
	\includegraphics[width=0.49\linewidth]{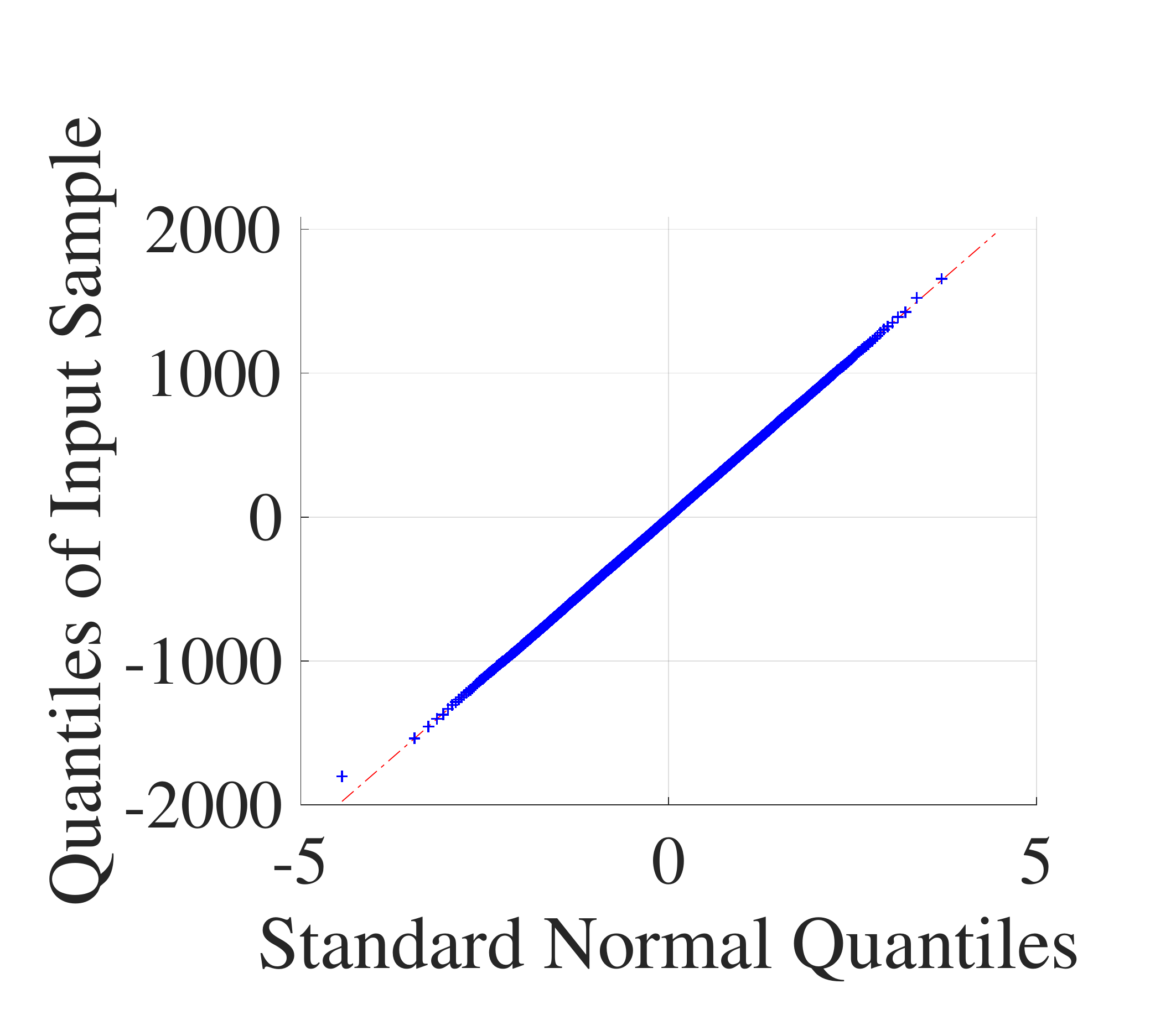}
	\includegraphics[width=0.49\linewidth]{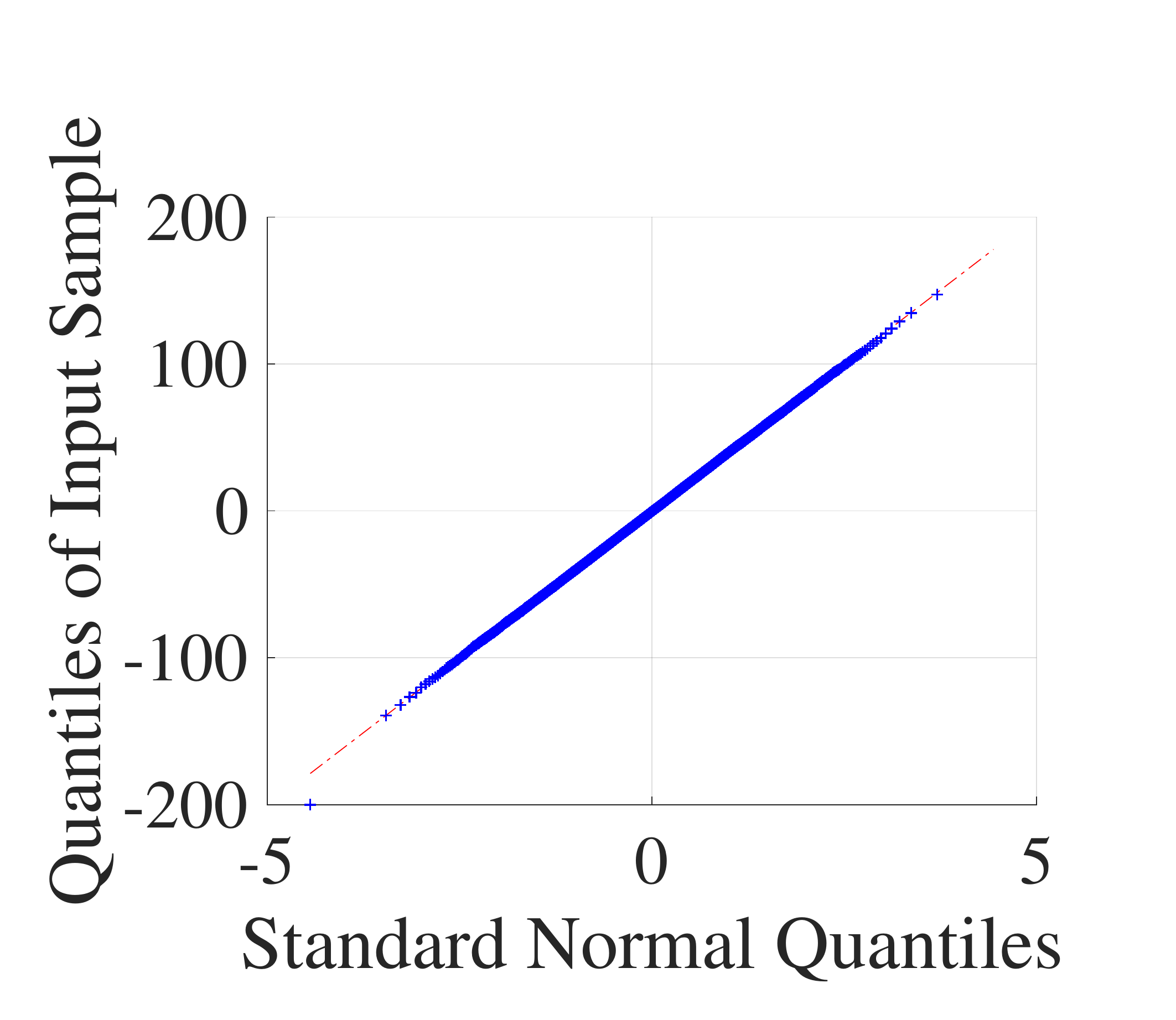}
	\caption{QQplots of $\text{vec}(\X_{f_X\leftarrow\X} -\X)$ at the first and 10-th iteration of the offline DTMP algorithm. The linear operator is chosen as the partial orthogonal discreate cosine transform operator given in \ref{poperator}, the video dataset is chosen as the ``Twoleaveshop2cor'' \cite{cavdataset}, and $m/(n_1 n_2)$ is set to 0.1.}
	\label{QQplot1}
\end{figure}

The MSE transfer function $f$ in (\ref{se_offline_gf}a) can be derived from Lines 2-3 of Algorithm \ref{batch_dtmp} by recalling that Lines 2-3 correspond to the LMMSE estimation of $\X$. From (\ref{m_v_m_y_X}b), the expression of $f$ is given by
\begin{align}
	f(v) = \frac{n_1 n_2}{m}(v+\sigma^2)-v.
\end{align}
We next determine the MSE transfer function $g$ in (\ref{se_offline_gf}b). A key observation is that $\X_{\y \rightarrow \X}$ in Line 2 of Algorithm \ref{batch_dtmp} can be modelled as
\begin{align}
	\X_{\y \rightarrow \X} = \X + \tau \N \label{Gauss_asmp}
\end{align}
where $\N\in \mathbb{R}^{n_1 \times n_2}$ is a random Gaussian matrix with entries draw from $\mathcal{N}(0,1)$ independently. As shown in Fig. \ref{QQplot1}, the QQplot of the residual term $\X_{f_X\leftarrow\X} -\X$ is plotted. From Fig. \ref{QQplot1}, we see that the residual term resembles an i.i.d. Gaussian noise at different iterations of offline DTMP. With the input model (\ref{Gauss_asmp}), the MSE transfer function of the right part of Fig. \ref{states} is given by 
\begin{align}
	v = g(\tau) = \frac{1}{n_1 n_2}\E\left[\|\mathcal{D}_{r}(\X+\sqrt{\tau}\N,\tau)-\X\|_F^2\right]
\end{align}
where $\mathcal{D}_{r}(\X+\sqrt{\tau}\N,\tau)$ is the output of the right part of Fig. \ref{states} with the input modelled by (\ref{Gauss_asmp}), and the expectation is taken over the probability model of (\ref{Gauss_asmp}).

\subsection{State Evolution of Online DTMP}
Different from the offline DTMP algorithm, in the online DTMP algorithm, the previous recovered backgrounds and foregrounds are available when we process the current frame which makes it difficult to track the performance of online DTMP on a given frame since the recovery process depends on previous recovery results. To track the performance of online-DTMP, we assume that for the current frame, the previous recovery results are known and the information provided to the current frame is fixed. Based on the assumption, the SE of online-DTMP can be established which has a similar form to the SE of given as follows:
\BS
\begin{align}
	\tau_t &= f_t(v_{t})=\frac{n_1}{m}(v_t+\sigma^2)-v_t\\
	v_{t+1} &= g_t(\tau_t)=\frac{1}{n_1}\E\left[\|\mathcal{D}_{r}(\x_t+\sqrt{\tau_t}\n; \L_{t}^{pre},\S_{t}^{pre})-\x_t\|_2^2\right].
\end{align}\label{se2}\ES
where $\mathcal{D}_{r}$ is the right nonlinear part of the online DTMP algorithm. Numerical results will be presented in the next section to verify the effectiveness of the developed state evolution. 

% /Similar to the SE of the offline DTMP algorithm, we will present the SE in the following numerical results section.

%/////////////////////////////////
\section{Numerical Results}\label{section_results}

\subsection{Simulation Settings}
The offline and online DTMP algorithms can be applied to compressed VBS with various linear operators such as Gaussian random linear operator and partial orthogonal linear operator. In practice, a general random linear operator may impose high storage and computation requirements on video implementations. To reduce the computational complexity and storage requirement, we choose the partial discrete cosine transform (DCT) operator in the experiments since there is no need to store the DCT operator in implementation and the fast DCT algorithm can be adopted in the computation. We generate the partial DCT operator $\mathcal{A}$ with its matrix form constructed as
\begin{align}
	\A = \boldsymbol{P} \D\label{poperator}
\end{align}
where $\A\in\mathbb{R}^{m\times hw}$ is the matrix form of linear operator $\mathcal{A}$, $\boldsymbol{P}\in \mathbb{R}^{m\times n}$ is a random selection matrix that selects rows randomly from $\D$, and $\D\in \mathbb{R}^{n \times n}$ is the DCT matrix. For offline video compression settings, $n = hw n_2$, and for online video compression settings, $n=hw$.

In the simulations, we choose eight video dataset, ``Twoleaveshop2cor'' ($288 \times 384$ pixels) from dataset "CAVIAR Test Case Scenarios" \cite{cavdataset}, ``CameraParameter'' ($240 \times 320$ pixels) from dataset \cite{limudataset}, ``ShoppingMall'' ($320\times 256$ pixels) from dataset "I2rdataset" \cite{i2rdataset} and ``tramcrossroad'', ``winterstreet'', ``PETS2016'', ``wetsnow'', and ``canoe'' from dataset "CDnet2014" \cite{wang2014cdnet} that contain different real application scenes such as `Bad weather'', ``Low Framerate'', ``Night videos'', ``Dynamic background'', ``Indoor activity''. The measurement rate is defined by $\frac{m}{n_1 n_2}$, the normalized mean square error (NMSE) is defined as $\frac{\|\hat{\L}+\hat{\S}-\X\|_F^2}{\|\X\|_F^2}$ and the peak signal-to-noise ratio (PSNR) is defined as $10\log(\frac{255^2}{\|\hat{\L}+\hat{\S}-\X\|_F^2})$. The rank of background matrix should be 1 for static background, however, most background change slowly. Thus, we set the rank of the background matrix to $2$ in our simulations. In the offline DTMP algorithm, we choose SURE-LET for the sparse denoiser $\mathcal{D}_{S}$, BM3D for the image denoiser $\mathcal{D}_{X}$, and the best-rank-$r$ for the low-rank denoiser $\mathcal{D}_{L}$.

\subsection{Offline Compressed VBS}
In the following, we compare the performance of the offline DTMP algorithm with counterpart algorithms including the TMP algorithm \cite{xue2018turbo} and the SpaRCS algorithm \cite{waters2011sparcs} for the offline compressed VBS problem and characterize the performance of the offline DTMP by the state evolution analysis. We compare these algorithms on video sequences ``CameraParameter'' and ``Twoleaveshop2cor''. The video matrix $\X$ is generated by choosing 500 frames from the beginning of the video sequence, reshaping each frame into a vector, and arranging these vectors column by column.

\begin{figure}
	\centering
	\includegraphics[width=\linewidth]{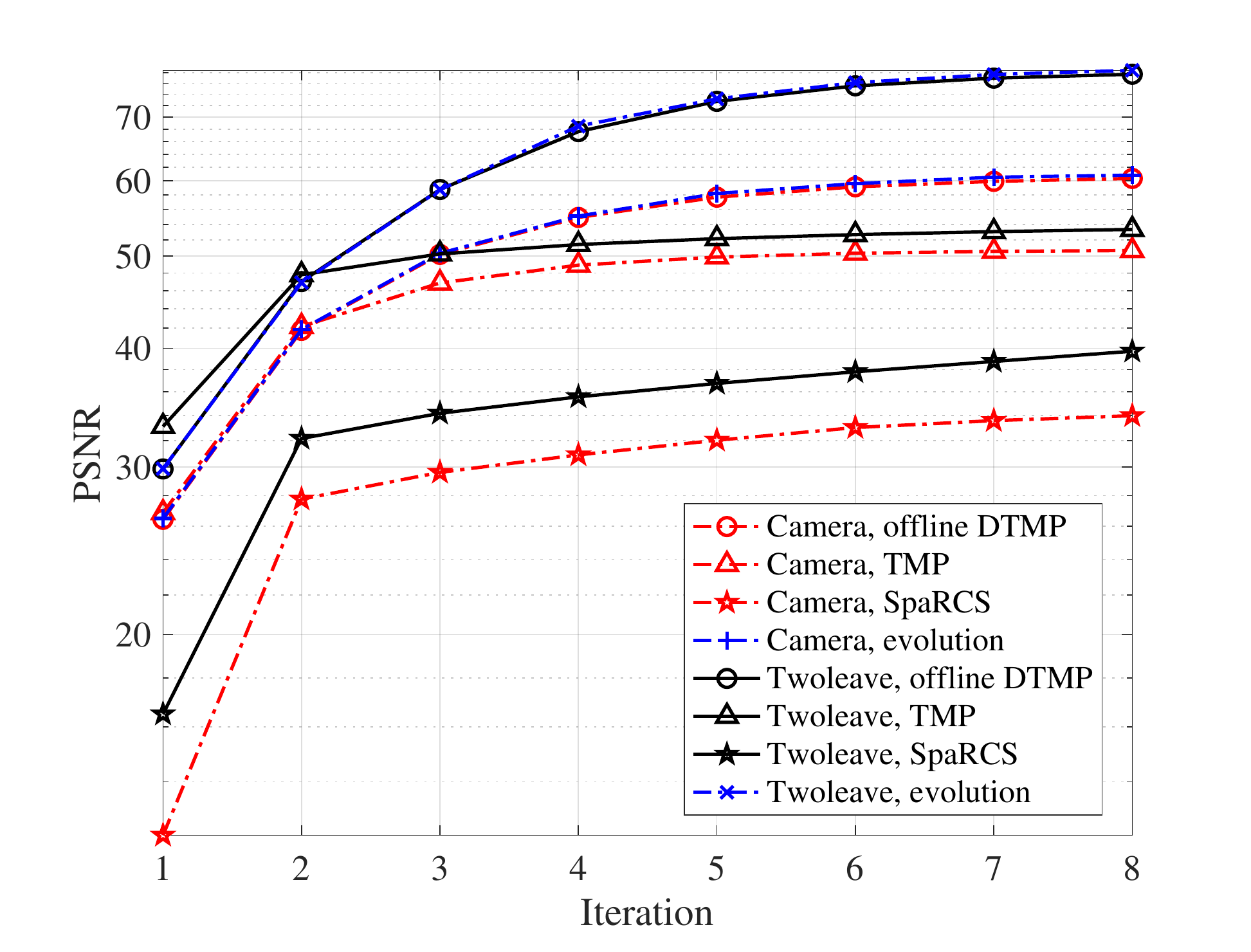}
	\caption{The PSNR performance of the offline DTMP algorithm, the TMP algorithm and the SpaRCS algorithm on video datasets ``CameraParameter'' and ``Twoleaveshop2cor'' and the SE of the offline DTMP algorithm. The measurement rate is set to 0.1.}\label{off_se_comp}
\end{figure}

Fig. \ref{off_se_comp} plots the PSNR of offline DTMP, TMP, and SpaRCS against the iteration number on different video sequences and the SE of the offline DTMP algorithm. In the figure, the curves with legend ``Camera'' corresponds to video ``CameraParameter'', and those with legend ``Twoleave'' correspond to ``Twoleaveshop2cor''. From the figure, we see that the offline DTMP algorithm achieves a much lower PSNR than TMP and SpaRCS in both datasets, and the state evolution analysis accurately characterize the performance of offline DTMP.

\begin{figure}
	\centering
	\includegraphics[width=0.49\linewidth]{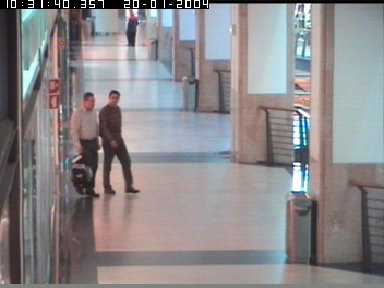}
	\includegraphics[width=0.49\linewidth]{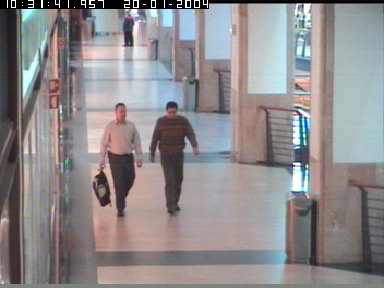}
	\includegraphics[width=0.32\linewidth]{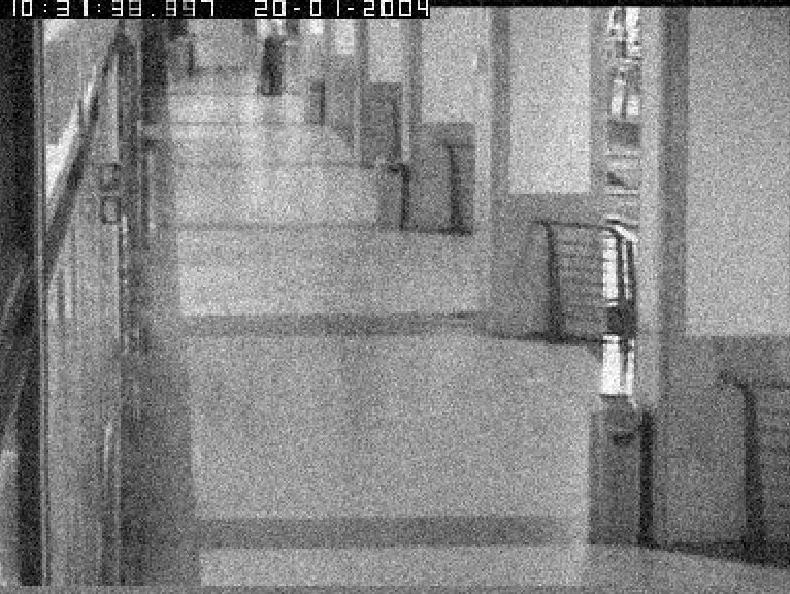}
	\includegraphics[width=0.32\linewidth]{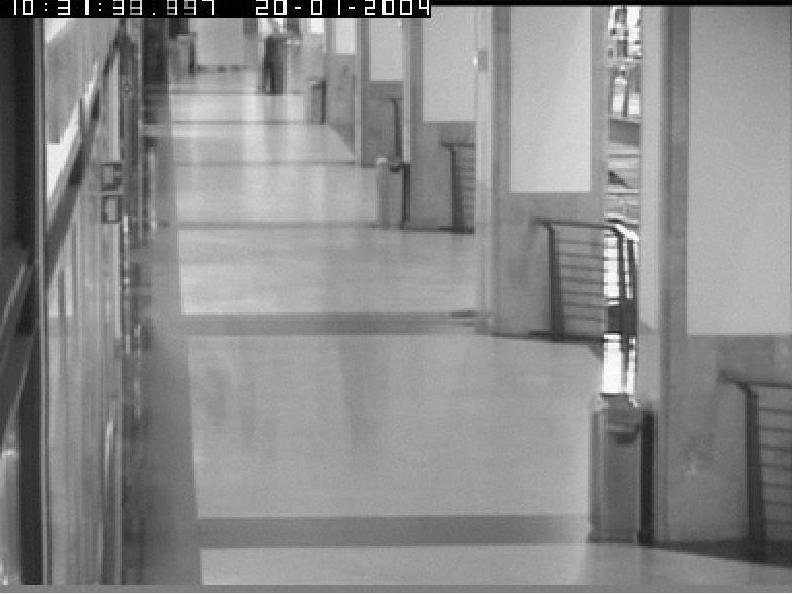}
	\includegraphics[width=0.32\linewidth]{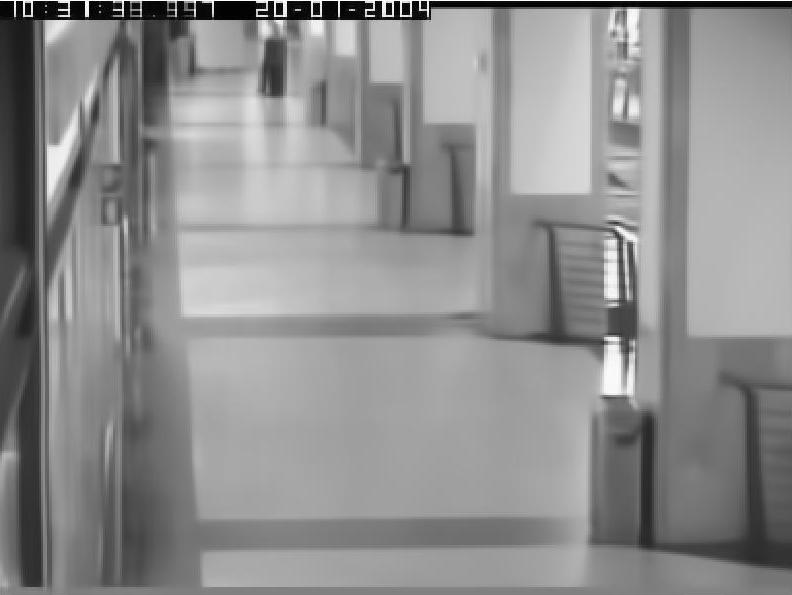}
	\includegraphics[width=0.32\linewidth]{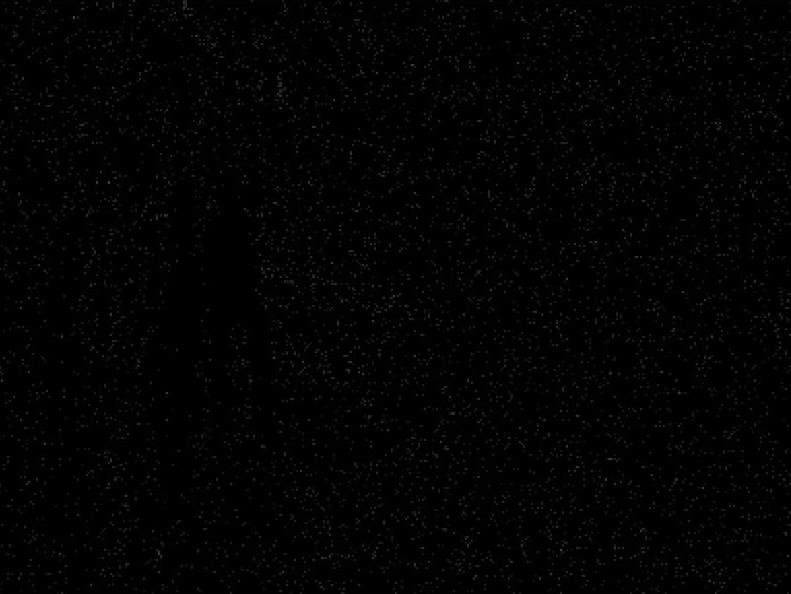}
	\includegraphics[width=0.32\linewidth]{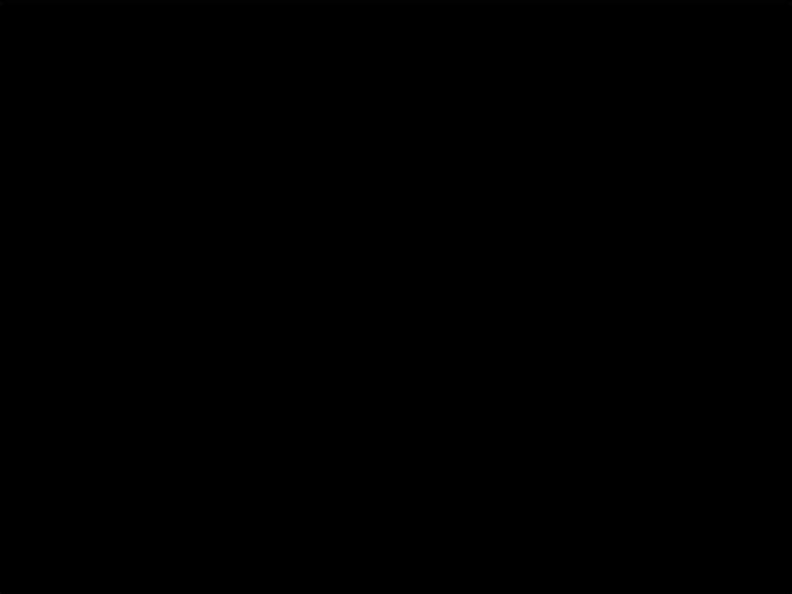}
	\includegraphics[width=0.32\linewidth]{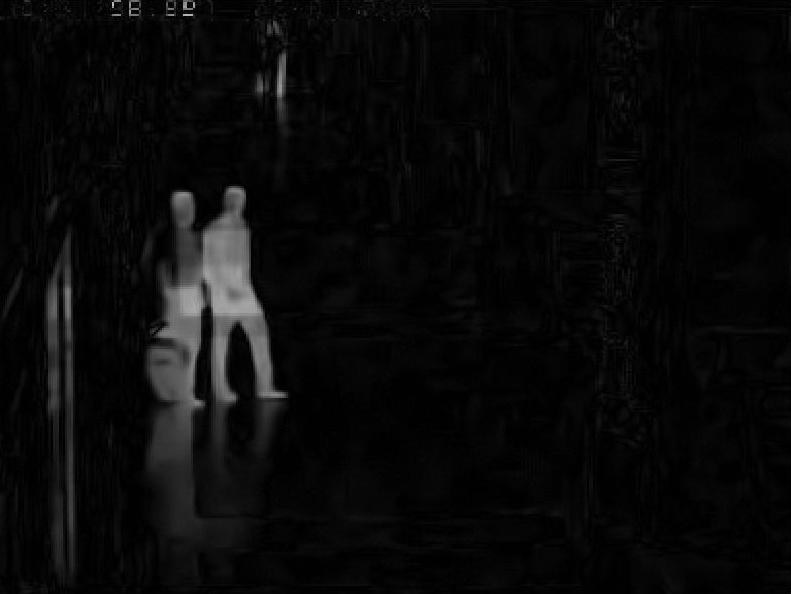}
	\includegraphics[width=0.32\linewidth]{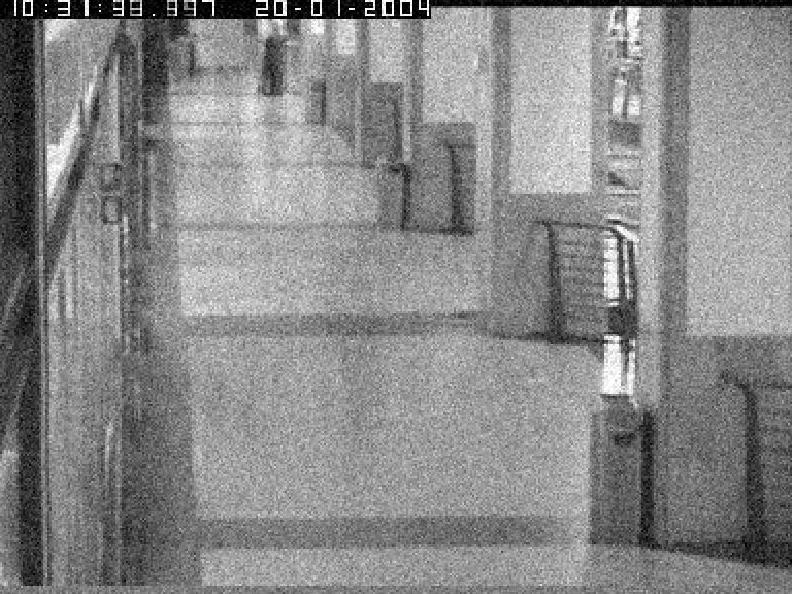}
	\includegraphics[width=0.32\linewidth]{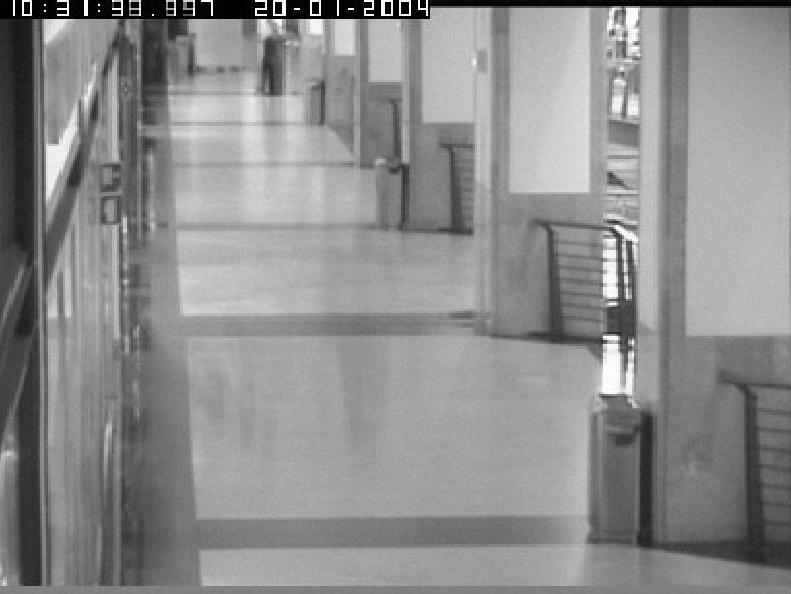}
	\includegraphics[width=0.32\linewidth]{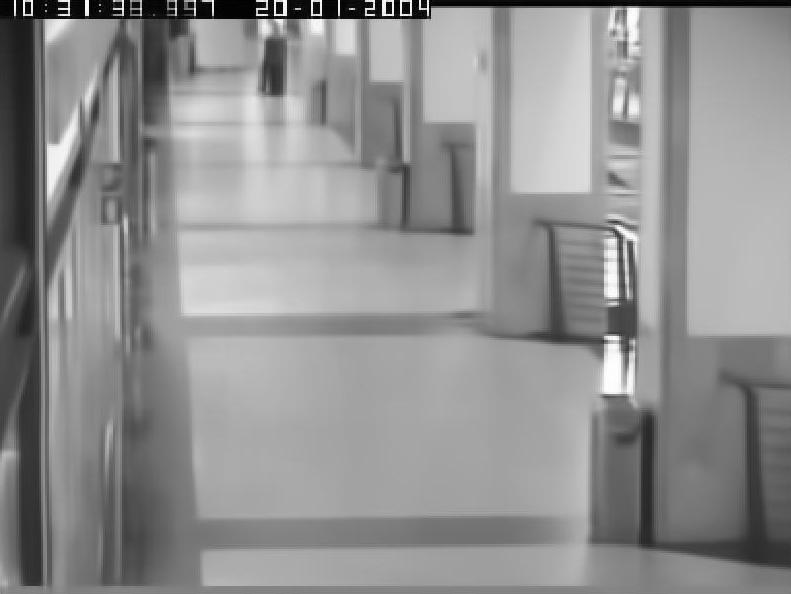}
	\includegraphics[width=0.32\linewidth]{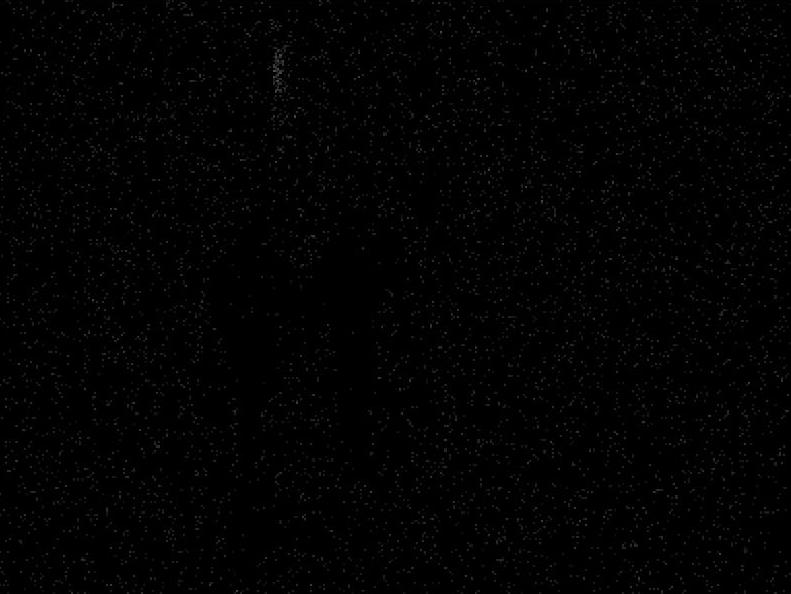}
	\includegraphics[width=0.32\linewidth]{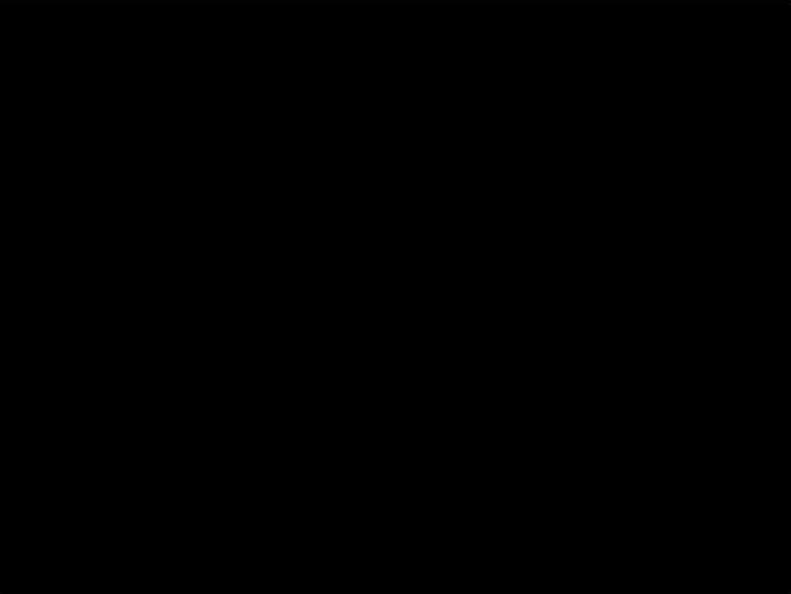}
	\includegraphics[width=0.32\linewidth]{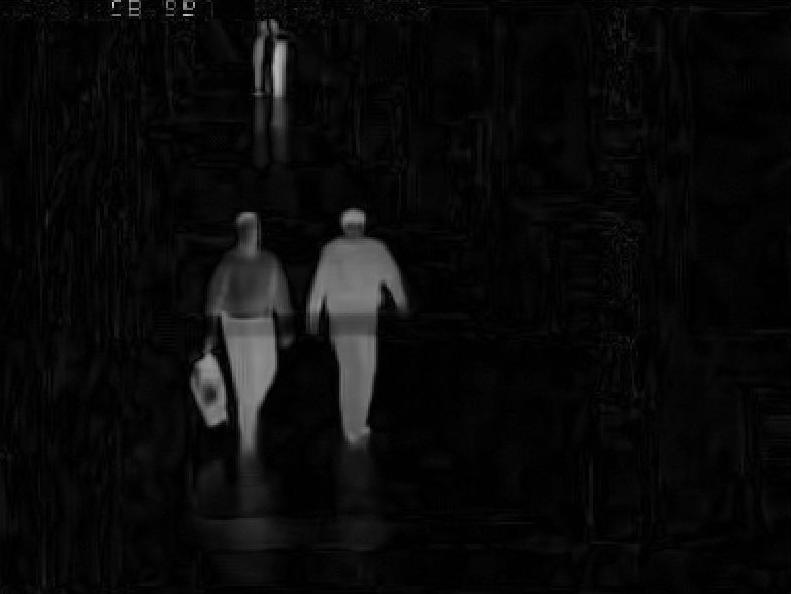}
	\caption{Top row: original images; lower left column: SpaRCS. lower middle column: TMP; lower right column: DTMP. The measurement rate is set to 0.1. Video dataset: ``Twoleaveshop2cor''.}\label{twoleave_01_offline}
\end{figure}

\begin{figure}
	\centering
	\includegraphics[width=0.49\linewidth]{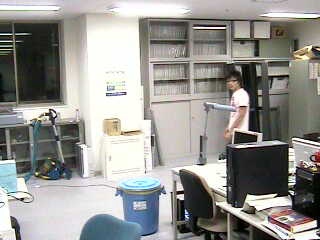}
	\includegraphics[width=0.49\linewidth]{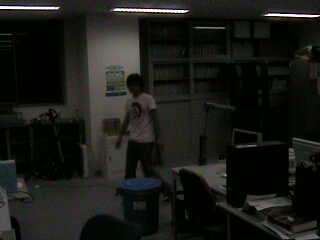}
	\includegraphics[width=0.325\linewidth]{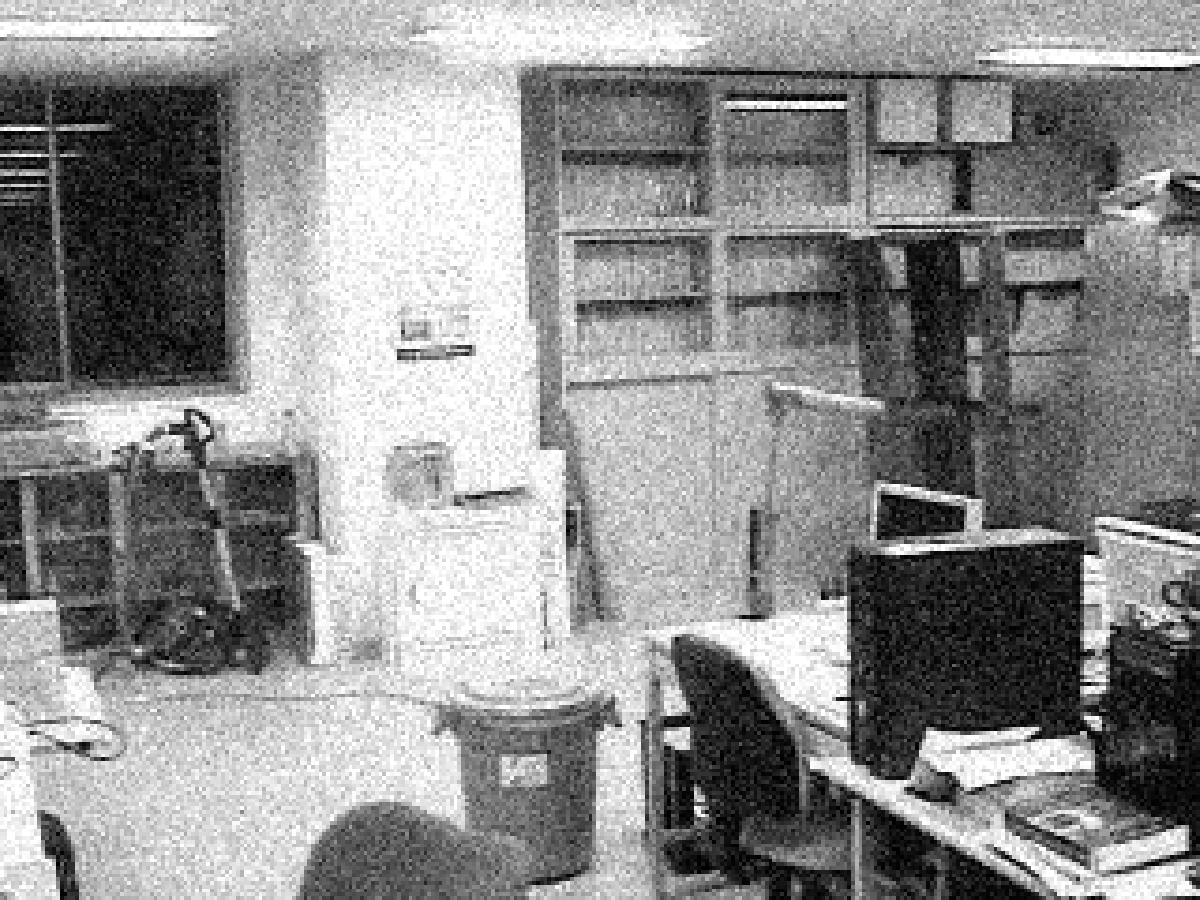}
	\includegraphics[width=0.325\linewidth]{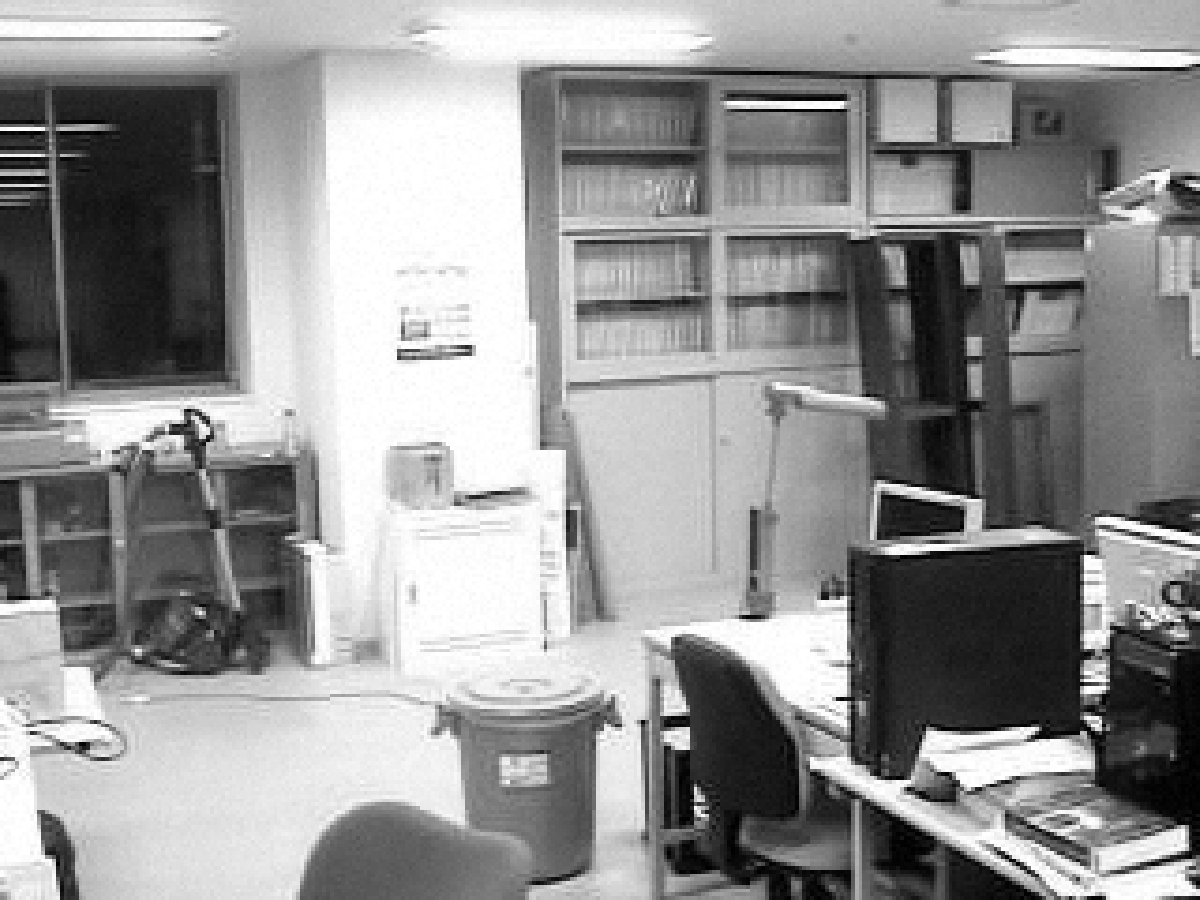}
	\includegraphics[width=0.325\linewidth]{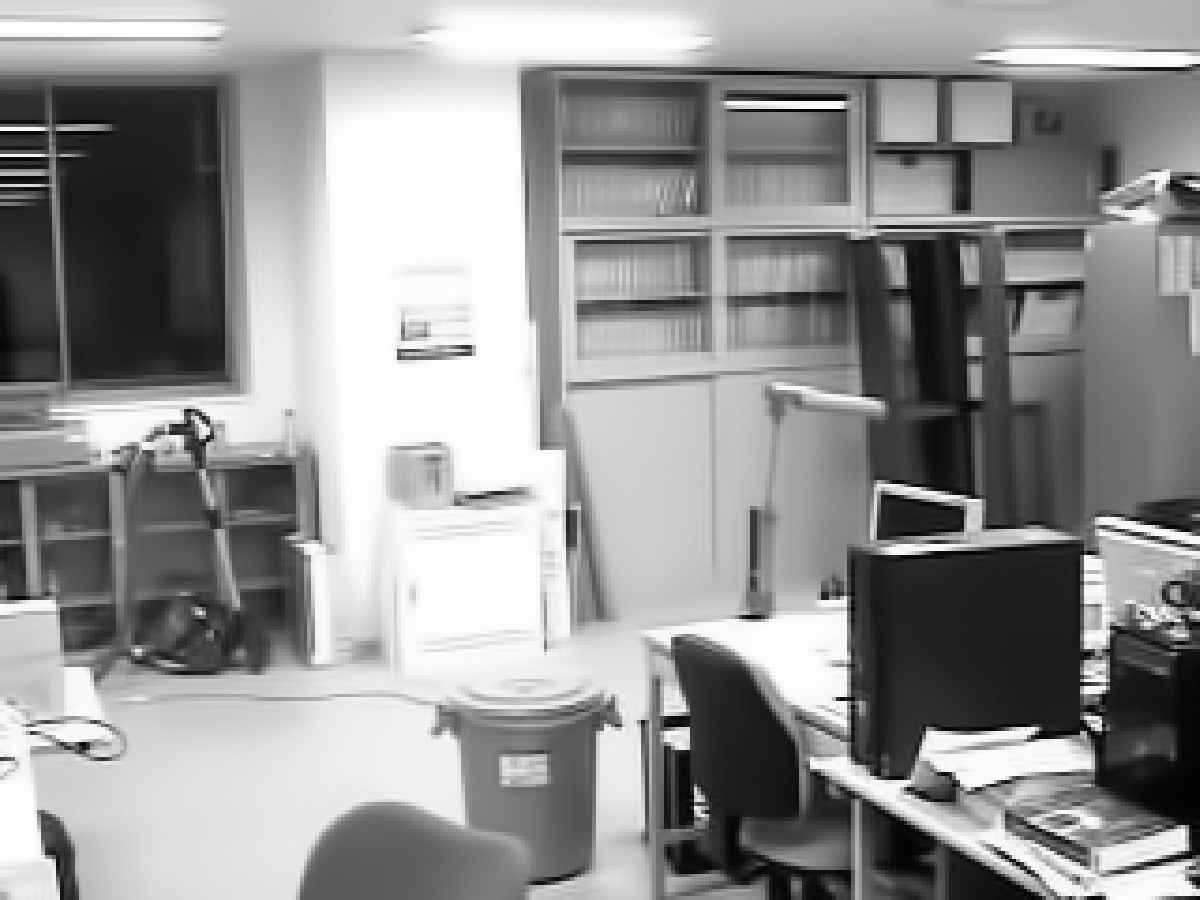}
	\includegraphics[width=0.325\linewidth]{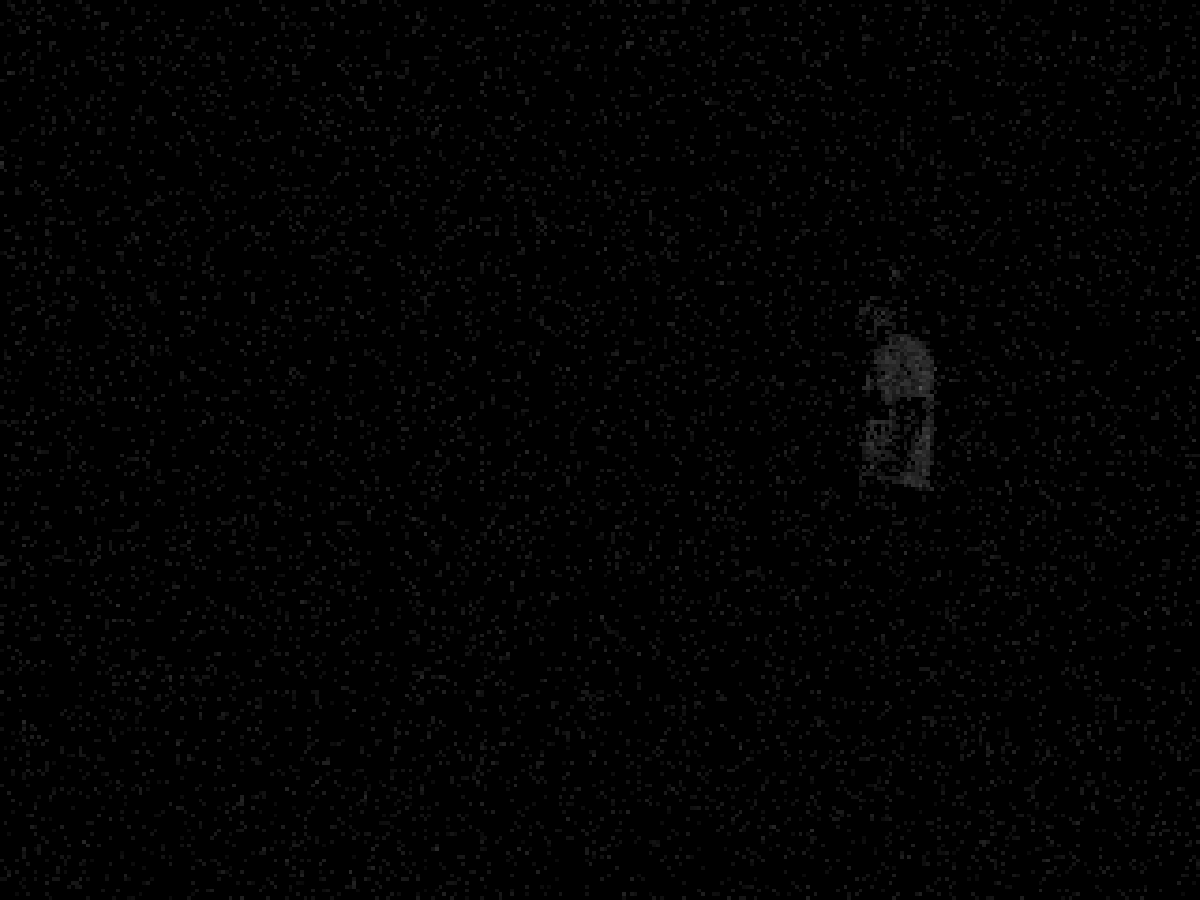}
	\includegraphics[width=0.325\linewidth]{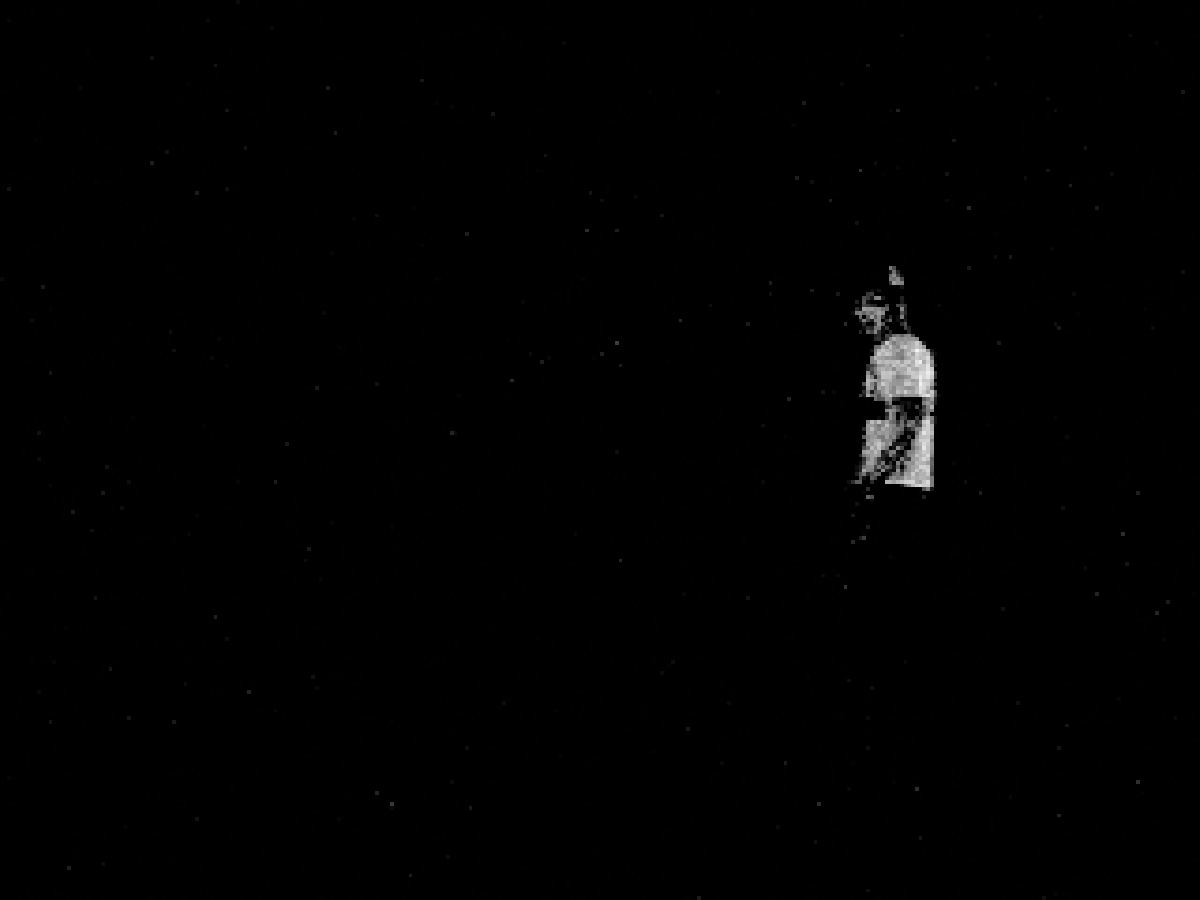}
	\includegraphics[width=0.325\linewidth]{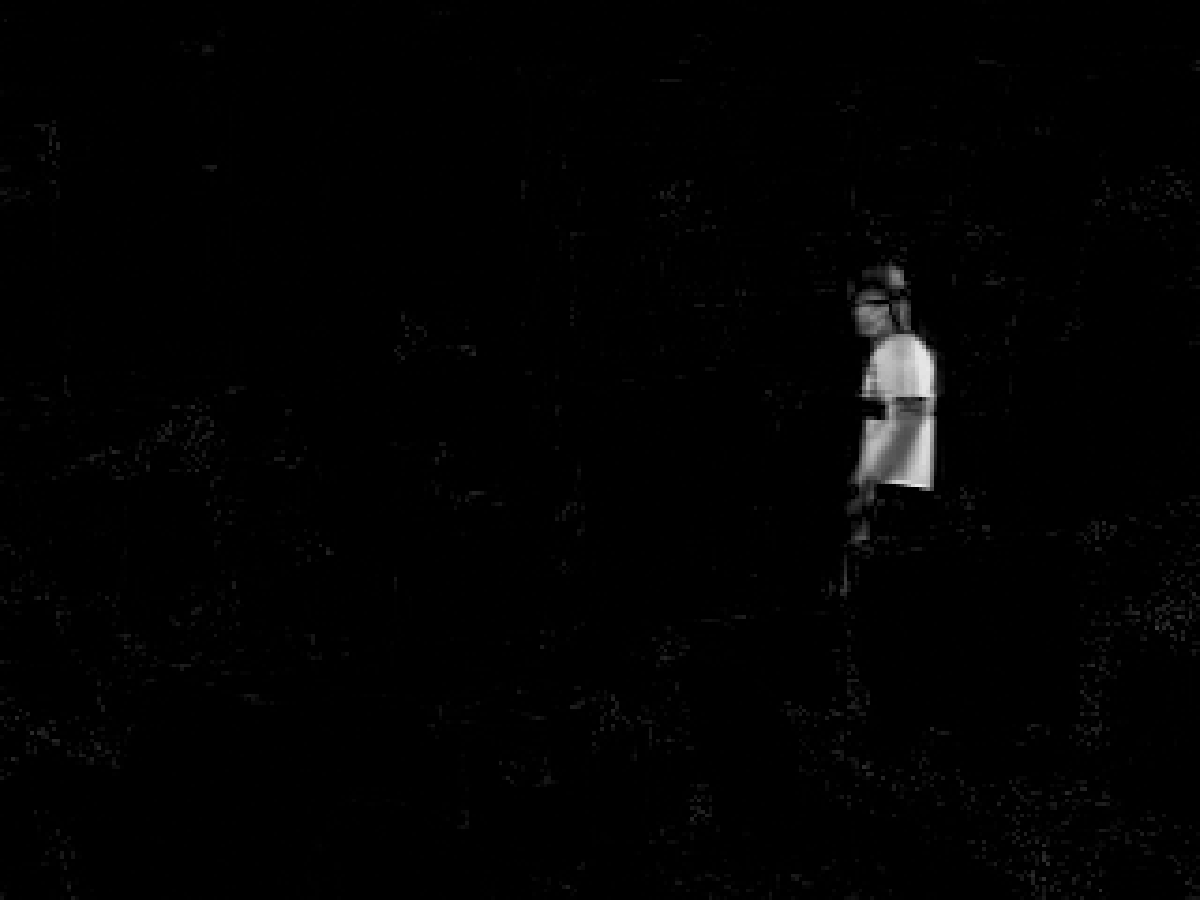}
	\includegraphics[width=0.325\linewidth]{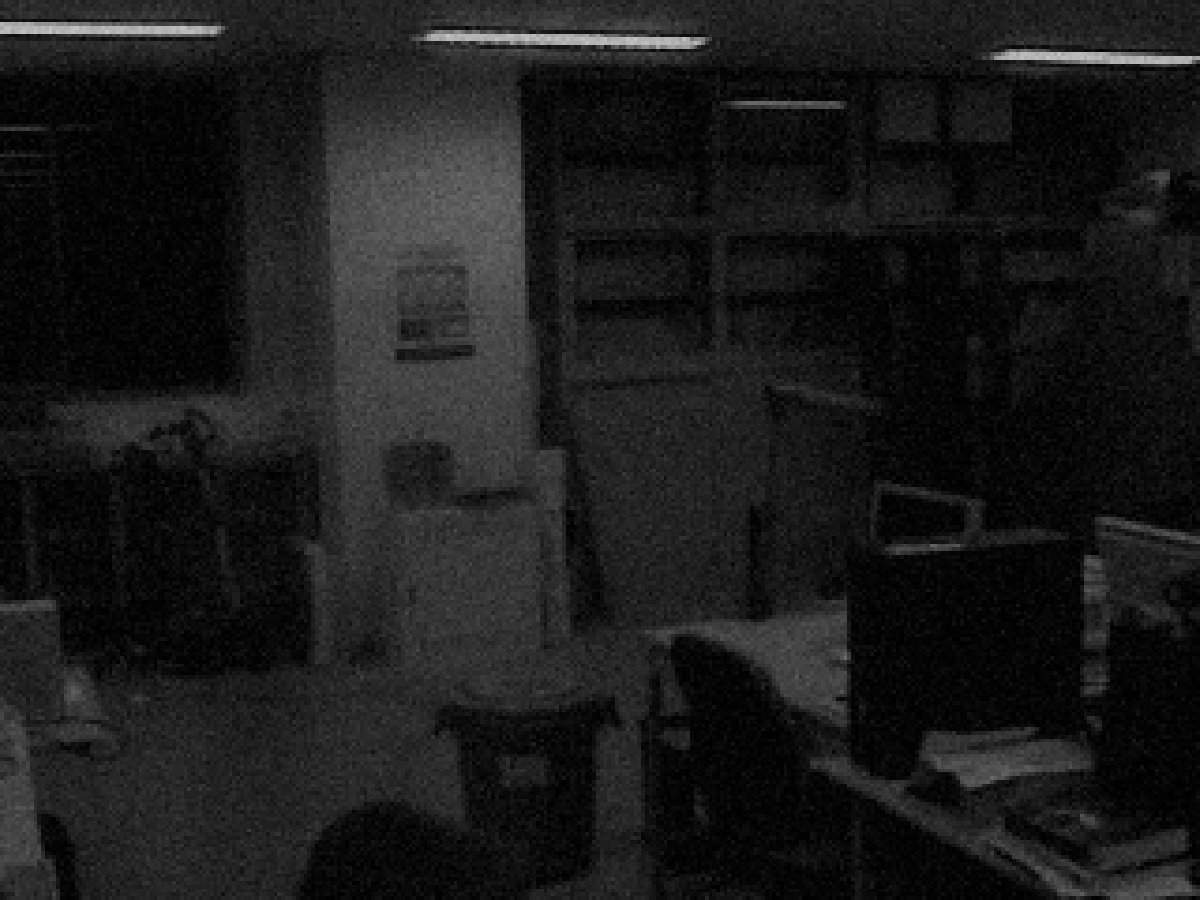}
	\includegraphics[width=0.325\linewidth]{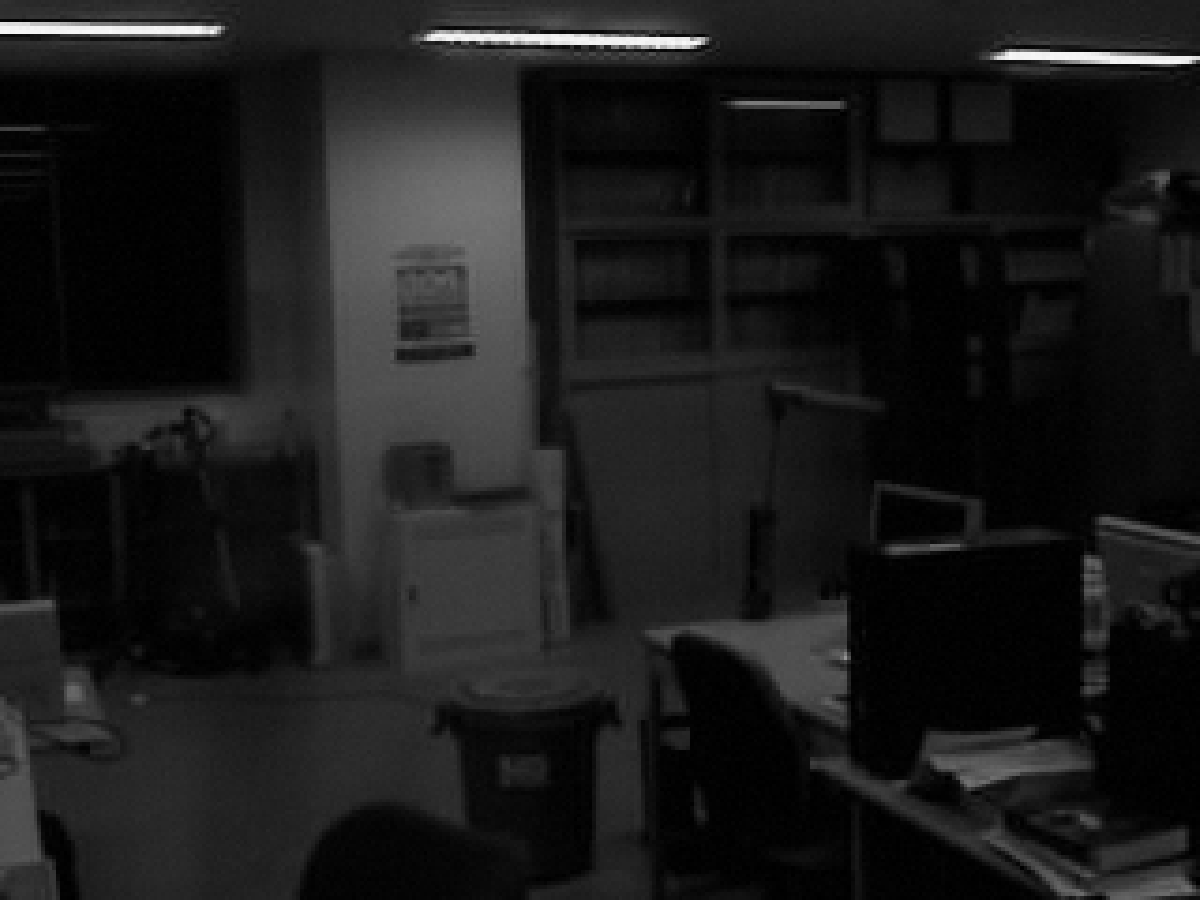}
	\includegraphics[width=0.325\linewidth]{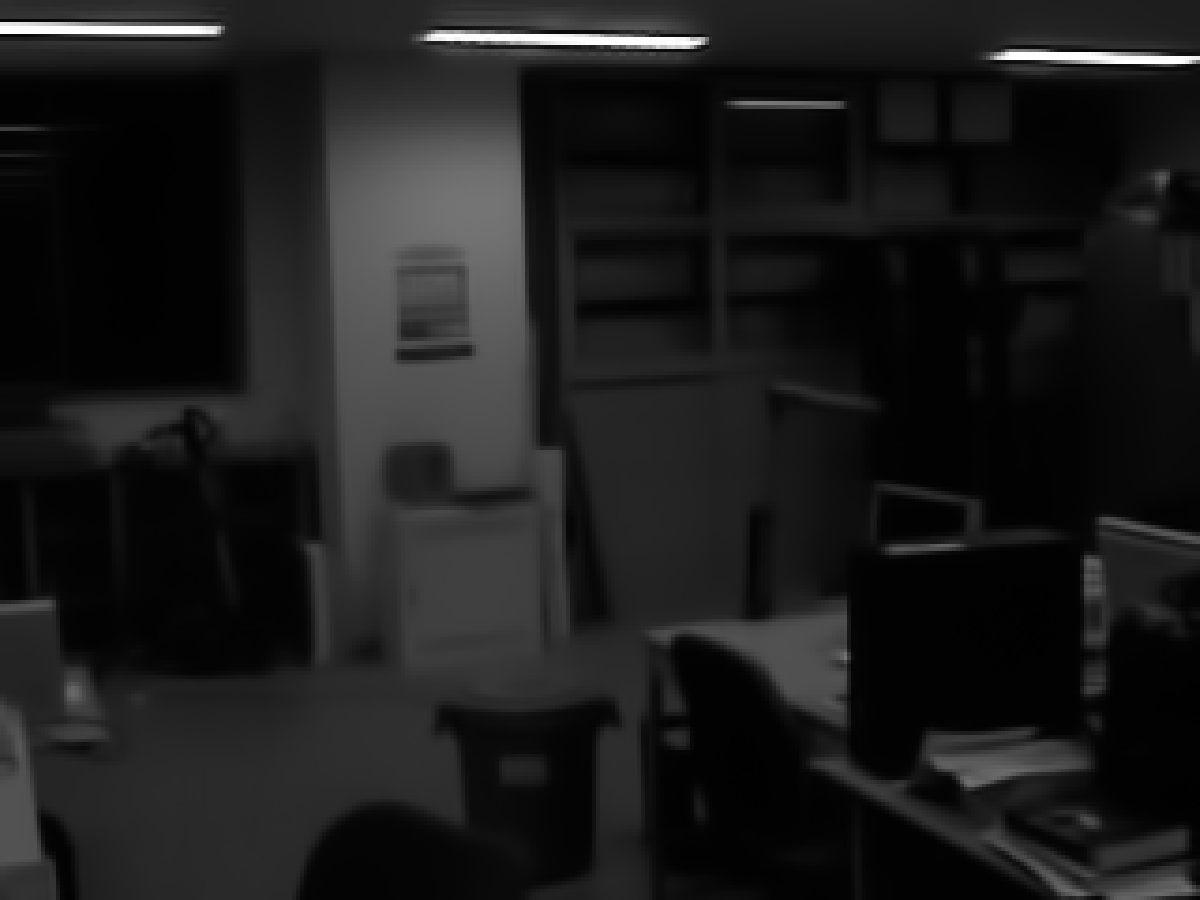}
	\includegraphics[width=0.325\linewidth]{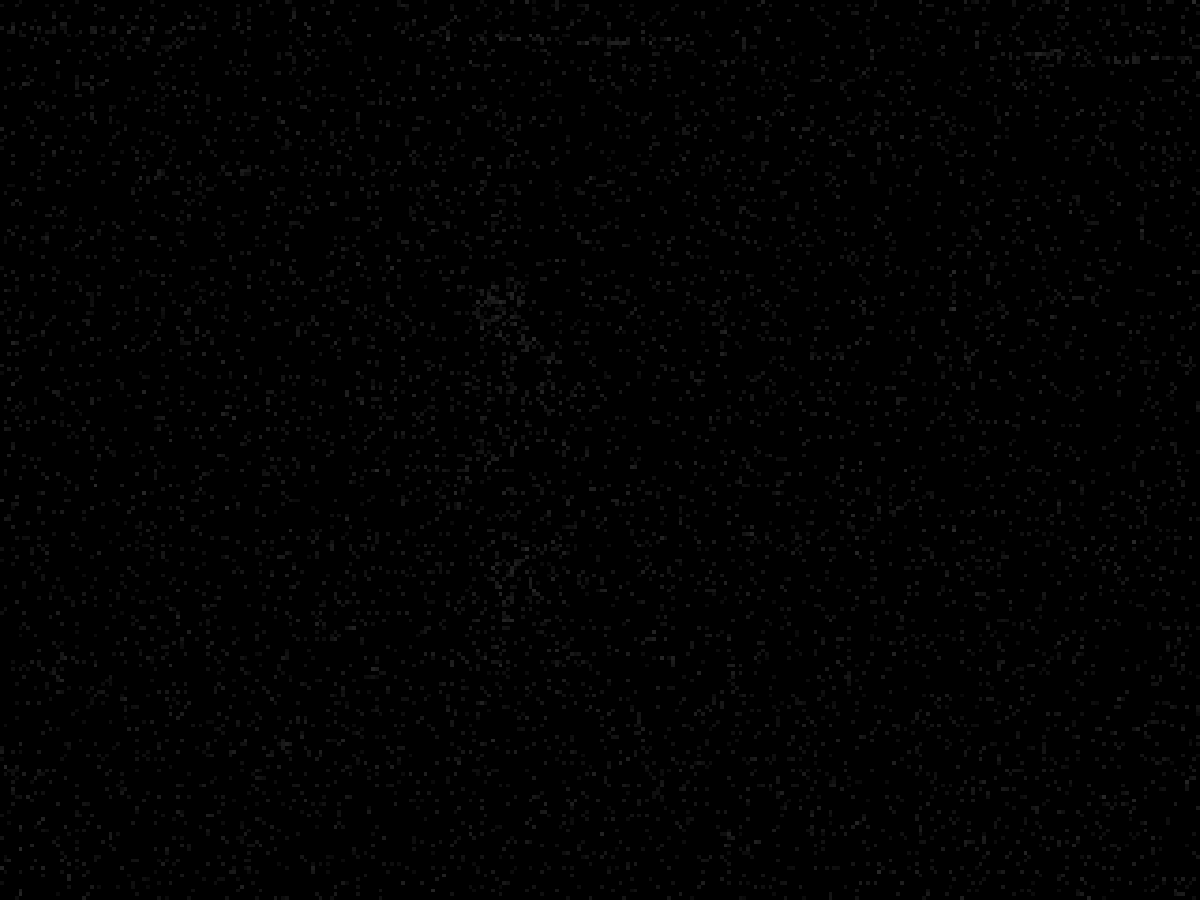}
	\includegraphics[width=0.325\linewidth]{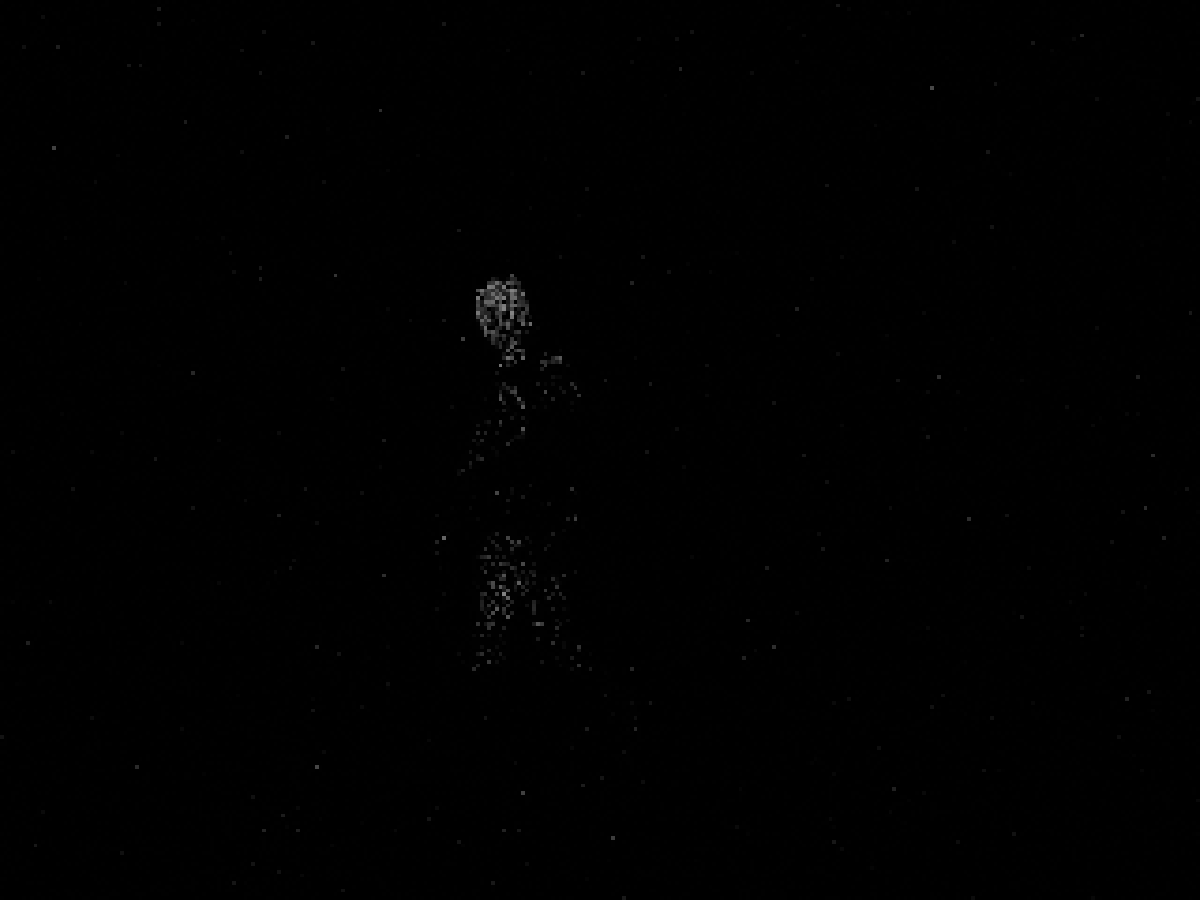}
	\includegraphics[width=0.325\linewidth]{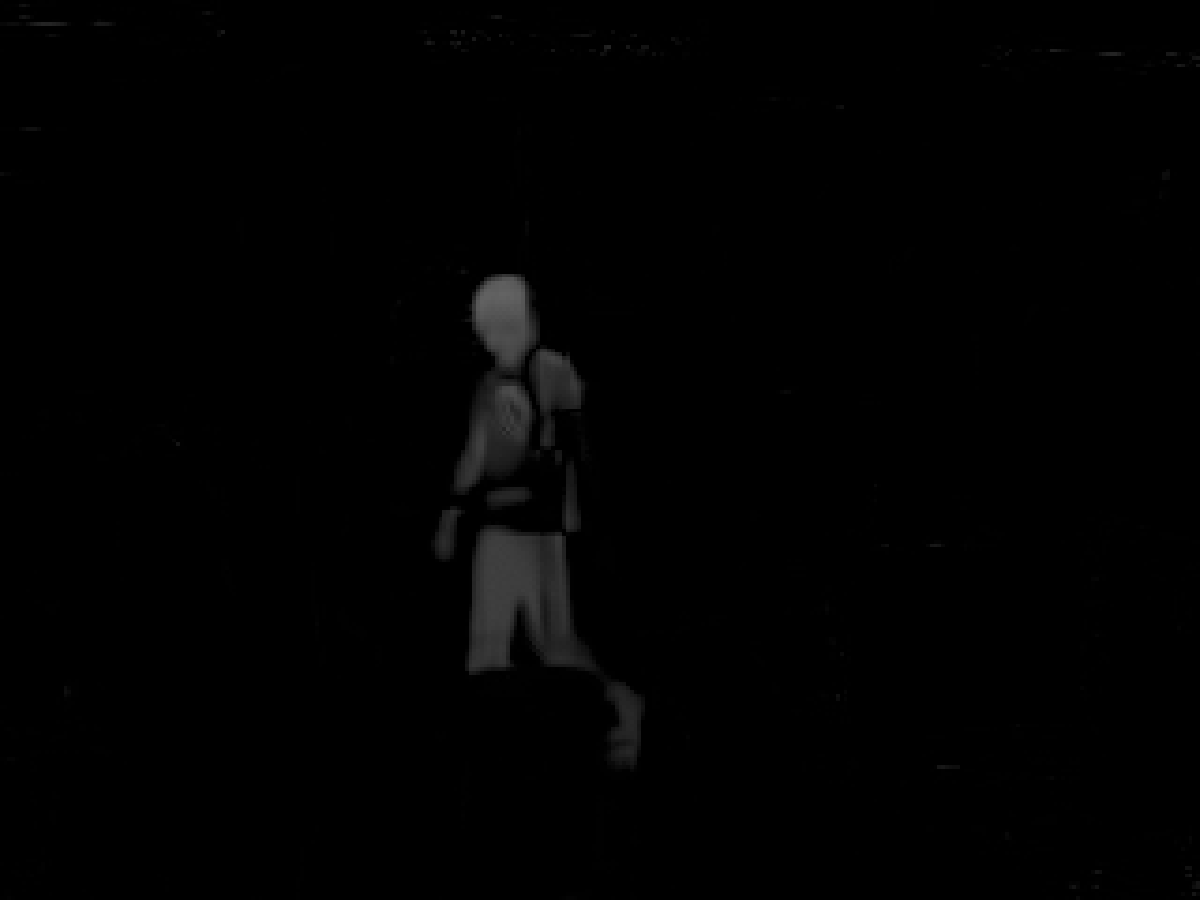}
	\caption{Top row: original images; lower left column: SpaRCS. lower middle column: TMP; lower right column: DTMP. The measurement rate is set to 0.1. Video dataset: ``CameraParameter''.}\label{camera_01_offline}
\end{figure}

In Fig. \ref{twoleave_01_offline} and Fig. \ref{camera_01_offline}, we compare the recovery of offline DTMP, TMP, and SpaRCS on the video sequences ``Twoleaveshop2cor'' and ``CameraParameter'', respectively. It is clear from these figures that the recovery background and the foreground of the offline DTMP algorithm demonstrates a better visual quality.

\begin{figure}[!ht]
	\centering
	\includegraphics[width=\linewidth]{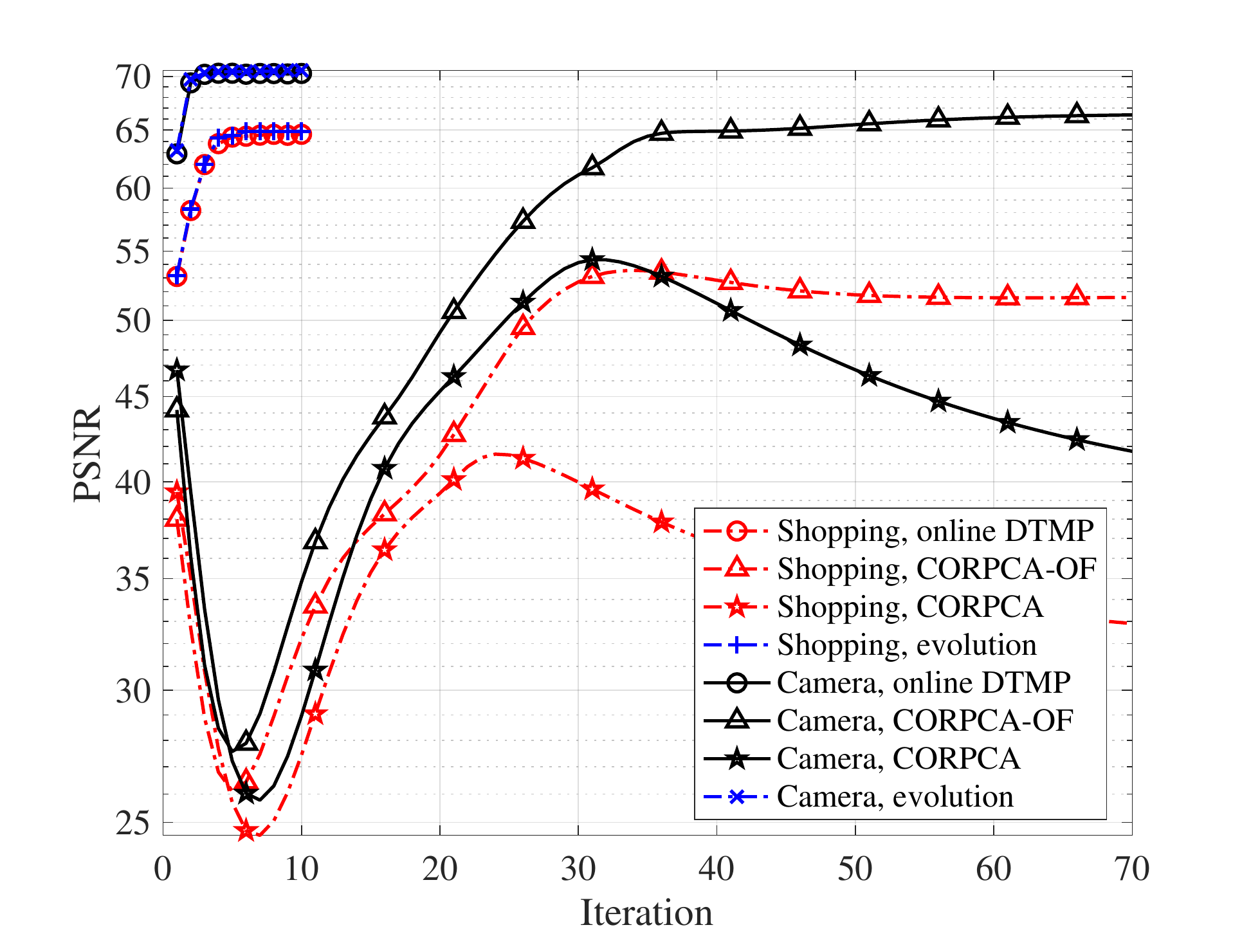}
	\caption{The PSNR performance of the online DTMP algorithm, the CORPCA algorithm and the CORPCA-OF algorithm on the $100$-th frame of video datasets ``CameraParameter'' and the $550$-th frame of ``ShoppingMall'' and the state evolution of the online DTMP algorithm. The measurement rate is set to 0.3.}\label{on_se_comp}
\end{figure}

\subsection{Online Compressed VBS}
In the following, we compare the performance of the online DTMP algorithm with the counterpart algorithms including CORPCA \cite{van2018compressive} and CORPCA-OF \cite{prativadibhayankaram2017compressive} for the online compressed VBS problem and characterize the performance of the online DTMP by its state evolution analysis. For all the online algorithms, we choose the first 70 frames of the video sequence, compress the video matrix, and then recover the background and foreground matrices using offline DTMP. The last 10 columns of the recovered background matrix and the last 3 columns of the foreground matrix are taken as $\L_t^{pre}$ and $\S_t^{pre}$ respectively in Algorithm \ref{online_dtmp} and as the input of in CORPCA and CORPCA-OF. The rest frames are measured frame by frame and recovered using the online separation algorithms whenever a compressed frame arrives.

In Fig. \ref{on_se_comp}, we plot the PSNR of online DTMP, CORPCA, and CORPCA-OF against the iteration number on video dataset ``CameraParameter'' and ``ShoppingMall''. In the figure, the curves with legend ``Camera'' correspond to video ``CameraParameter'', and those with legend ``Shopping'' correspond to ``ShoppingMall''. From the figure, we see that the online DTMP algorithm achieves a much higher PSNR  with less iteration time comparing with CORPCA and CORPCA-OF in both datasets, and the state evolution analysis accurately characterize the performance of online DTMP on both videos.

In Table \ref{table-time}, we present the recovery speed of different algorithms on different video sequences (11 video sequences from dataset "CDnet2014", on for each video category.) with different measurement rates. The recovery speed is expressed in frames per second (FPS). We see from the table that the online DTMP algorithm have the highest recovery speed on all cases (about 3-10 times faster than CORPCA and CORPCA-OF).

\begin{figure}
	\centering
	\includegraphics[width=0.24\linewidth]{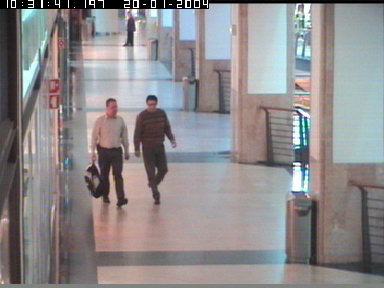}
	\includegraphics[width=0.24\linewidth]{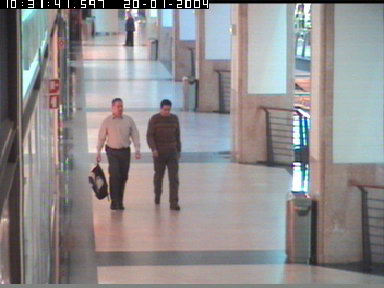}
	\includegraphics[width=0.24\linewidth]{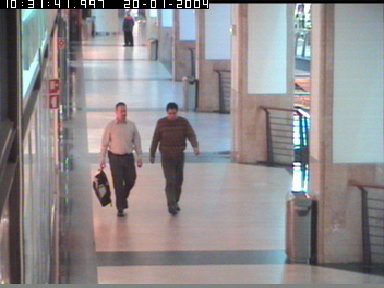}
	\includegraphics[width=0.24\linewidth]{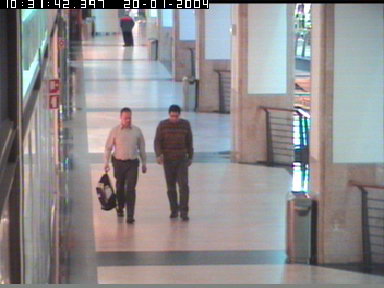}
	\includegraphics[width=0.24\linewidth]{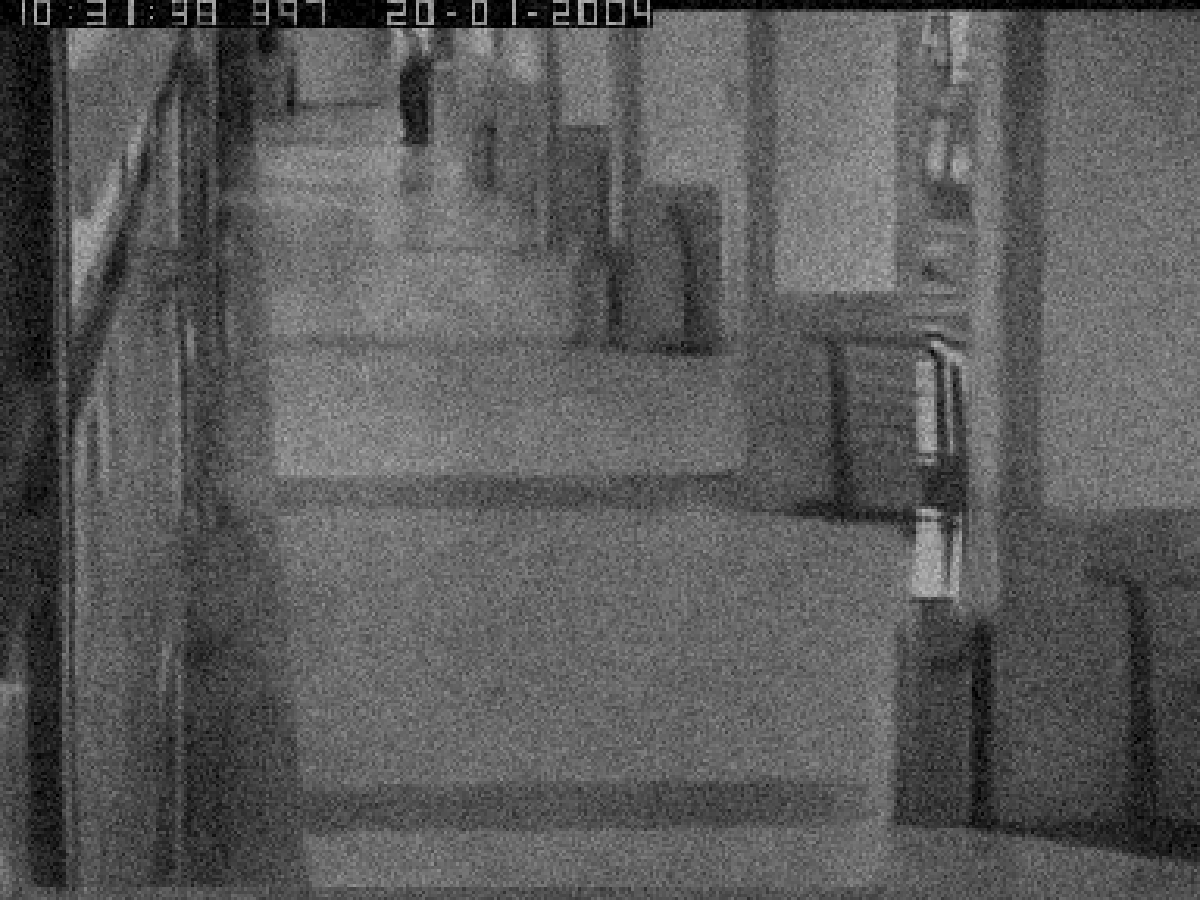}
	\includegraphics[width=0.24\linewidth]{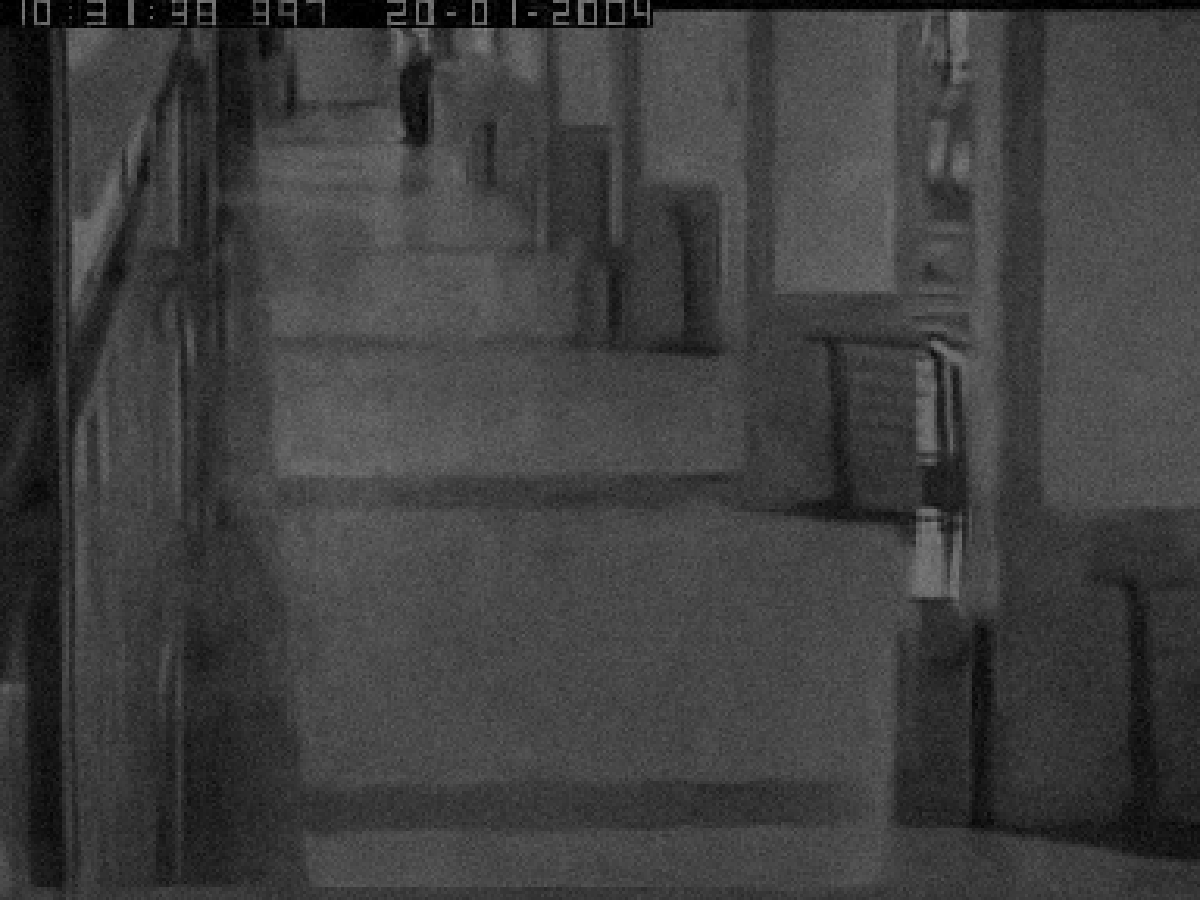}
	\includegraphics[width=0.24\linewidth]{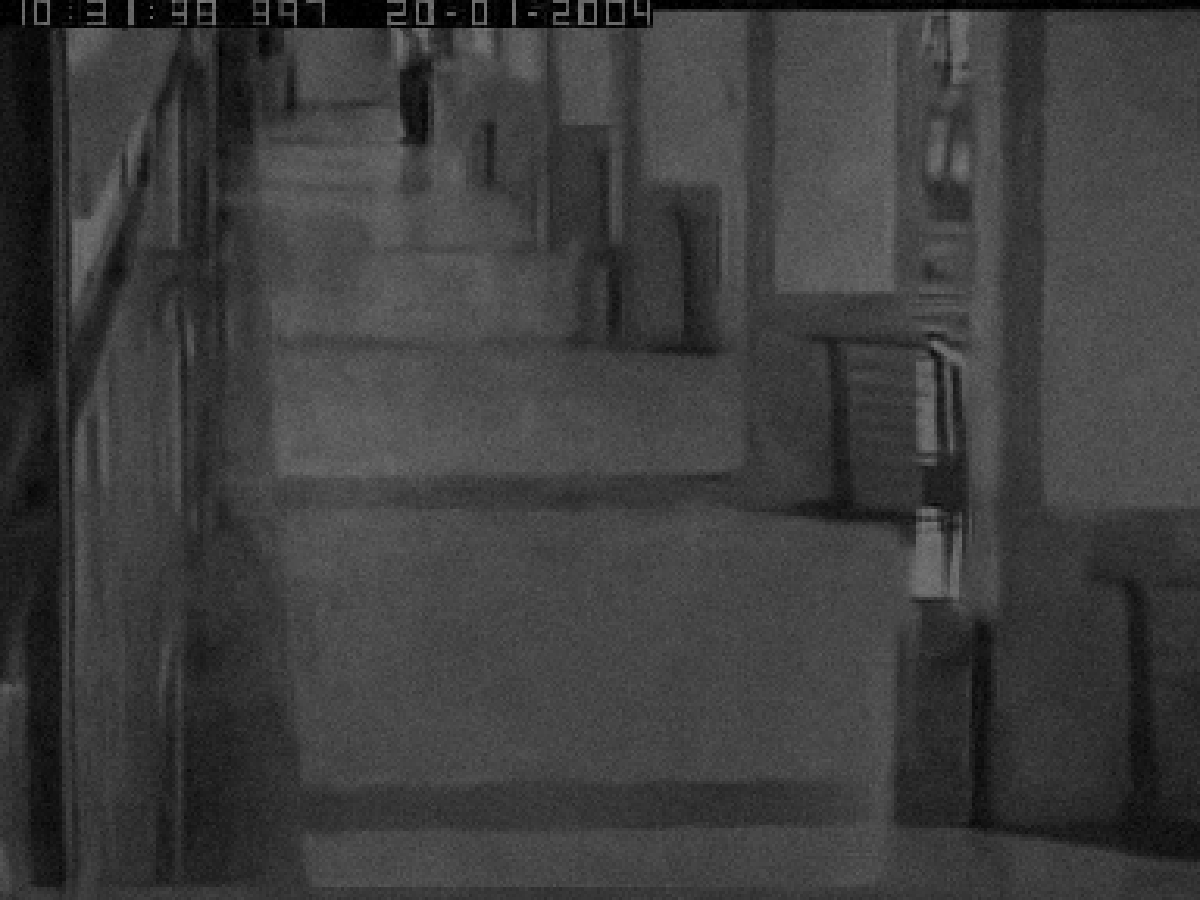}
	\includegraphics[width=0.24\linewidth]{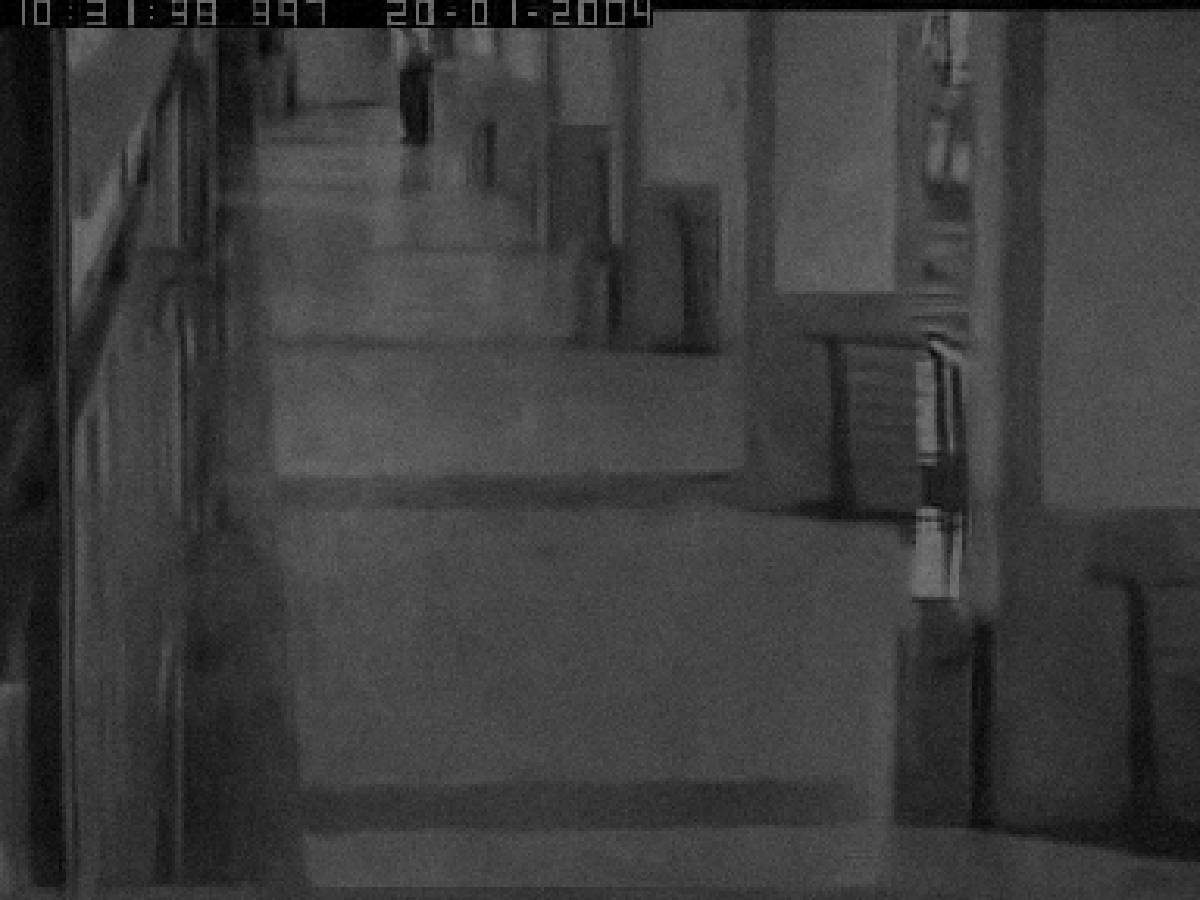}
	\includegraphics[width=0.24\linewidth]{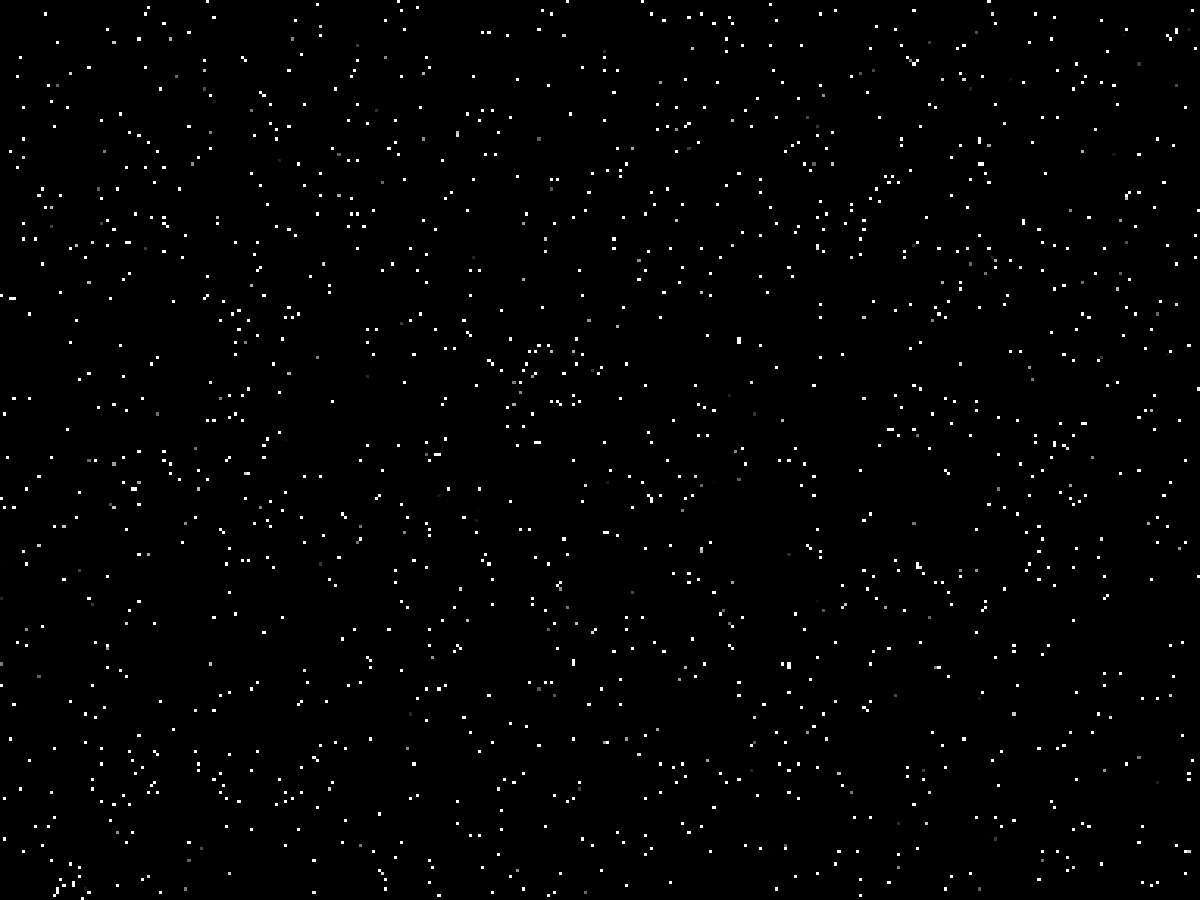}
	\includegraphics[width=0.24\linewidth]{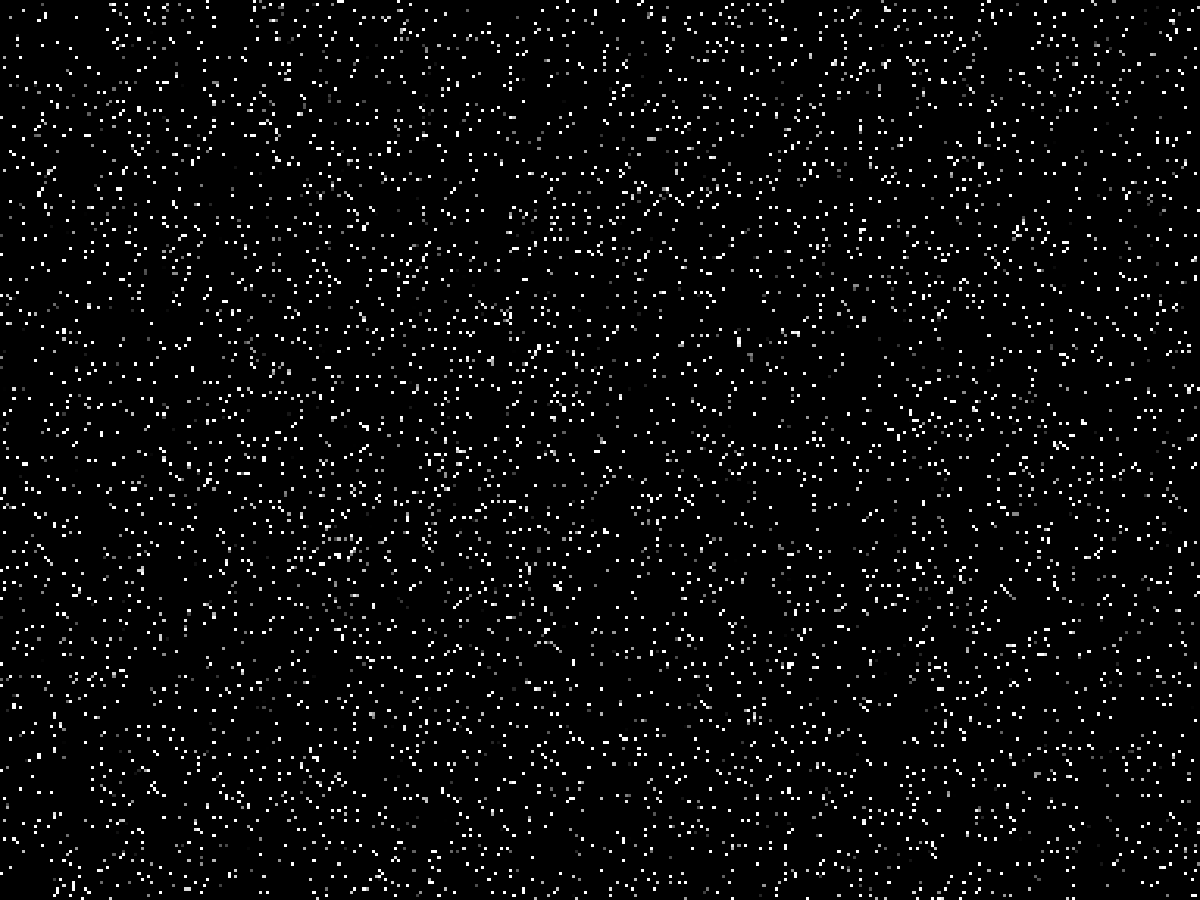}
	\includegraphics[width=0.24\linewidth]{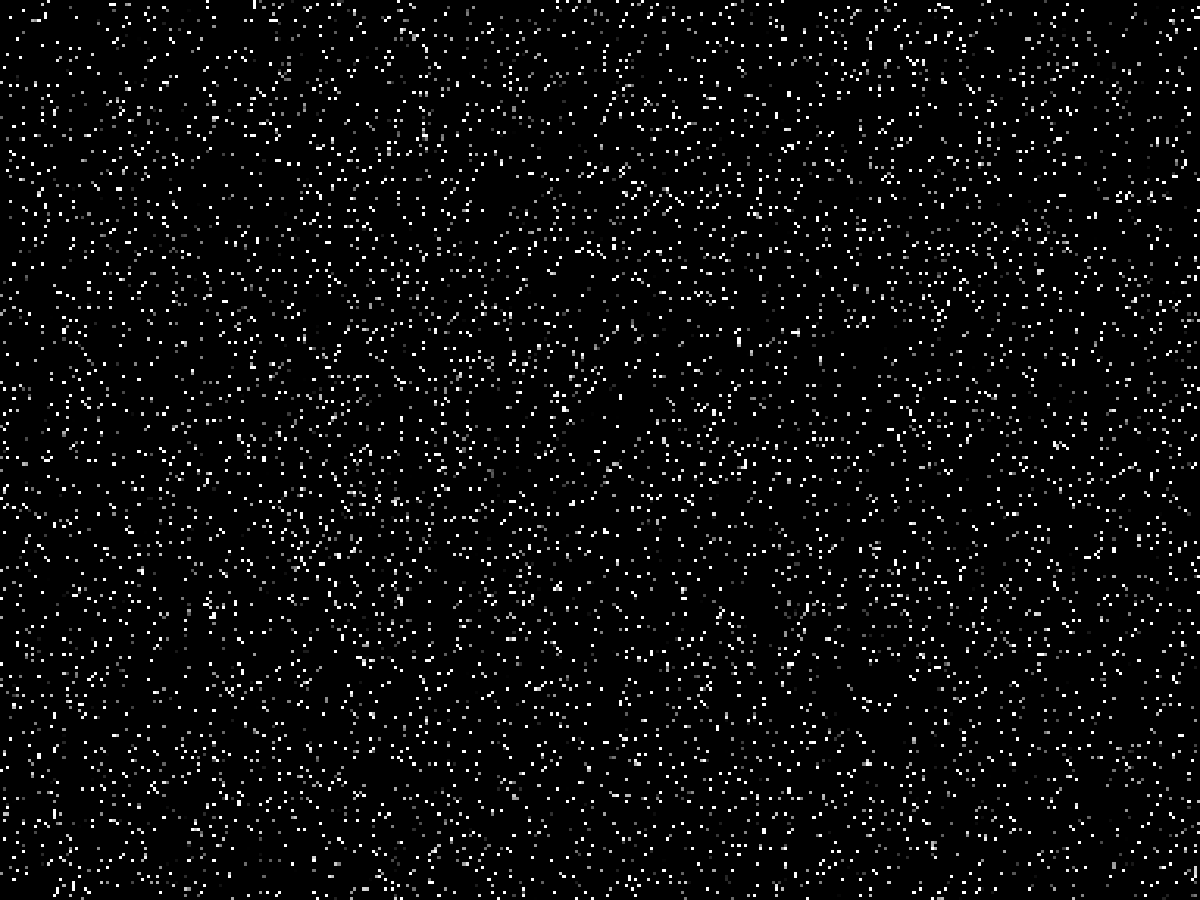}
	\includegraphics[width=0.24\linewidth]{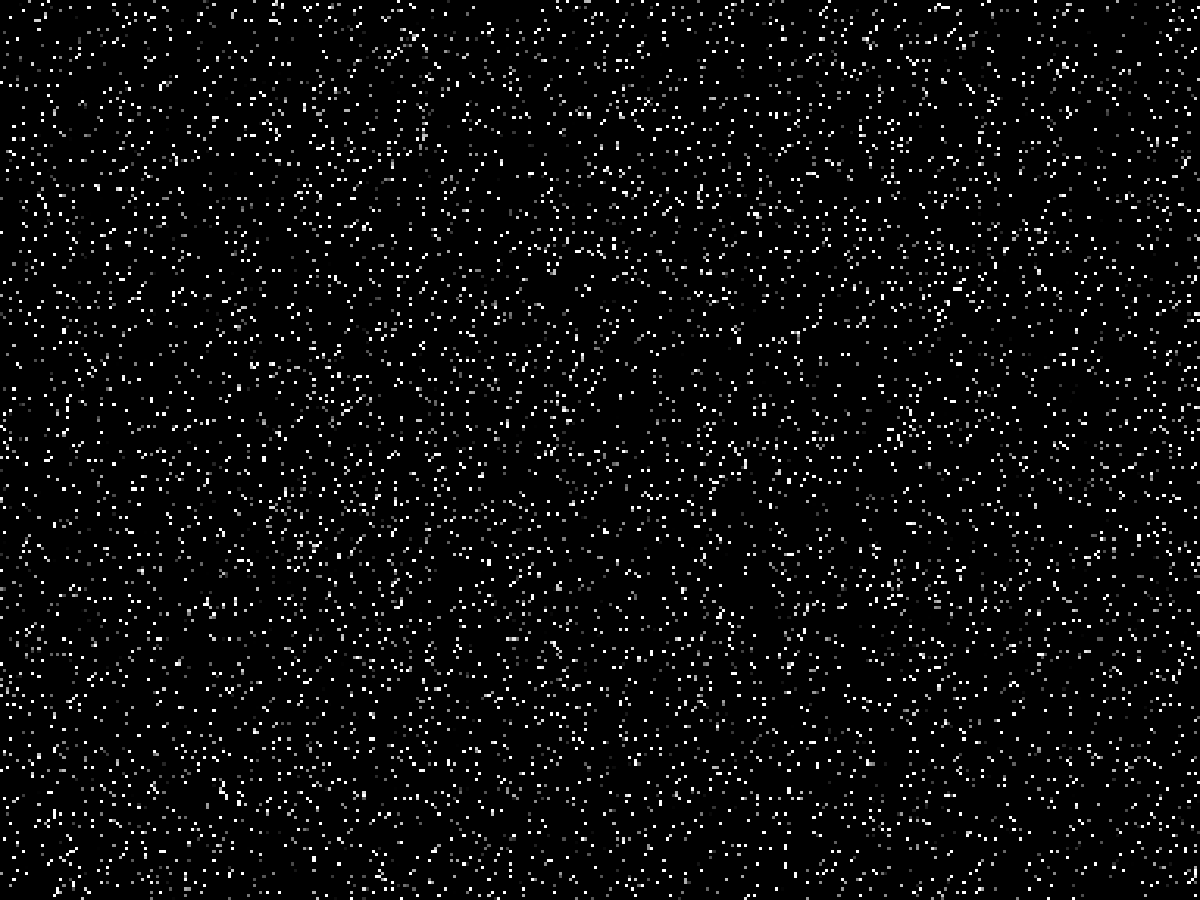}

	\includegraphics[width=0.24\linewidth]{./figs/corpca_back_251.png}
	\includegraphics[width=0.24\linewidth]{./figs/corpca_back_261.png}
	\includegraphics[width=0.24\linewidth]{./figs/corpca_back_271.png}
	\includegraphics[width=0.24\linewidth]{./figs/corpca_back_281.png}
	\includegraphics[width=0.24\linewidth]{./figs/corpca_fore_251.png}
	\includegraphics[width=0.24\linewidth]{./figs/corpca_fore_261.png}
	\includegraphics[width=0.24\linewidth]{./figs/corpca_fore_271.png}
	\includegraphics[width=0.24\linewidth]{./figs/corpca_fore_281.png}

	\includegraphics[width=0.24\linewidth]{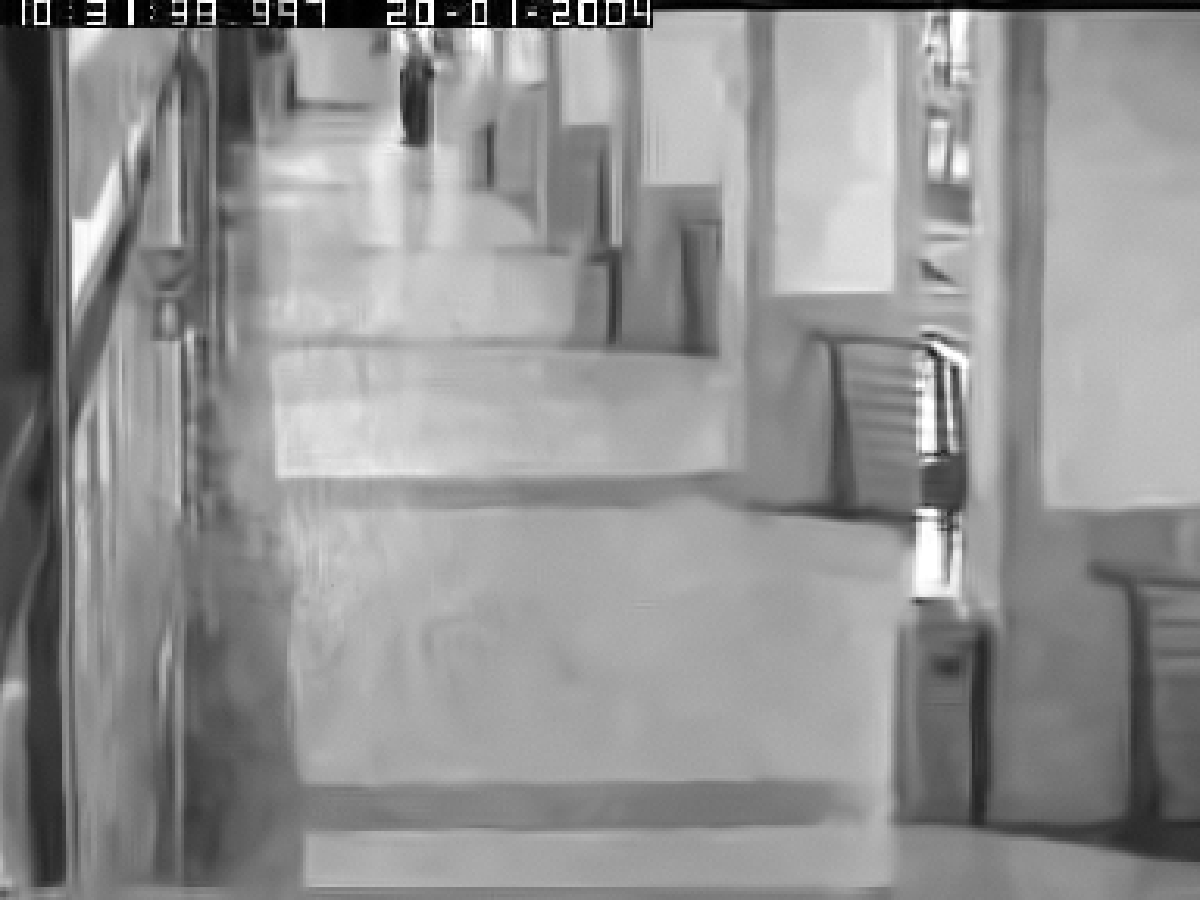}
	\includegraphics[width=0.24\linewidth]{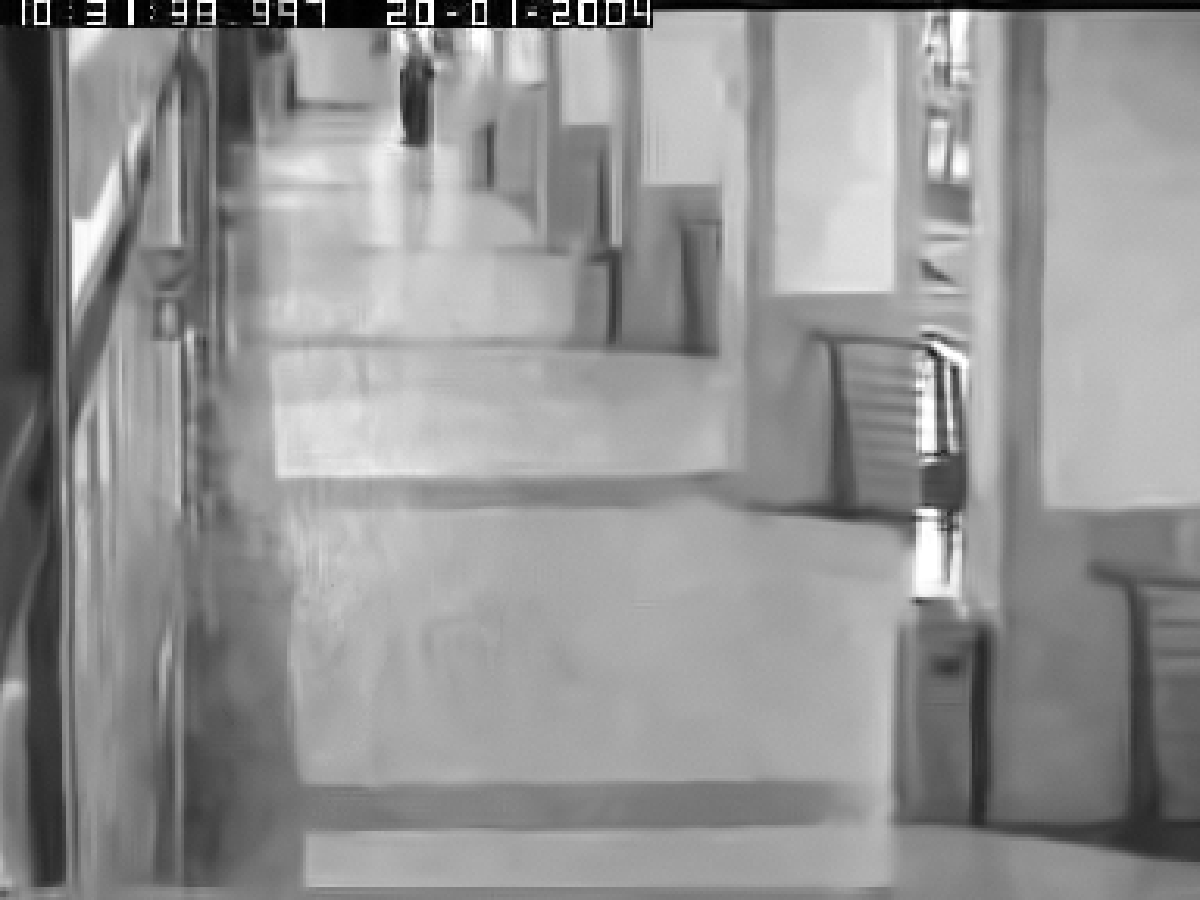}
	\includegraphics[width=0.24\linewidth]{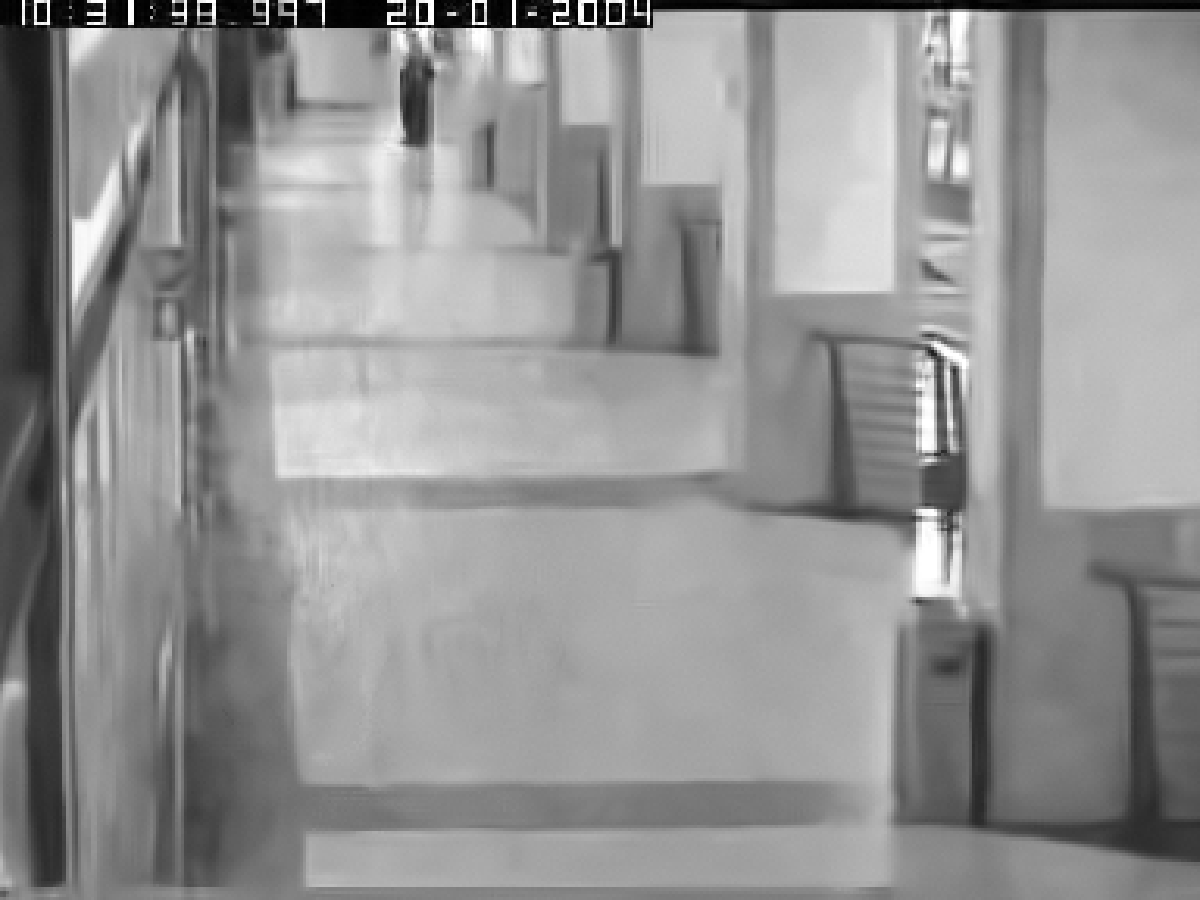}
	\includegraphics[width=0.24\linewidth]{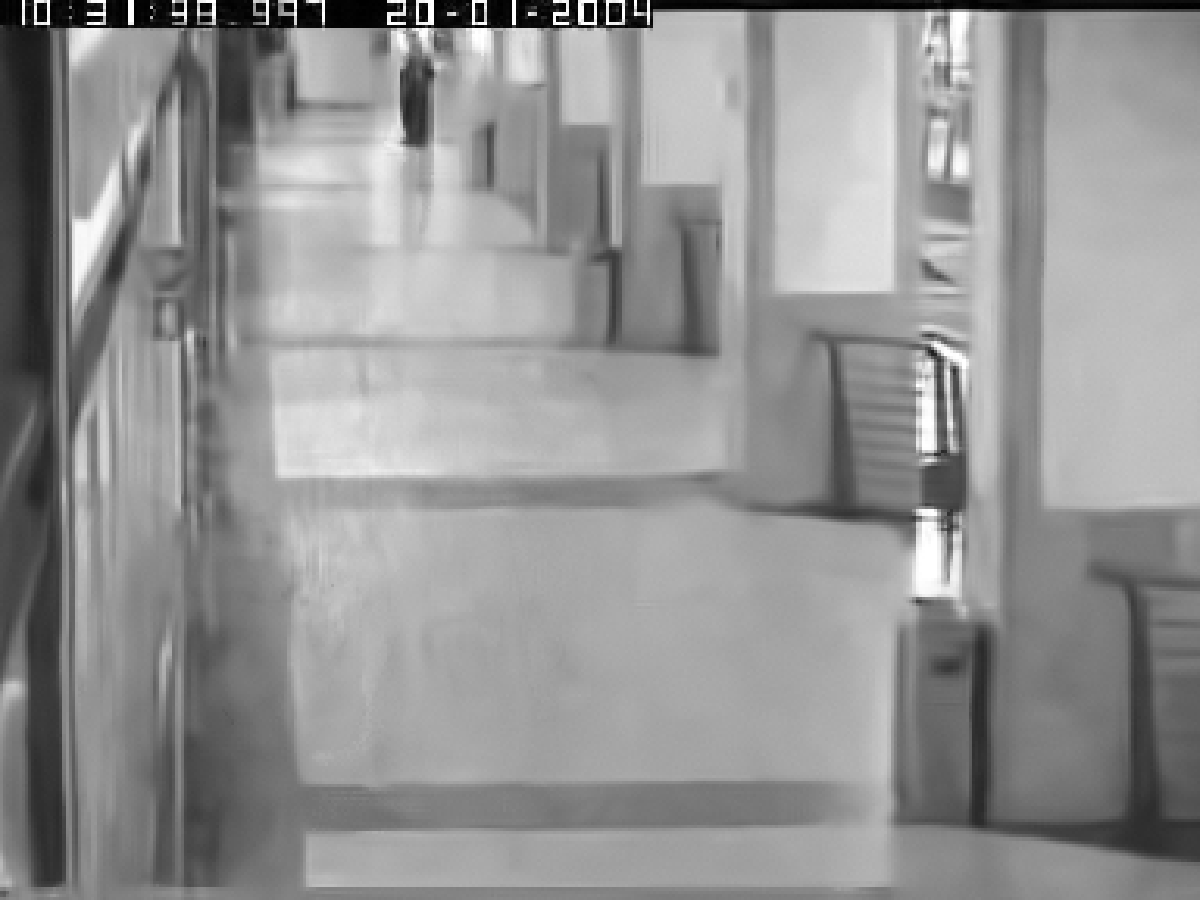}
	\includegraphics[width=0.24\linewidth]{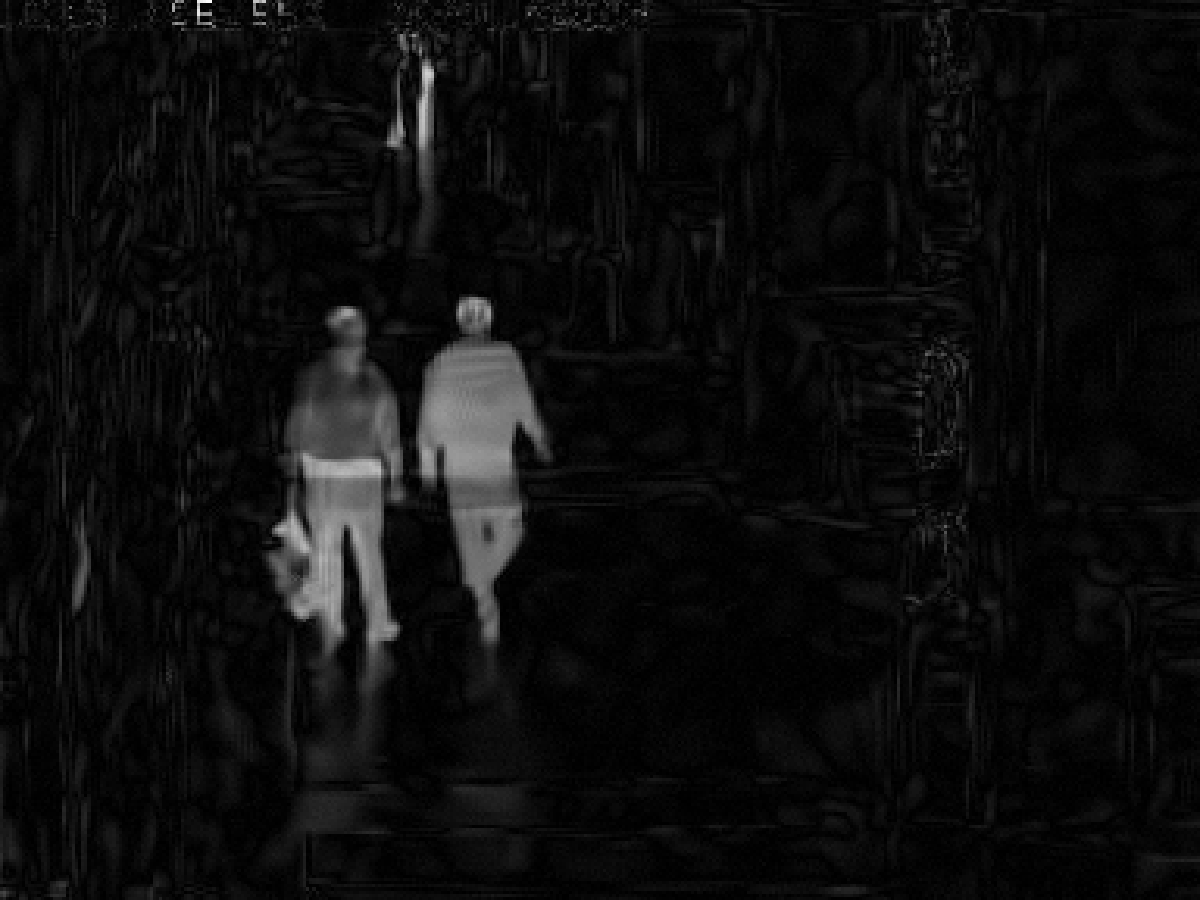}
	\includegraphics[width=0.24\linewidth]{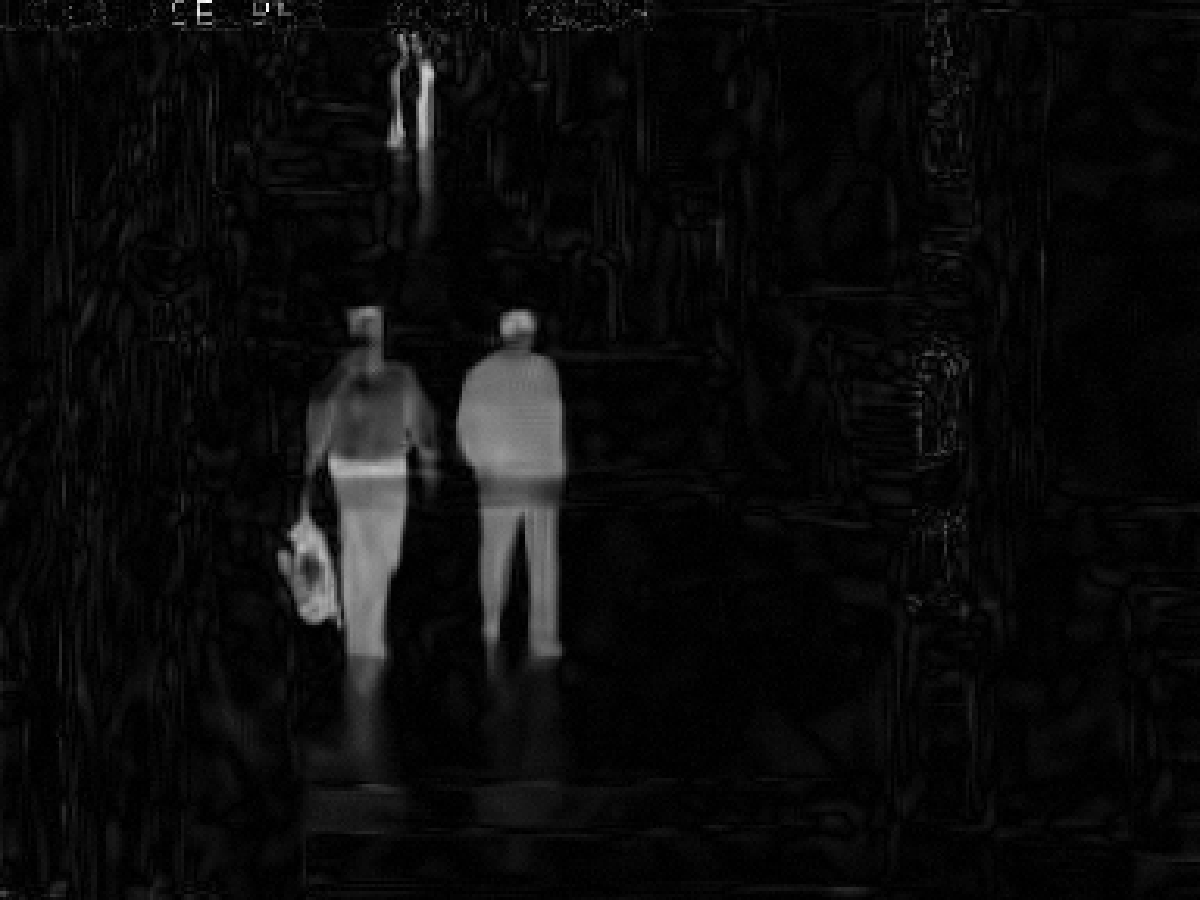}
	\includegraphics[width=0.24\linewidth]{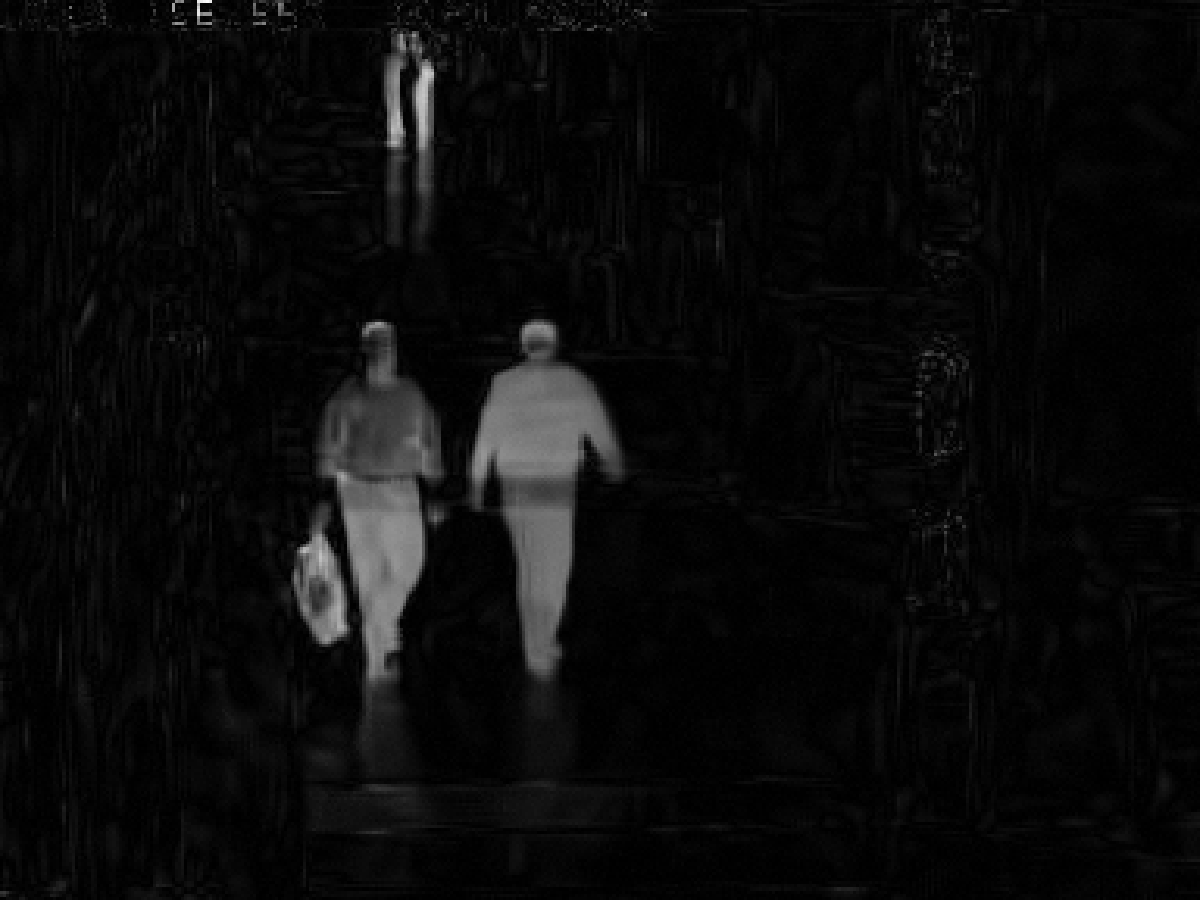}
	\includegraphics[width=0.24\linewidth]{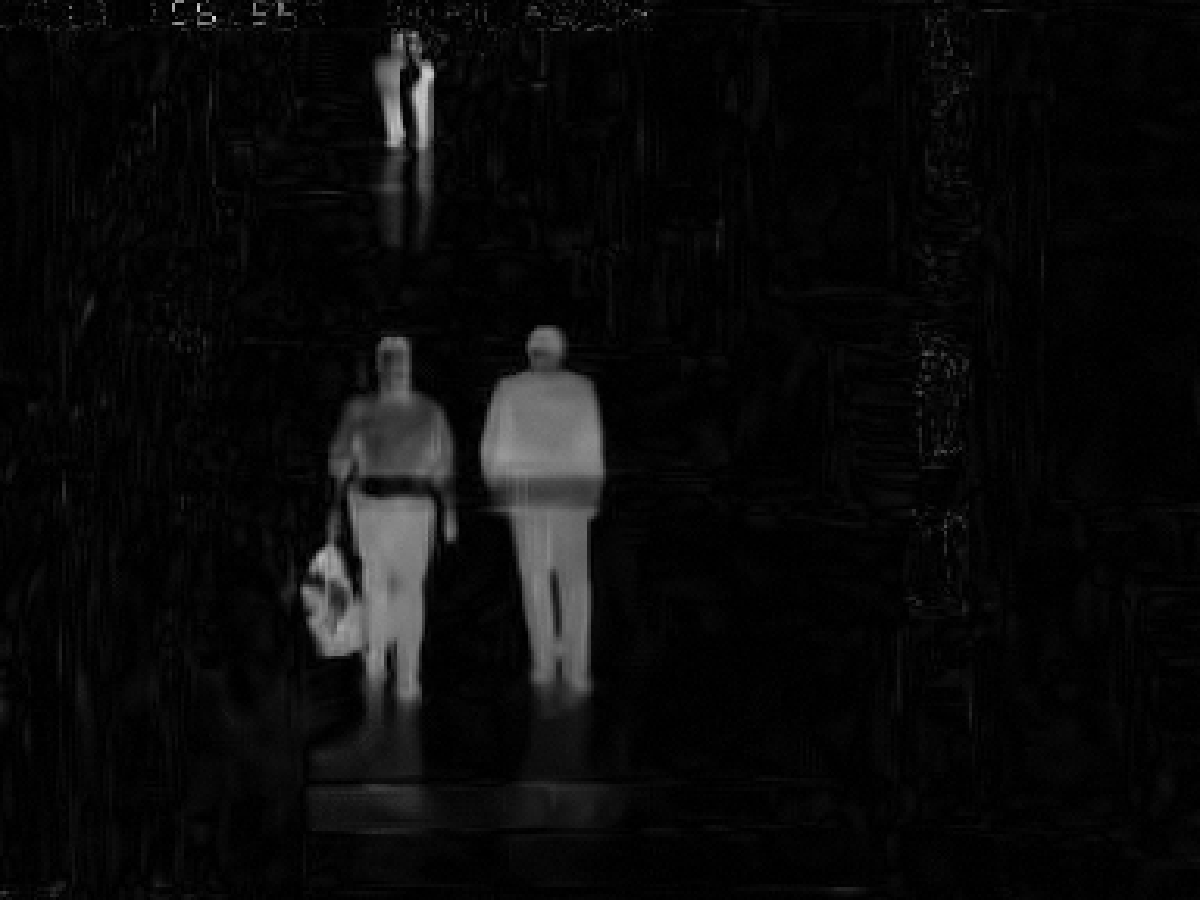}
	
	\caption{Recovery results of the online DTMP algorithm, the CORPCA algorithm, and the CORPCA-OF algorithm on video dataset ``Twoleaveshop2cor''. The measurement rate is 0.1.}\label{online-twoleave}
\end{figure}

In Fig. \ref{online-twoleave}, we compare the recovery results of online DTMP, CORPCA, and CORPCA-OF on a video sequence. In the figure, the images in the first row are the original video frames, these in the second and third rows are the recovered background and foreground frames using CORPCA, these in the fourth and fifth rows are the recovered background and foreground frames using CORPCA-OF, and these in the sixth and seventh rows are recovered background and foreground frames using online DTMP. As shown in this figure, CORPCA and CORPCA-OF fail to recover the foreground frames when the measurement is low, whereas the online DTMP algorithm succeeds in the recovery and demonstrates a much better visual quality.

\begin{table}[]
\centering
\setlength\tabcolsep{2pt}
\caption{Running time comparison (in FPS) of different algorithms for online compressed VBS.}
\begin{tabular}{l|l|l|l|l}
\hline
video &measurement rate &Online DTMP & CORPCA & CORPCA-OF \\ \hline
\multirow{3}{*}{tramcrossroad}&$m/n=0.5$   &  3.98 & 0.563 &  0.394         \\ 
&$m/n=0.3$   & 3.96 & 0.562 &  0.392         \\ 
&$m/n=0.1$   & 3.44 & 0.565 &  0.393 \\ \hline
\multirow{3}{*}{winterstreet} &$m/n=0.5$   & 1.103 & 0.162 &   0.116        \\
&$m/n=0.3$    & 1.104 & 0.163 & 0.118       \\ 
&$m/n=0.1$      & 1.06 & 0.159 & 0.114 \\ \hline  
\multirow{3}{*}{PET2016}&$m/n=0.5$   & 0.698 & 0.101 & 0.068          \\
&$m/n=0.3$   & 0.684 & 0.102 & 0.069          \\ 
&$m/n=0.1$   & 0.689 & 0.101 & 0.068 \\ \hline 
% ,720,576
\multirow{3}{*}{wetsnow}& $m/n=0.5$  & 0.741 & 0.107 & 0.070          \\
&$m/n=0.3$    & 0.707 & 0.108 & 0.069          \\ 
&$m/n=0.1$    & 0.712 & 0.108 & 0.069 \\ \hline  
\multirow{3}{*}{canoe}&$m/n=0.5$   & 4.05 & 0.589 & 0.388       \\
&$m/n=0.3$    & 3.96 & 0.595 & 0.391      \\ 
&$m/n=0.1$  & 1.32 & 0.591 & 0.390 \\  \hline  
\multirow{3}{*}{intermittentPan}&$m/n=0.5$   & 1.26  &  0.174 & 0.124       \\
&$m/n=0.3$    & 1.25  &  0.176  &  0.126     \\ 
&$m/n=0.1$  & 1.24 & 0.176  &  0.126\\  \hline  

\multirow{3}{*}{diningRoom}&$m/n=0.5$   & 2.518 &  0.3536  & 0.258      \\
&$m/n=0.3$    &2.224   & 0.347  &  0.254     \\ 
&$m/n=0.1$  & 2.234 &   0.3436  &  0.255 \\  \hline  

\multirow{3}{*}{busStation}&$m/n=0.5$   & 2.79  &  0.384  &  0.292      \\
&$m/n=0.3$    &2.68 &   0.389  &  0.298     \\ 
&$m/n=0.1$  & 2.71  &  0.391  &  0.291 \\  \hline 

\multirow{3}{*}{sofa}&$m/n=0.5$   & 3.340 &  0.481  & 0.336  \\
&$m/n=0.3$    & 3.178 & 0.490 &  0.320     \\ 
&$m/n=0.1$  & 3.180 & 0.489 &  0.326 \\  \hline  

\multirow{3}{*}{sidewalk}&$m/n=0.5$   & 3.41 & 0.477 & 0.395       \\
&$m/n=0.3$    & 3.35 &   0.451 &  0.379     \\ 
&$m/n=0.1$  & 3.37 &  0.452 &  0.377 \\  \hline  

\multirow{3}{*}{turbulence1}&$m/n=0.5$   & 0.584 &  0.091 &  0.052       \\
&$m/n=0.3$    & 0.547 &  0.086 &  0.049     \\ 
&$m/n=0.1$  & 0.549 & 0.088 & 0.049 \\  \hline  

\end{tabular}\label{table-time}
\end{table}

\section{Conclusion}\label{sec_conc}
In this paper, we proposed a denoising-based message passing framework termed DTMP for compressed video background and foreground separation in both offline and online measurement modes. DTMP exploits the textual feature of a natural image and the continuity of the foreground in a video to improve the background and foreground separation performance. We also developed the state evolution to accurately characterize the behavior of the proposed algorithms. Numerical results demonstrate that DTMP can separate the background and foreground of the compressed video with a much higher resulting PSNR and exhibits a substantially better visual quality compared to the counterpart algorithms. The work in this paper can be potentially improved in a number of aspects. For example, from Fig. \ref{twoleave_01_offline} \ref{camera_01_offline}, \ref{online-twoleave}, we see that the foreground of an image usually exhibits block sparsity, i.e. the non-zeros appear in clusters. This block sparsity, however, has not been appropriately exploited in our proposed algorithms. In addition, some key parameters of the proposed algorithms, such as the rank of the matrix $\L$ and the noise level $\sigma^2$, are set empirically in simulations. These parameters can be learned automatically by using data-driven machine learning techniques. We will explore these possibilities in our future work.

\bibliographystyle{IEEEtran}

\begin{thebibliography}{10}
\providecommand{\url}[1]{#1}
\csname url@samestyle\endcsname
\providecommand{\newblock}{\relax}
\providecommand{\bibinfo}[2]{#2}
\providecommand{\BIBentrySTDinterwordspacing}{\spaceskip=0pt\relax}
\providecommand{\BIBentryALTinterwordstretchfactor}{4}
\providecommand{\BIBentryALTinterwordspacing}{\spaceskip=\fontdimen2\font plus
\BIBentryALTinterwordstretchfactor\fontdimen3\font minus
  \fontdimen4\font\relax}
\providecommand{\BIBforeignlanguage}[2]{{%
\expandafter\ifx\csname l@#1\endcsname\relax
\typeout{** WARNING: IEEEtran.bst: No hyphenation pattern has been}%
\typeout{** loaded for the language `#1'. Using the pattern for}%
\typeout{** the default language instead.}%
\else
\language=\csname l@#1\endcsname
\fi
#2}}
\providecommand{\BIBdecl}{\relax}
\BIBdecl

\bibitem{shah2014video}
M.~Shah, J.~D. Deng, and B.~J. Woodford, ``Video background modeling: recent approaches, issues and our proposed techniques,'' \emph{Mach. Vision. Appl.}, vol.~25, no.~5, pp. 1105--1119, July, 2014.

\bibitem{bouwmans2019background}
T.~Bouwmans, B.~Garcia, ``Background subtraction in real applications: Challenges, current models and future directions,'' \emph{Comput. Sci. Rev.}, Feb. 2020.

\bibitem{lee2002back}
B.~Lee, and M.~Hedley, ``Background estimation for video surveillance'' \emph{IVCNZ}, Jan. 2002.

\bibitem{mcfarlane1995segmentation}
McFarlane, J.~Nigel, and C.~Schofield, ``Segmentation and tracking of piglets in images,'' \emph{Mach. Vision. Appl.}, vol. 8, no.~3, pp. 187--193, Mar. 1995.

\bibitem{zheng2006extracting}
J. Zheng, Y. Wang, N.~L. Nihan, and M.~E. Hallenbeck, ``Extracting roadway background image: Mode-based approach,'' \emph{Transportation research record}, vol. 1944, no.~1, pp. 82--88, Jan. 2006.

\bibitem{wren1997pfinder}
C.~E. Wren, A. Azarbayejani, T. Darrell, and A.~P. Pentland, ``Pfinder: Real-time tracking of the human body,'' \emph{IEEE Trans. Pattern. Anal. Mach. Intell.}, vol. 19, no.~7, pp. 780--785, July, 1997.

\bibitem{stauffer1999adaptive}
C. Stauffer, and W.~E. Grimson, ``Adaptive background mixture models for real-time tracking,'' in \emph{Proc. IEEE Comput. Soc. Conf. Comput. Vis. Pattern Recognit.}, vol. 2, pp. 246--252, Fort Collins, Colorado, June, 1999.

\bibitem{elgammal2000non}
A.~Elgammal, D.~Harwood, and L.~Davis, ``Non-parametric model for background subtraction,'' in \emph{Proc. ECCV 2020}, pp. 751--767, Dublin, Ireland, June, 2000.

\bibitem{oliver2000bayesian}
N.~M.~Oliver, B.~Rosario, and A.~P.~Pentland, ``A Bayesian computer vision system for modeling human interactions,'' \emph{IEEE Trans. Pattern Anal. Mach. Intell.}, vol. 22. no.~8, pp. 831--843, Aug. 2000.

\bibitem{bouwmans2014robust}
T.~Bouwmans and E.~H. Zahzah, ``Robust {PCA} via principal component pursuit: A review for a comparative evaluation in video surveillance,'' \emph{Comput. Vis. Image Und.}, vol. 122, pp. 22--34, May, 2014.

\bibitem{sigari2008fuzzy}
M.~H.~Sigari, N.~Mozayani, and H.~Pourreza, ``Fuzzy running average and fuzzy background subtraction: concepts and application,'' \emph{Int. J. Comput. Sci. Netw. Secur.}, vol.~8, no.~2, pp. 138--143, Feb. 2008.

\bibitem{el2008type}
F.~El Baf, T.~Ouwmans, and B.~Vachon, ``Type-2 fuzzy mixture of Gaussians model: application to background modeling,'' in \emph{Int. Symp. Vis. Comput.}, pp. 772--781, Las Vegas, NV, USA, July, 2008.

\bibitem{zhang2006fusing}
H.~Zhang, and D.~Xu, ``Fusing color and texture features for background model,'' in \emph{Proc. FSKD 2006}, pp. 887--893, Xian, China, Sep. 2006.

\bibitem{el2008fuzzy}
F.~Baf, T.~Bouwmans, and B.~Vachon, ``Fuzzy integral for moving object detection'' in \emph{Proc. FUZZ-IEEE 2008}, pp.~1729-1736, Hong-Kong, China, June, 2008.

\bibitem{bouwmans2019deep}
T.~Bouwmans, and S.~Javed, M.~Sultana, and S.~Jung, ``Deep neural network concepts for background subtraction: A systematic review and comparative evaluation'' in \emph{Neural Networks}, vol.~117, pp.~8-66, Sep. 2019.

\bibitem{candes2011robust}
E.~J. Cand{\`e}s, X.~Li, Y.~Ma, and J.~Wright, ``Robust principal component analysis?'' \emph{J. ACM}, vol.~58, no.~3, pp.~1--37, May, 2011.

\bibitem{xu2010robust}
H.~Xu, C.~Caramanis, and S.~Sanghavi, ``Robust pca via outlier pursuit,'' in \emph{Proc. Adv. Neural Inf. Process. Syst.}, Vancouver, CANADA, Dec. 2010, pp. 2496--2504.

\bibitem{ding2011bayesian}
X.~Ding, L.~He, and L.~Carin, ``Bayesian robust principal component analysis,'' \emph{IEEE Trans. Image Process.}, vol.~20, no.~12, pp. 3419--3430, May, 2011.

\bibitem{donoho2006compressed}
D.~L. Donoho \emph{et~al.}, ``Compressed sensing,'' \emph{IEEE Trans. Inform. Theory}, vol.~52, no.~4, pp. 1289--1306, Apr. 2006.

\bibitem{candes2009exact}
E.~J. Cand{\`e}s and B.~Recht, ``Exact matrix completion via convex optimization,'' \emph{Found. Comput. Math.}, vol.~9, no.~6, p. 717, Apr. 2009.

\bibitem{ganesh2012principal}
A.~Ganesh, K.~Min, J.~Wright, and Y.~Ma, ``Principal component pursuit with reduced linear measurements,'' in \emph{Proc. IEEE Int. Symp. Info. Theory}, Cambridge, MA, USA, July, 2012, pp. 1281--1285.

\bibitem{waters2011sparcs}
A.~E. Waters, A.~C. Sankaranarayanan, and R.~Baraniuk, ``Sparcs: Recovering low-rank and sparse matrices from compressive measurements,'' in \emph{Proc. Adv. Neural Inf. Process. Syst.}, Granada, SPAIN, Dec. 2011, pp. 1089--1097.

\bibitem{aravkin2014variational}
A.~Aravkin, S.~Becker, V.~Cevher, and P.~Olsen, ``A variational approach to stable principal component pursuit,'' \emph{arXiv preprint arXiv:1406.1089}, 2014.


\bibitem{xue2018turbo}
Z.~Xue, X.~Yuan, and Y.~Yang, ``Turbo-type message passing algorithms for compressed robust principal component analysis,'' \emph{IEEE J. Sel. Top. Signa. Process.}, vol.~12, no.~6, pp.~1182--1196, Dec. 2018.

\bibitem{van2018compressive}
H.~Van~Luong, N.~Deligiannis, J.~Seiler, S.~Forchhammer, and A.~Kaup,
  ``Compressive online robust principal component analysis via n-$l_1$ minimization,'' \emph{IEEE Trans. Image Process.}, vol.~27, no.~9, pp. 4314--4329, Sep. 2018.


\bibitem{prativadibhayankaram2017compressive}
S.~Prativadibhayankaram, H.~Van~Luong, T.~H. Le, and A.~Kaup, ``Compressive online robust principal component analysis with optical flow for video foreground-background separation,'' in \emph{Proc. Int. Symp. Info. Comm. Tech.}, Nha Trang, Vietnam, 2017, Dec. pp. 385--392.

\bibitem{xue2017denoising}
Z.~Xue, J.~Ma, and X.~Yuan, ``Denoising-based turbo compressed sensing,'' \emph{IEEE Access}, vol.~5, pp. 7193--7204, Apr. 2017.


% ////////////////////////
\bibitem{culibrk2006neural}
D.~Culibrk, O.~Marques, D.~Socek, H.~Kalva, and B.~Furht, ``A neural network approach to bayesian background modeling for video object segmentation.'' \emph{IEEE Trans. Neural Netw.}, vol.~18, no.~6, pp.~1614--1627, Nov. 2007.

\bibitem{luque2008video}
R.~M.~Luque, D.~L{\'o}pez-Rodr{\'\i}guez, E.~Merida-Casermeiro, and E.~J~Palomo, ``Video object segmentation with multivalued neural networks,'' in \emph{Proc. Eighth Int. Conf. Hybrid Intell. Syst.}, pp.~613--618, Barcelona, Spain, Sept. 2008.

\bibitem{luque2008neural}
R.~M.~Luque, E.~Dom{\'\i}nguez, E.~J~Palomo, and J.~Mu{\~n}oz, ``A neural network approach for video object segmentation in traffic surveillance,'' in \emph{Proc. Int. Conf. Image Anal. Recogn.}, pp.~151--158, Barcelona, Spain, Sept. 2008.


\bibitem{butler2005real}
D.~E Butler, M.~Bove, and S.~Sridharan, ``Real-time adaptive foreground/background segmentation,'' \emph{EURASIP J. Adv. Signal Process.}, no.~14, vol.~2005, pp.~841926, Jan. 2005.

\bibitem{kim2004background}
K.~Kim, T.~Chalidabhongse, D.~Harwood, and L.~Davis, ``Background modeling and subtraction by codebook construction,'' in \emph{Proc. Int. Conf. Image Process.}, pp.~3061--3064, Singapore, Singapore, Oct. 2004.

\bibitem{xiao2006background}
M.~Xiao, C.~Han, and X.~Kang, ``A background reconstruction for dynamic scenes,'' in \emph{Proc. Int. Conf. Info. Fusion}, pp.~1--7, Florence, Italy, July. 2006.

\bibitem{bouwmans2011recent}
T.~Bouwmans, ``Recent advanced statistical background modeling for foreground detection - a systematic survey,'' \emph{Recent Patents Compt. Sci.}, vol.~4, no.~3, pp. 147--176, Mar. 2011.

\bibitem{bouwmans2014traditional}
T.~Bouwmans, ``Traditional and recent approaches in background modeling for foreground detection: An overview,'' \emph{Comput. Sci. Rev.}, vol.~11, no.~3, pp. 31--66, Mar. 2014.



\bibitem{zonoobi2013low}
D.~Zonoobi, Dornoosh, and A.~Kassim, ``Low rank and sparse matrix reconstruction with partial support knowledge for surveillance video processing,'' in \emph{Proc. Int. Conf. Image Process.}, pp.~335--339, Melbourne, VIC, Australia, Sep. 2013.

\bibitem{li2014recursive}
S.~Li, and H.~Qi, ``Recursive low-rank and sparse recovery of surveillance video using compressed sensing,'' in \emph{Proc. Int. Conf. Distributed Smart Cameras}, pp.~1--6, Venezia Mestre, Italy, Nov. 2014.

\bibitem{pan2017online}
P.~Pan, J.~Feng, L.~Chen, and Y.~Yang, ``Online compressed robust PCA,'' in \emph{Proc. Int. Joint Conf. Neural Networks (IJCNN)}, pp.~1041--1048, Anchorage, AK, USA, July, 2017.

\bibitem{van2018online}
H.~Van Luong, N.~Deligiannis, S.~Forchhammer, and A.~Kaup, ``Online decomposition of compressive streaming data using n-l1 cluster-weighted minimization,'' in \emph{Proc. Data Compression}, pp.~62--69, Snowbird, UT, USA, July, 2018.


\bibitem{van2018compressivecluster}
H.~Van Luong, N.~Deligiannis, S.~Forchhammer, and A.~Kaup, ``Compressive online decomposition of dynamic signals via n-l1 minimization with clustered priors,'' in \emph{Proc. Stat. Signal Process. Workshop}, pp.~846--850, Freiburg, Germany, Aug. 2018.

\bibitem{prativadibhayankaram2018compressive}
S.~Prativadibhayankaram, H.~Van Luong, T. Ha Le, and A.~Kaup, ``Compressive online video background--foreground separation using multiple prior information and optical flow,'' \emph{J. Imaging}, Vol.~4, no.~7, pp.~90--113, May, 2018.

\bibitem{kang2015robust}
B.~Kang, and W.~Zhu, ``Robust moving object detection using compressed sensing,'' \emph{Proc. IET Image Process.}, Vol.~9, no.~9, pp.~811--819, Aug. 2015.

\bibitem{rangan2011generalized}
S.~Rangan, ``Generalized approximate message passing for estimation with random linear mixing,'' in \emph{Proc. IEEE Int. Symp. Inform. Theory}, St. Petersburg, Russia, pp.~2168--2172, Aug. 2011.

\bibitem{parker2014bilinear}
J.~Parker, P.~Schniter, and C.~Volkan, ``Bilinear generalized approximate message passing—Part I: Derivation,'' \emph{IEEE Trans. Signal Process.}, vol.~62, no.~22, pp.~5839--5853, Sep. 2014.

\bibitem{parker2014bilinear2}
J.~Parker, P.~Schniter, and C.~Volkan, ``Bilinear generalized approximate message passing—Part II: Applications,'' \emph{IEEE Trans. Signal Process.}, vol.~62, no.~22, pp.~5854--5867, Sep. 2014.

\bibitem{parker2016parametric}
J.~Parker, P.~Schniter,  ``Parametric bilinear generalized approximate message passing'' \emph{IEEE J. Sel. Top. Signal Process.}, vol.~10, no.~4, pp.~795--808, Mar. 2016.



% \bibitem{luque2008dipolar}
% R.~M.~Luque, D.~L{\'o}pez-Rodr{\'\i}guez, E.~Dominguez, and E.~J~Palomo, ``A dipolar competitive neural network for video segmentation,'' in \emph{Proc. Ibero-American Conf. Artifi. Intell.}, pp.~103--112, Lisbon, Portugal, Oct. 2008.

% //////////
\bibitem{xue2019tarm}
Z.~Xue, X.~Yuan, J.~Ma, and Y.~Ma, ``{TARM}: A turbo-type algorithm for affine rank minimization,'' \emph{IEEE Trans. Signal Process.}, vol.~67, no.~22, pp. 5730--5745, Oct. 2019.

\bibitem{donoho2010message}
D.~L. Donoho, A.~Maleki, and A.~Montanari, ``Message passing algorithms for compressed sensing: I. motivation and construction,'' in \emph{Proc. IEEE ITW 2010}, Cairo, Jan. 2010, pp. 1--5.

\bibitem{ma2015turbo}
J.~Ma, X.~Yuan, and L.~Ping, ``Turbo compressed sensing with partial dft
  sensing matrix,'' \emph{IEEE Signal Proc. Let.}, vol.~22, no.~2, pp. 158--161, Aug. 2015.

\bibitem{kschischang2001factor}
F.~R. Kschischang, B.~J. Frey, H.-A. Loeliger \emph{et~al.}, ``Factor graphs and the sum-product algorithm,'' \emph{IEEE Trans. Inform. Theory}, vol.~47, no.~2, pp. 498--519, Feb. 2001.

\bibitem{chang2000adaptive}
S.~G. Chang, B.~Yu, and M.~Vetterli, ``Adaptive wavelet thresholding for image denoising and compression,'' \emph{IEEE Trans. Image Process.}, vol.~9, no.~9, pp. 1532--1546, Sep. 2000.

\bibitem{yaroslavsky2001transform}
L.~P. Yaroslavsky, K.~O. Egiazarian, and J.~T. Astola, ``Transform domain image restoration methods: review, comparison, and interpretation,'' in \emph{Proc. Soc. Photo-opt. Ins.}, pp. 155--170, Jan. 2001.

\bibitem{dabov2006image}
K.~Dabov, A.~Foi, V.~Katkovnik, and K.~Egiazarian, ``Image denoising with block-matching and 3d filtering,'' in \emph{Proc. Soc. Photo-opt. Ins.}, vol. 6064. pp.~606414, 2006.

\bibitem{cai2010singular}
J.-F. Cai, E.~J. Cand{\`e}s, and Z.~Shen, ``A singular value thresholding
  algorithm for matrix completion,'' \emph{SIAM J. Optimiz.} vol.~20, no.~4, pp. 1956--1982, Mar. 2010.

\bibitem{eckart1936approximation}
C.~Eckart and G.~Young, ``The approximation of one matrix by another of lower rank,'' \emph{Psychometrika}, vol.~1, no.~3, pp. 211--218, Sep. 1936.

\bibitem{donoho1995noising}
D.~L. Donoho, ``De-noising by soft-thresholding,'' \emph{IEEE Trans. Inform. Theory}, vol.~41, no.~3, pp. 613--627, May, 1995.

\bibitem{blu2007sure}
T.~Blu and F.~Luisier, ``The sure-let approach to image denoising,'' \emph{IEEE Trans. Image Process.}, vol.~16, no.~11, pp. 2778--2786, Nov. 2007.

\bibitem{brox2011large}
T.~Brox and J.~Malik, ``Large displacement optical flow: descriptor matching in variational motion estimation,'' \emph{IEEE Trans. Pattern. Anal. Mach. Intell.}, vol.~33, no.~3, pp. 500--513, Mar. 2011.

\bibitem{cavdataset}
T.~U. o.~E. School~of informatics, ``Caviar test case scenarios,'' 2004.
  [Online]. Available: \url{http://groups.inf.ed.ac.uk/vision/CAVIAR/CAVIARDATA1/}

\bibitem{limudataset}
L.~K. University, ``Dataset: Detection of moving objects,'' 2008. [Online]. Available: \url{http://limu.ait.kyushu-u.ac.jp/dataset/en/}


\bibitem{i2rdataset}
M.~Narayana, ``Moving camera systems (hand-held cameras, automobile cameras),'' 2013. [Online]. Available: \url{http://vis-www.cs.umass.edu/~narayana/castanza/I2Rdataset/}

\bibitem{wang2014cdnet}
Y.~Wang, P.~Jodoin, F.~Porikli, J.~Konrad, Y.~Benezeth, and P.~Ishwar, ``CDnet 2014: An expanded change detection benchmark dataset,'' in \emph{Proc. IEEE Conf. Comp. Vis. Pattern Recogn.}, pp.~387--394, Washington, DC, USA, June. 2014.



\end{thebibliography}

\end{document}